\documentclass{article}
\usepackage{microtype}
\usepackage{graphicx}
\usepackage{booktabs}
\usepackage{multirow}
\usepackage{placeins}
\usepackage{hyperref}
\usepackage[T1]{fontenc}

\usepackage[preprint]{icml2026}
\usepackage{amsmath,amssymb}
\usepackage{xcolor}
\usepackage{colortbl}

\hypersetup{colorlinks=true,linkcolor=blue!50!black,citecolor=blue!50!black,urlcolor=blue!50!black}
\usepackage{enumitem}
\usepackage{tcolorbox}
\definecolor{reload}{HTML}{2B6CD4}
\definecolor{reloadbg}{HTML}{E8F0FC}
\newcommand{\RQ}[1]{\textbf{\color{reload}RQ#1}}
\newtcolorbox{rqanswer}[1]{colback=reloadbg,colframe=reload,boxrule=0.6pt,arc=3pt,
  left=5pt,right=5pt,top=3pt,bottom=3pt,before skip=8pt,after skip=4pt,
  before upper={\textbf{\color{reload}Answer to RQ#1.}\ }}

\newcommand{\ci}[2]{\ensuremath{[#1,\,#2]}}
\newsavebox{\fitboxbox}
\newcommand{\fitbox}[1]{\sbox{\fitboxbox}{#1}\ifdim\wd\fitboxbox>\linewidth\resizebox{\linewidth}{!}{\usebox{\fitboxbox}}\else\usebox{\fitboxbox}\fi}

\icmltitlerunning{Extending Reasoning in Typed Decision Models via Looping}

\begin{document}

\twocolumn[
  \icmltitle{Extending Reasoning in Typed Decision Models via Adaptive Looping}
  \begin{icmlauthorlist}
    \icmlauthor{David Berghaus}{li,iais}
    \icmlauthor{Arber Zela}{epfl}
    \icmlauthor{Behzad Shomali}{li,bonn}
    \icmlauthor{Markus Frey}{li,iais}\\
    \icmlauthor{Mehdi Ali}{li,iais}
    \icmlauthor{Rafet Sifa}{li,iais,bonn}
    \icmlauthor{Volkan Cevher}{epfl}
  \end{icmlauthorlist}
  \icmlaffiliation{li}{Lamarr Institute for Machine Learning and Artificial Intelligence, Bonn, Germany}
  \icmlaffiliation{iais}{Fraunhofer IAIS, Sankt Augustin, Germany}
  \icmlaffiliation{bonn}{University of Bonn, Bonn, Germany}
  \icmlaffiliation{epfl}{EPFL, Lausanne, Switzerland}
  \icmlcorrespondingauthor{David Berghaus}{david.berghaus@iais.fraunhofer.de}
  \icmlkeywords{looped transformers, decision models, adaptive computation, test-time compute}
  \vskip 0.3in
]
\printAffiliationsAndNotice{}

\begin{abstract}
Typed decision models choose among a fixed set of answer options in a single forward pass, so unlike generative language models they cannot spend more test-time compute on hard inputs by generating more tokens. Looping a block of layers offers this compute in latent space without adding parameters, but prior work found no gain from loops retrofitted into conventionally pretrained models. We show that such loops help when trained together with the decision model. Our method, RELOAD (retrofitted loops for adaptive decisions), fine-tunes a pretrained encoder or decoder with a damped loop over its middle block and a lightweight controller picks the number of passes adaptively for each input. Across six models from 0.4B to 4B parameters, looping improves both accuracy and calibration, mostly on problems that need longer chains of reasoning than any seen in training, where encoders gain up to 23 points. Neither matched training compute nor more data closes this gap, and separate layers for each pass come close only with nearly twice the parameters. The controller keeps most of the gain at about three quarters of the compute.
\end{abstract}

\section{Introduction}
\begin{figure*}[t]
\centering
\includegraphics[width=\linewidth]{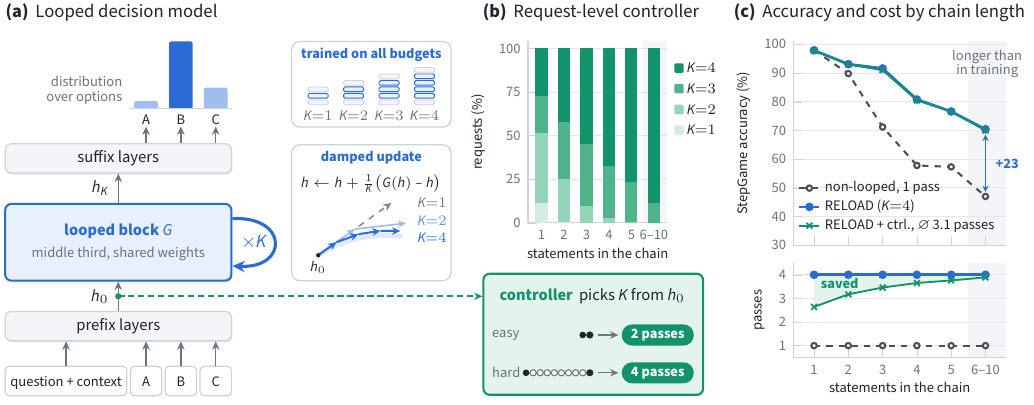}
\caption{Overview of RELOAD. (a) A middle block $G$ of a pretrained model is applied $K$ times with a damped update, and the hidden state at each answer option is scored. One model is trained on all budgets $K\in\{1,2,3,4\}$. (b) A controller picks the number of passes for each request from the prefix output $h_0$, before the loop runs. Bars show the share of StepGame test requests at each chosen $K$, by the number of statements that must be chained (Von, three seeds). (c) Top: StepGame test accuracy by chain length, for the same model. Bottom: mean number of passes per request. The controller matches the accuracy of four passes with 3.1 passes on average and saves most on short chains, where extra passes add little. Training problems need at most five statements.}
\label{fig:overview}
\end{figure*}
Typed decision models \citep{deusser2026jev,chung2026jevtom,von,kev} read a context, a question and a set of candidate answers and output a distribution over the candidates in a single forward pass. Open models such as Von \citep{von} and Kev \citep{kev} are fine-tuned from pretrained encoders and decoders \citep{warner2025modernbert,weller2025ettin}, and one pass per request makes them cheap to serve wherever software must choose among fixed options, for example to classify, route or verify. The same design fixes their computation. Generative language models spend more test-time compute on hard inputs by writing out intermediate steps \citep{wei2022chain} and allocating that compute by difficulty \citep{snell2024scaling}, whereas decision models process every input at the same depth. This limits decisions that require chaining several inferences, such as composing spatial relations or following a multi-step proof.

Looping a block of layers offers test-time compute in latent space: the block is applied repeatedly to refine the hidden state \citep{dehghani2019universal,saunshi2025reasoning}, and because the passes share weights, the extra computation adds no parameters. Prior looped decision models build on language models pretrained to loop and report no gain from loops retrofitted into conventionally pretrained models \citep{gan2026sansi,song2026lev}. We show that a retrofitted loop does help once it is trained together with the decision model. Our method, \textbf{RELOAD} (\textbf{re}trofitted \textbf{lo}ops for \textbf{a}daptive \textbf{d}ecisions), fine-tunes a pretrained model with a damped loop over its middle block on all budgets from one to four passes, and a request-level controller picks the number of passes for each input before the loop runs (Figure~\ref{fig:overview}). Our contributions are:

\begin{enumerate}
\item \textbf{Loop training extends the reasoning horizon.} Across three bidirectional encoders and three causal decoders from 0.4B to 4B parameters, RELOAD improves accuracy and Brier score over equally tuned single-pass models, most on problems longer than any in training (Section~\ref{sec:main}).
\item \textbf{The gain comes from the trained loop.} For Von, neither single-pass training with matched compute nor looping a single-pass model only at inference closes the gap, and more data does not make looping redundant. Depth up-scaling comes close only with nearly twice the parameters (Section~\ref{sec:controls}).
\item \textbf{Per-request budgets save compute.} The controller recovers most of the gain of four passes at about three quarters of their compute, allocating passes mainly by input length (Section~\ref{sec:controller}).
\item \textbf{Analysis of the trained loop.} All budgets follow one ODE trajectory, so for Von a midpoint integrator at test time reaches the accuracy of eight passes with four. The passes act on the statements that must be composed, and the gain requires looping a block of layers rather than each layer in place (Section~\ref{sec:analysis}).
\end{enumerate}

\section{Related work}
\paragraph{Test-time compute.} Chain-of-thought prompting lets generative models spend more computation on hard inputs \citep{wei2022chain}, and allocating it per input can be more efficient than a fixed budget \citep{snell2024scaling}. Recurrent-depth language models reason in continuous space instead of text \citep{geiping2025scaling,zhu2025ouro}. All of these generate tokens, whereas our models score a fixed set of options.
\paragraph{Looped transformers.} Universal Transformers apply one block recurrently \citep{dehghani2019universal}. Looping enables algorithmic computation \citep{giannou2023looped,yang2024looped} and, when trained from scratch, nearly matches a deeper model on reasoning but not on memorization \citep{saunshi2025reasoning}. Recurrent-depth language models are pretrained to loop \citep{geiping2025scaling,zhu2025ouro}, also for many budgets in one model \citep{jeddi2026loopformer,ladhak2026elastic}, with supervised intermediate loops \citep{shomali2026loopmtp} or with adaptive per-layer loops and memory banks \citep{frey2026adaptive}.
\paragraph{Loops as ODE solvers.} Residual networks can be read as discretized differential equations \citep{haber2017stable,chen2018neural}, which has guided architecture design \citep{lu2018beyond}. \citet{queiruga2020continuous} swap the integrator of trained continuous-in-depth networks; in their experiments this works for networks trained with a higher-order method but fails for those trained with forward Euler. Our loop is trained with Euler steps of four sizes, and the midpoint method still improves it.
\paragraph{Retrofitting loops into pretrained models.} SanSi \citep{gan2026sansi} and LEV \citep{song2026lev} build looped decision models on language models pretrained to loop, such as Ouro, and report no gain from loops retrofitted into conventionally pretrained models. Their main models loop the whole layer stack, so four steps take about four times as long as one \citep{song2026lev}; four of our middle-block passes take less than twice as long (Appendix Table~\ref{tab:cost}). \citet{song2026lev} also find no gain from gated loops added with LoRA to a conventionally pretrained Qwen3.5-0.8B. On Von, we find that the update rule makes the difference: our damped loop still gains with LoRA, whereas the gated update brings no gain even with full fine-tuning (Appendix Table~\ref{tab:retrofit}). Training-Free Looped Transformers \citep{chen2026trainingfree} apply the same damped update to a middle block of frozen decoders; in our configuration the loop helps only after loop training. Other work continues pretraining decoders with random recurrence depth \citep{mcleish2025retrofitted}, compresses them into recursive models \citep{bae2025relaxed}, or grows them by duplicating layers \citep{kim2024solar,wu2024llamapro}, our parameter-heavy baseline. Layer-wise analyses of tabular foundation models also motivate looping as iterative refinement \citep{rezaeibalef2026onelayer}.

\paragraph{Adaptive computation.} ACT \citep{graves2016adaptive}, PonderNet \citep{banino2021pondernet}, early-exit encoders \citep{xin2020deebert,zhou2020pabee}, CALM \citep{schuster2022calm} and RecurTrace \citep{wang2026recurtrace} decide after each step or token whether to stop. Recurrent reasoners trained from scratch on puzzles such as Sudoku and mazes, including HRM \citep{wang2025hrm} and TRM \citep{jolicoeur2025trm}, also learn when to halt. Fixed-point reasoners iterate toward a learned fixed point and stop once it is reached \citep{movahedi2026fprm,huang2026eqr}, also on the hypersphere \citep{zela2026sfprm}, building on deep equilibrium models \citep{bai2019deq}. The Depth-Adaptive Transformer \citep{elbayad2020depth} instead predicts depth per sequence. Our controller decides once per request, from features computed before the loop.

\section{Method}\label{sec:method}
\paragraph{Decision models.} A decision model reads an input $x$ (a question, its context and $m$ answer options) and outputs a distribution $p(\cdot\mid x)$ over the options in one forward pass, through a head $f$ that gives one logit per option. Encoders follow Von \citep{von}\footnote{\url{https://github.com/wfzyx/von} (Apache-2.0).}: the input is \texttt{question context [SEP] [MASK] option$_1$ \dots\ [MASK] option$_m$}, each option attends only to the shared prefix and itself, and option position ids restart after the prefix, so scores do not depend on option order. The head $f$ scores the hidden state at each \texttt{[MASK]}. Decoders follow Kev, an open reconstruction of Jev-style decision models \citep{kev}: a block-causal mask lets each option read the shared prefix but not the other options, and $f$ is a pointer head that scores each option's end token against a final decision token.

\paragraph{Looped middle block.} Let the network consist of layers $L_0,\dots,L_{N-1}$ applied to the token embeddings $e(x)$, followed by the decision head $f$. We split the layers into a prefix $P=L_{a-1}\circ\dots\circ L_0$, a block $G=L_b\circ\dots\circ L_a$ of $B=b-a+1$ layers and a suffix $S=L_{N-1}\circ\dots\circ L_{b+1}$. With budget $K$ and $k=0,\dots,K-1$, the model computes
\begin{align}
h_0 &= P\bigl(e(x)\bigr), \nonumber\\
h_{k+1} &= h_k + \tfrac{1}{K}\bigl(G(h_k) - h_k\bigr), \label{eq:loop}\\
p(\cdot\mid x) &= \mathrm{softmax}\bigl(f(S(h_K))\bigr). \nonumber
\end{align}
At $K{=}1$ the update gives $h_1=G(h_0)$ and recovers the original network. All passes share the weights of $G$, so looping adds no parameters, and budget $K$ costs $N+(K-1)B$ layer applications rather than $K$ full forward passes. In practice four passes take less than twice as long as one and need no extra inference memory (Appendix Table~\ref{tab:cost}). In every model, $G$ is roughly the middle third of the layers (counted from 0), without tuning per model: layers 9--17 of the 28-layer encoders, which form three complete groups of one global- and two local-attention layers; layers 9--18 of Qwen3-0.6B and Qwen3-1.7B (28 layers); and layers 12--23 of Qwen3-4B (36 layers). Each layer keeps its own attention type and mask.

\paragraph{The loop as an ODE.} Equation~\eqref{eq:loop} is the explicit Euler method with step size $1/K$ \citep{chen2026trainingfree} for
\begin{equation}
\frac{dh}{dt} = G\bigl(h(t)\bigr) - h(t), \qquad h(0) = h_0, \qquad t\in[0,1],
\label{eq:ode}
\end{equation}
so every budget discretizes the same trajectory and approaches its end state $h(1)$ as $K$ grows. The budget sets the step size, not the length of the trajectory; fixed-point reasoners instead iterate until the state stops changing \citep{zela2026sfprm,movahedi2026fprm}. As for residual networks read as dynamical systems \citep{haber2017stable,lu2018beyond,chen2018neural}, this view yields two predictions, tested in Section~\ref{sec:analysis}: for smooth $G$ the error relative to $h(1)$ falls as $1/K$ (Appendix~\ref{app:ode}), and the $K$ evaluations of $G$ can be spent on a more accurate integrator at test time.

\paragraph{Training on all budgets.} Let $p_K(\cdot\mid x)$ be the output at budget $K$ and $\ell$ the cross-entropy plus $0.5\times$ the Brier score. Every training step minimizes
\begin{equation}
\mathcal{L}(x,y) = \frac{1}{4}\sum_{K=1}^{4} \ell\bigl(p_K(\cdot\mid x),\,y\bigr),
\label{eq:allbudgets}
\end{equation}
with one backward pass per budget, so that one model can run at any budget. This supervises the endpoints of four trajectories with step sizes $1/K$, whereas \citet{song2026lev} supervise intermediate readouts of a single recurrence.

\paragraph{Request-level controller.} The controller picks the budget once per request, before the loop runs, from the prefix output $h_0$, which all budgets share, so its features cost nothing extra. Let $\bar h_{\mathcal{C}}$ and $\bar h_{\mathcal{O}}$ be the means of $h_0$ over the question and context tokens and over the option markers, and $n$ the input length. The features
\begin{equation}
z = \mathrm{PCA}_{64}\bigl(\mathrm{std}\bigl[\,\bar h_{\mathcal{C}},\ \bar h_{\mathcal{O}},\ \log n,\ m\,\bigr]\bigr)
\label{eq:ctrl_features}
\end{equation}
are standardized and reduced to 64 dimensions. To fit the controller, we run the trained model at every budget $K\in\{1,2,3,4\}$ on development data and record whether each answer is correct. One logistic regression per budget estimates the probability of a correct answer, $\hat P_K=\sigma(w_K^\top z+b_K)$, and at test time the controller picks
\begin{equation}
K^\star = \arg\max_{K}\ \hat P_K - \lambda\,(K-1).
\label{eq:ctrl}
\end{equation}
An extra pass is thus used only if it is predicted to raise the probability of a correct answer by more than the cost weight $\lambda$, which sets the trade-off between accuracy and compute. The controller adds no parameters to the network and is fitted on a CPU. The stopper of \citet{song2026lev}, in contrast, decides after each step whether the answer already agrees with the deepest one, and needs no correct-answer labels.

\section{Experiments}\label{sec:experiments}
We address four research questions, each answered at the end of its subsection:
\begin{enumerate}[label=\RQ{\arabic*},leftmargin=*,itemsep=1pt,topsep=3pt,parsep=0pt]
\item Does loop training help, and where? (\S\ref{sec:main})
\item Is the gain due to the trained loop, rather than to extra compute, data or parameters? (\S\ref{sec:controls})
\item Can passes be chosen per request? (\S\ref{sec:controller})
\item How does the trained loop work, and which design choices matter? (\S\ref{sec:analysis})
\end{enumerate}

\subsection{Setup}\label{sec:setup}
\paragraph{Models.} We use three bidirectional encoders and three causal decoders: Von \citep{von}, a 395M-parameter ModernBERT-large \citep{warner2025modernbert} already trained as a decision model; the general-purpose Ettin encoders \citep{weller2025ettin} with 400M and 1B parameters and a newly initialized scorer; and the Qwen3-0.6B, 1.7B and 4B base models \citep{yang2025qwen3} with Kev's input encoding and pointer head.

\paragraph{Training.} All models are fully fine-tuned for two epochs on one broad mixture: public classification and question-answering sets, adversarial natural-language inference, Kev's policy and compositional decision sets, and multi-step reasoning data (object tracking, Boolean circuits, ProofWriter, CLUTRR, and StepGame stories of at most five statements). All arms of a model share data, hyperparameters and seeds, with learning rates chosen on development data. The controller is fitted on development sets, including StepGame stories of up to ten statements as in the longer-reasoning test group. Appendix~\ref{app:training} gives the details.

\begin{table*}[t]
\centering
\scriptsize\setlength{\tabcolsep}{3pt}\renewcommand{\arraystretch}{0.92}
\caption{Accuracy (\%) and Brier score (lower is better), mean $\pm$ standard deviation over three seeds. In-distribution: held-out ProofWriter, StepGame with 1--5 statements and Kev decision. Longer reasoning: StepGame with 6--10 statements. Out-of-distribution: Kev transfer and eight unseen benchmarks, zero-shot. Qwen3 models use Kev's encoding. RELOAD ($K{=}k$): one loop-trained checkpoint run with $k$ passes; the single pass is trained separately. Every budget, including $K{=}1$: Appendix Table~\ref{tab:main_full}; per-set results: Appendix~\ref{app:perset}. Layers: layer applications per question; for the controller ($\lambda{=}0.005$, fitted on development data) per group (in-distribution / longer / out-of-distribution). Bold: best per model and column.}
\label{tab:main}
\resizebox{\textwidth}{!}{\let\scriptsize\tiny\begin{tabular}{llrrrrrrr}
\toprule
 & & \multicolumn{2}{c}{In-distribution} & \multicolumn{2}{c}{Longer reasoning} & \multicolumn{2}{c}{Out-of-distribution} & \\
\cmidrule(lr){3-4}\cmidrule(lr){5-6}\cmidrule(lr){7-8}
Model & Variant & Acc. & Brier & Acc. & Brier & Acc. & Brier & Layers \\
\midrule
 & Single pass & 92.8\,{\scriptsize$\pm$0.3} & 0.101\,{\scriptsize$\pm$0.004} & 70.6\,{\scriptsize$\pm$1.3} & 0.404\,{\scriptsize$\pm$0.025} & 50.9\,{\scriptsize$\pm$0.2} & 0.728\,{\scriptsize$\pm$0.005} & 28  \\
\rowcolor{reloadbg}\cellcolor{white} & RELOAD + controller & 93.6\,{\scriptsize$\pm$0.2} & 0.088\,{\scriptsize$\pm$0.002} & 85.4\,{\scriptsize$\pm$1.0} & 0.203\,{\scriptsize$\pm$0.014} & 53.8\,{\scriptsize$\pm$1.0} & 0.682\,{\scriptsize$\pm$0.008} & 35/53/39  \\
\rowcolor{reloadbg}\cellcolor{white}\multirow{-3}{*}{Von (395M)}& RELOAD ($K{=}4$) & \textbf{93.9}\,{\scriptsize$\pm$0.2} & \textbf{0.085}\,{\scriptsize$\pm$0.002} & \textbf{85.7}\,{\scriptsize$\pm$0.4} & \textbf{0.198}\,{\scriptsize$\pm$0.009} & \textbf{54.1}\,{\scriptsize$\pm$0.7} & \textbf{0.675}\,{\scriptsize$\pm$0.007} & 55  \\
\midrule
 & Single pass & 93.0\,{\scriptsize$\pm$0.4} & 0.100\,{\scriptsize$\pm$0.003} & 72.3\,{\scriptsize$\pm$0.6} & 0.384\,{\scriptsize$\pm$0.008} & 47.9\,{\scriptsize$\pm$0.2} & 0.714\,{\scriptsize$\pm$0.015} & 28  \\
\rowcolor{reloadbg}\cellcolor{white} & RELOAD + controller & 93.9\,{\scriptsize$\pm$0.2} & 0.087\,{\scriptsize$\pm$0.001} & \textbf{87.0}\,{\scriptsize$\pm$1.7} & 0.183\,{\scriptsize$\pm$0.027} & 50.0\,{\scriptsize$\pm$0.6} & 0.696\,{\scriptsize$\pm$0.013} & 35/52/37  \\
\rowcolor{reloadbg}\cellcolor{white}\multirow{-3}{*}{Ettin-400M}& RELOAD ($K{=}4$) & \textbf{94.0}\,{\scriptsize$\pm$0.2} & \textbf{0.085}\,{\scriptsize$\pm$0.000} & \textbf{87.0}\,{\scriptsize$\pm$1.8} & \textbf{0.179}\,{\scriptsize$\pm$0.029} & \textbf{50.5}\,{\scriptsize$\pm$0.8} & \textbf{0.686}\,{\scriptsize$\pm$0.017} & 55  \\
\midrule
 & Single pass & 93.9\,{\scriptsize$\pm$0.3} & 0.089\,{\scriptsize$\pm$0.003} & 80.2\,{\scriptsize$\pm$1.4} & 0.292\,{\scriptsize$\pm$0.013} & 48.7\,{\scriptsize$\pm$0.7} & 0.717\,{\scriptsize$\pm$0.028} & 28  \\
\rowcolor{reloadbg}\cellcolor{white} & RELOAD + controller & 94.5\,{\scriptsize$\pm$0.1} & 0.079\,{\scriptsize$\pm$0.002} & 90.6\,{\scriptsize$\pm$0.7} & 0.135\,{\scriptsize$\pm$0.005} & 52.6\,{\scriptsize$\pm$0.9} & 0.655\,{\scriptsize$\pm$0.009} & 34/54/39  \\
\rowcolor{reloadbg}\cellcolor{white}\multirow{-3}{*}{Ettin-1B}& RELOAD ($K{=}4$) & \textbf{94.6}\,{\scriptsize$\pm$0.2} & \textbf{0.078}\,{\scriptsize$\pm$0.003} & \textbf{90.7}\,{\scriptsize$\pm$0.8} & \textbf{0.134}\,{\scriptsize$\pm$0.005} & \textbf{53.2}\,{\scriptsize$\pm$0.9} & \textbf{0.649}\,{\scriptsize$\pm$0.010} & 55  \\
\midrule
 & Single pass & 86.1\,{\scriptsize$\pm$0.4} & 0.194\,{\scriptsize$\pm$0.009} & 48.1\,{\scriptsize$\pm$0.8} & 0.629\,{\scriptsize$\pm$0.015} & 52.6\,{\scriptsize$\pm$0.7} & 0.642\,{\scriptsize$\pm$0.009} & 28  \\
\rowcolor{reloadbg}\cellcolor{white} & RELOAD + controller & 87.9\,{\scriptsize$\pm$0.2} & 0.171\,{\scriptsize$\pm$0.003} & 54.3\,{\scriptsize$\pm$1.0} & 0.571\,{\scriptsize$\pm$0.013} & 52.8\,{\scriptsize$\pm$0.8} & \textbf{0.641}\,{\scriptsize$\pm$0.018} & 37/44/41  \\
\rowcolor{reloadbg}\cellcolor{white}\multirow{-3}{*}{Qwen3-0.6B}& RELOAD ($K{=}4$) & \textbf{88.0}\,{\scriptsize$\pm$0.2} & \textbf{0.169}\,{\scriptsize$\pm$0.003} & \textbf{55.2}\,{\scriptsize$\pm$0.6} & \textbf{0.561}\,{\scriptsize$\pm$0.009} & \textbf{53.4}\,{\scriptsize$\pm$1.0} & 0.642\,{\scriptsize$\pm$0.016} & 58  \\
\midrule
 & Single pass & 89.3\,{\scriptsize$\pm$0.5} & 0.153\,{\scriptsize$\pm$0.009} & 57.7\,{\scriptsize$\pm$1.4} & 0.532\,{\scriptsize$\pm$0.015} & 54.6\,{\scriptsize$\pm$1.4} & 0.630\,{\scriptsize$\pm$0.022} & 28  \\
\rowcolor{reloadbg}\cellcolor{white} & RELOAD + controller & 90.6\,{\scriptsize$\pm$0.5} & 0.135\,{\scriptsize$\pm$0.005} & 59.9\,{\scriptsize$\pm$1.1} & 0.511\,{\scriptsize$\pm$0.018} & 56.2\,{\scriptsize$\pm$1.1} & 0.610\,{\scriptsize$\pm$0.012} & 38/51/43  \\
\rowcolor{reloadbg}\cellcolor{white}\multirow{-3}{*}{Qwen3-1.7B}& RELOAD ($K{=}4$) & \textbf{90.9}\,{\scriptsize$\pm$0.2} & \textbf{0.131}\,{\scriptsize$\pm$0.003} & \textbf{60.2}\,{\scriptsize$\pm$1.4} & \textbf{0.506}\,{\scriptsize$\pm$0.017} & \textbf{56.5}\,{\scriptsize$\pm$0.8} & \textbf{0.605}\,{\scriptsize$\pm$0.008} & 58  \\
\midrule
 & Single pass & 90.9\,{\scriptsize$\pm$1.4} & 0.132\,{\scriptsize$\pm$0.014} & 62.0\,{\scriptsize$\pm$4.5} & 0.507\,{\scriptsize$\pm$0.045} & 60.4\,{\scriptsize$\pm$0.9} & 0.578\,{\scriptsize$\pm$0.016} & 36  \\
\rowcolor{reloadbg}\cellcolor{white} & RELOAD + controller & 92.8\,{\scriptsize$\pm$0.5} & 0.105\,{\scriptsize$\pm$0.006} & 71.8\,{\scriptsize$\pm$0.2} & 0.398\,{\scriptsize$\pm$0.009} & 61.5\,{\scriptsize$\pm$0.1} & 0.570\,{\scriptsize$\pm$0.007} & 46/62/51  \\
\rowcolor{reloadbg}\cellcolor{white}\multirow{-3}{*}{Qwen3-4B}& RELOAD ($K{=}4$) & \textbf{93.0}\,{\scriptsize$\pm$0.5} & \textbf{0.102}\,{\scriptsize$\pm$0.007} & \textbf{72.1}\,{\scriptsize$\pm$0.4} & \textbf{0.393}\,{\scriptsize$\pm$0.009} & \textbf{62.0}\,{\scriptsize$\pm$0.2} & \textbf{0.569}\,{\scriptsize$\pm$0.006} & 72  \\
\bottomrule
\end{tabular}}
\end{table*}
\paragraph{Test groups.} We group the test sets by their distance from the training data (Appendix~\ref{app:data}).
\emph{In-distribution}: held-out splits of training tasks, namely ProofWriter \citep{tafjord2021proofwriter}, which asks whether a statement follows from facts and rules; StepGame \citep{shi2022stepgame} with 1--5 statements, which asks for the spatial relation between two objects; and Kev decision, with typed decisions from Kev's training sources \citep{kev}.
\emph{Longer reasoning}: StepGame with 6--10 statements, longer than any training story. Depending on where the two queried objects lie on the described path, a problem needs one to all of its statements, so some chains are longer than any in training.
\emph{Out-of-distribution}: sets never used for training or model selection, namely Kev transfer, with sources and rule structures absent from training, and eight reasoning benchmarks evaluated zero-shot: ParaRules \citep{tafjord2021proofwriter}, JustLogic \citep{justlogic}, ProverQA \citep{proverqa}, BoardgameQA \citep{boardgameqa}, MuSR \citep{musr}, ReClor \citep{reclor}, LogiQA~2.0 \citep{logiqa2} and ReSQ \citep{resq}.

\paragraph{Metrics.} Decision models are used for their probabilities, for example to set thresholds or to defer uncertain cases, so besides accuracy we report the Brier score \citep{brier1950verification}, the squared error $\sum_j (p_j-y_j)^2$ between the predicted distribution $p$ and the one-hot answer $y$, averaged over questions. It ranges from 0 to 2 and is low only for a model that is both accurate and calibrated. We report macro means over the sets of each group, averaged over three seeds, and paired differences with 95\% intervals from a bootstrap over seeds and then items within each set.

\subsection{RQ1: Does loop training help, and where?}\label{sec:main}

\begin{figure*}[t]
\centering
\includegraphics[width=\linewidth]{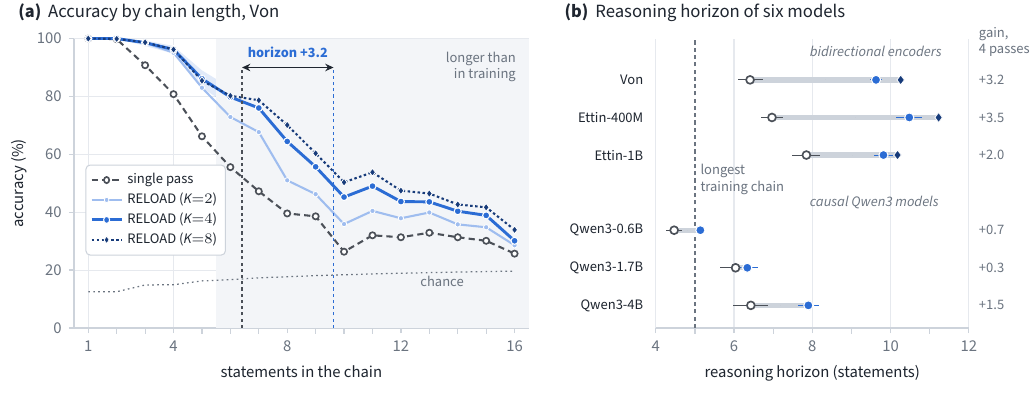}
\caption{Reasoning horizon on generated StepGame stories whose statements all lie on the chain (three seeds; not used for training or model selection; training chains have at most five statements). (a) Von's accuracy by chain length. Band: 95\% bootstrap interval for $K{=}4$. Chance: always answering the most frequent label. Dashed lines: horizon of the single pass and of $K{=}4$, where a fitted sigmoid lies halfway between ceiling and chance (Appendix~\ref{app:horizon}). (b) Horizon of the single pass (open circle) and of RELOAD at $K{=}4$ (filled circle) and $K{=}8$ (diamond, hidden for Qwen3 where it coincides with $K{=}4$). Gray bar: single pass to $K{=}8$; whiskers: standard deviation over seeds (single pass, $K{=}4$); right: gain of $K{=}4$ over the single pass.}
\label{fig:horizon}
\end{figure*}

\begin{table*}[t]
\centering
\footnotesize\setlength{\tabcolsep}{4pt}
\caption{Controls for Von: accuracy (\%) and Brier score (lower is better), mean $\pm$ standard deviation over three seeds. Train: training layer applications relative to two-epoch single-pass training. Layers: layer applications per question at inference ($K{=}4$ for looped rows). Looped only at inference: the single-pass checkpoint run with four passes. Half the data: a nested subset with about half of the training questions. Groups as in Table~\ref{tab:main}, but out of distribution only Kev transfer, since the zero-shot benchmarks were not run for every control. More controls: Appendix~\ref{app:controls}. Bold: best per column.}
\label{tab:controls}
\fitbox{\begin{tabular}{lrrrrrrrrr}
\toprule
 & & & & \multicolumn{2}{c}{In-distribution} & \multicolumn{2}{c}{Longer reasoning} & \multicolumn{2}{c}{Kev transfer} \\
\cmidrule(lr){5-6}\cmidrule(lr){7-8}\cmidrule(lr){9-10}
Variant & Params & Train & Layers & Acc. & Brier & Acc. & Brier & Acc. & Brier \\
\midrule
Single pass & 395M & 1.0 & 28 & 92.8\,{\scriptsize$\pm$0.3} & 0.101\,{\scriptsize$\pm$0.004} & 70.6\,{\scriptsize$\pm$1.3} & 0.404\,{\scriptsize$\pm$0.025} & 76.4\,{\scriptsize$\pm$1.8} & 0.344\,{\scriptsize$\pm$0.023}  \\
Single pass, 12 epochs & 395M & 6.0 & 28 & \textbf{94.1}\,{\scriptsize$\pm$0.3} & 0.107\,{\scriptsize$\pm$0.004} & 74.9\,{\scriptsize$\pm$0.3} & 0.455\,{\scriptsize$\pm$0.007} & 72.5\,{\scriptsize$\pm$4.2} & 0.479\,{\scriptsize$\pm$0.065}  \\
Single pass, looped only at inference & 395M & 1.0 & 55 & 91.7\,{\scriptsize$\pm$0.4} & 0.117\,{\scriptsize$\pm$0.003} & 70.7\,{\scriptsize$\pm$1.4} & 0.399\,{\scriptsize$\pm$0.013} & 73.6\,{\scriptsize$\pm$0.7} & 0.382\,{\scriptsize$\pm$0.018}  \\
Depth up-scaling & 726M & 5.9 & 55 & \textbf{94.1}\,{\scriptsize$\pm$0.4} & \textbf{0.083}\,{\scriptsize$\pm$0.005} & 84.3\,{\scriptsize$\pm$1.0} & 0.227\,{\scriptsize$\pm$0.012} & \textbf{79.8}\,{\scriptsize$\pm$0.2} & \textbf{0.293}\,{\scriptsize$\pm$0.011}  \\
\rowcolor{reloadbg}RELOAD ($K{=}4$), half the data & 395M & 3.1 & 55 & 92.1\,{\scriptsize$\pm$0.5} & 0.108\,{\scriptsize$\pm$0.005} & 76.5\,{\scriptsize$\pm$2.3} & 0.316\,{\scriptsize$\pm$0.028} & 79.4\,{\scriptsize$\pm$1.1} & 0.308\,{\scriptsize$\pm$0.014}  \\
\rowcolor{reloadbg}RELOAD ($K{=}4$) & 395M & 5.9 & 55 & 93.9\,{\scriptsize$\pm$0.2} & 0.085\,{\scriptsize$\pm$0.002} & \textbf{85.7}\,{\scriptsize$\pm$0.4} & \textbf{0.198}\,{\scriptsize$\pm$0.009} & 77.5\,{\scriptsize$\pm$1.9} & 0.323\,{\scriptsize$\pm$0.029}  \\
\bottomrule
\end{tabular}}
\end{table*}

\paragraph{Main results.} Table~\ref{tab:main} compares RELOAD with a single-pass model trained with the same recipe. Looping improves mean accuracy in every group for every model, most on longer reasoning: the encoders gain 10 to 15 points and the Qwen3 models 2.5 to 10, and on chains longer than any in training the encoders gain 17 to 23 points (Figure~\ref{fig:overview}c for Von). Run with one pass, the same checkpoint is within about two points of the single-pass model on longer reasoning (Appendix Table~\ref{tab:main_full}), so the gain comes from the extra passes. Out of distribution the gains are smaller and vary across benchmarks, and single sets can get worse, such as Kev transfer for Qwen3-0.6B (Appendix~\ref{app:perset}). Looping can also substitute for parameters: averaged over the test sets of Appendix Table~\ref{tab:perset}, RELOAD on Ettin-400M is more accurate and has a lower Brier score than the single-pass Ettin-1B. All our models are far more accurate than untrained base models and released decision models (Appendix~\ref{app:baselines}).

\paragraph{Probability quality.} Looping lowers the mean Brier score for every model, also after temperature scaling (Appendix Tables~\ref{tab:perset} and~\ref{tab:calib}), and four passes roughly halve the Brier score of the encoders on longer reasoning (Table~\ref{tab:main}).

\paragraph{Reasoning horizon.} Since the StepGame test set has few chains longer than five statements, we generate stories from its templates in which every statement belongs to the chain, with chain lengths from 1 to 16 (Appendix~\ref{app:horizon}). We summarize each accuracy curve by its \emph{reasoning horizon}, the chain length at which a fitted sigmoid lies halfway between the model's ceiling and chance. Four passes extend the horizon of the encoders by 2 to 3.5 statements, for Von from 6.4 to 9.6 (Figure~\ref{fig:horizon}); two or three passes give most of this, and eight passes, beyond the trained budgets, add a little more. With the story length fixed at ten statements, so that longer chains do not mean longer inputs, Von's horizon still grows from 4.4 to 8.6.

\paragraph{Distractors.} The loop also finds a chain inside a longer story. Five statements off the chain, which also make the stories longer than any in training, cost the single-pass encoders 16 to 19 points on chains of one to five statements, but RELOAD ($K{=}4$) only 3 to 4 (Appendix Table~\ref{tab:distractors}).

\paragraph{Object tracking.} The gain is not specific to StepGame. In synthetic object tracking, four objects in four boxes are swapped several times and the question asks where one object ends up; training covers up to four swaps. Four passes raise Von's accuracy from 77\% to 98\% with six swaps and from 44\% to 75\% with eight, and Ettin-400M gains similarly (Appendix Table~\ref{tab:probe}).

\paragraph{Encoders and decoders.} The causal Qwen3 models gain less throughout: 2.5 to 10 points on longer reasoning, only 0.3 to 1.5 statements of horizon (Figure~\ref{fig:horizon}b), and almost no protection against distractors, which cost them 19 to 23 points with or without the loop (Appendix Table~\ref{tab:distractors}). On longer reasoning the bidirectional encoders are also much more accurate, even than Qwen3-4B with four to ten times their parameters, whereas out of distribution, where many benchmarks require world knowledge and reading comprehension, the larger Qwen3 models are stronger. A plausible reason for the encoders' advantage is that every story token can attend to every other, whereas a causal model reads the story left to right, although the two families also differ in pretraining and input format.

\begin{rqanswer}{1}
Looping improves accuracy in every test group for all six models and lowers their mean Brier score, most on longer reasoning, where the encoders gain 10 to 15 points and extend their reasoning horizon by 2 to 3.5 statements.
\end{rqanswer}

\subsection{RQ2: Is the gain due to the trained loop?}\label{sec:controls}
Besides the loop, RELOAD differs from the single-pass baseline in training and inference compute, and its gain might shrink with more data or be matched by more parameters. Table~\ref{tab:controls} separates these factors for Von.

\paragraph{Training compute.} Training on all four budgets costs about six times the layer applications of single-pass training. Twelve single-pass epochs match this compute and reach RELOAD's in-distribution accuracy, but the final checkpoint stays more than 10 points behind on longer reasoning and becomes overconfident out of distribution (Appendix~\ref{app:controls}).

\paragraph{Inference compute without loop training.} Running the single-pass model with four passes, as in the training-free setting of \citet{chen2026trainingfree}, costs as much as RELOAD at inference, but in our middle-block configuration mean accuracy falls with every added pass (Figure~\ref{fig:frontier}). The same passes help only after loop training.

\paragraph{Training data.} The gain persists as the training mixture grows fourfold, and RELOAD trained on half the data beats the single pass trained on all of it on longer reasoning and Kev transfer, though not in distribution (Appendix~\ref{app:controls}).

\paragraph{Parameters.} The natural alternative to looping is to spend the same compute on new layers. Following depth up-scaling \citep{kim2024solar} (see also block expansion, \citealp{wu2024llamapro}), we give passes 2--4 their own trained copies of the block, adding 331M parameters at the same inference cost. This model, nearly twice the size, matches RELOAD in distribution, comes within about one and a half points on longer reasoning and is better on Kev transfer. For Ettin-400M it is on par with RELOAD on longer reasoning and again better on Kev transfer (Appendix~\ref{app:controls}). Extra layers can thus replace weight sharing, but only at nearly twice the parameters.

\begin{rqanswer}{2}
For Von, the gain persists at matched training compute and with four times the data, and looping without loop training does not help. Extra layers can replace the shared loop, but only at nearly twice the parameters.
\end{rqanswer}

\subsection{RQ3: Can passes be chosen per request?}\label{sec:controller}
\begin{figure}[t]
\centering
\includegraphics[width=\linewidth]{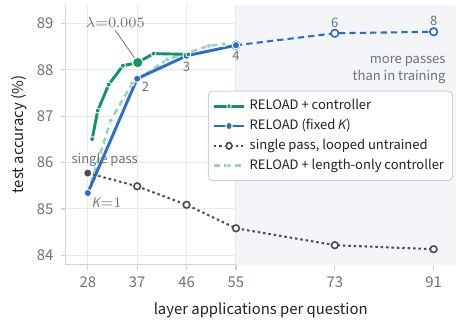}
\caption{Accuracy versus compute (Von, three seeds): mean test accuracy over ProofWriter, StepGame, Kev decision and Kev transfer against layer applications per question. Blue: RELOAD at fixed budgets (shaded: beyond the trained budgets). Green: RELOAD with the controller as $\lambda$ varies, and with a controller that sees only input length and option count. Gray: the single pass, and the same model looped without loop training.}
\label{fig:frontier}
\end{figure}
Table~\ref{tab:main} includes the controller at $\lambda{=}0.005$ for every model: it stays within one point of four passes in every group at about three quarters of their compute. For Von it traces the upper envelope of the fixed budgets, and at the compute of two passes it is almost as accurate as four (Figure~\ref{fig:frontier}).

\paragraph{Baselines.} A mix of the two neighboring fixed budgets with the same mean compute, which ignores the input, is 2.5 points less accurate than the controller on longer reasoning for the encoders, with a higher Brier score, and about as accurate elsewhere; for the Qwen3 models the controller is not significantly better (Appendix Table~\ref{tab:ctrl_baselines}). A controller that sees only the input length and the number of options needs more compute for the same accuracy (Figure~\ref{fig:frontier}).

\paragraph{Halting rules.} Rules that stop once the answer stops changing or becomes confident need about 55\% more compute for the same accuracy (Appendix Table~\ref{tab:ctrl_baselines}). Our budgets do not build on each other: two passes take half-size steps rather than continuing the single pass, so deciding whether to stop requires running every budget from the start. Recurrent models trained with intermediate exits on a single trajectory avoid this cost, but are outside this comparison.

\paragraph{What the controller responds to.} The states before the loop could in principle encode how many statements must be chained, but the controller's budget follows the length of the story: on StepGame it uses one pass for single-statement stories and four for stories of seven or more statements. Since long stories tend to contain long chains, most passes go where they help most (Figure~\ref{fig:overview}b). Among stories of equal length, however, short and long chains get about the same budget (Appendix Table~\ref{tab:hops}), so the states before the loop do not appear to expose the chain length (Appendix~\ref{app:interventions}).

\begin{rqanswer}{3}
Yes. The controller stays within one point of four passes at about three quarters of their compute, mainly by giving longer inputs more passes.
\end{rqanswer}

\subsection{RQ4: How does the trained loop work?}\label{sec:analysis}
\begin{figure*}[t]
\centering
\begin{minipage}[c]{0.37\linewidth}
\centering
\includegraphics[width=\linewidth]{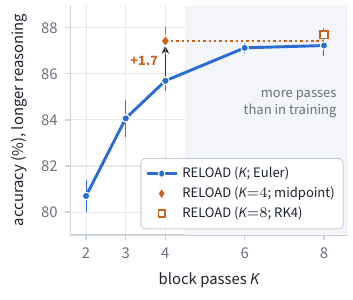}
\end{minipage}\hfill
\begin{minipage}[c]{0.6\linewidth}
\centering
\footnotesize\setlength{\tabcolsep}{4pt}
\fitbox{\begin{tabular}{@{}lrrrr@{}}
\toprule
 & In-dist. & \multicolumn{2}{c}{Longer reasoning} & OOD \\
\cmidrule(lr){2-2}\cmidrule(lr){3-4}\cmidrule(l){5-5}
Method & Acc. & Acc. & Brier & Acc. \\
\midrule
RELOAD ($K{=}4$; Euler) & 93.9\,{\scriptsize$\pm$0.2} & 85.7\,{\scriptsize$\pm$0.4} & 0.198\,{\scriptsize$\pm$0.009} & 54.1\,{\scriptsize$\pm$0.7} \\
RELOAD ($K{=}4$; midpoint) & 93.8\,{\scriptsize$\pm$0.2} & 87.4\,{\scriptsize$\pm$0.6} & 0.177\,{\scriptsize$\pm$0.004} & 53.9\,{\scriptsize$\pm$0.9} \\
\midrule
RELOAD ($K{=}8$; Euler) & 93.9\,{\scriptsize$\pm$0.1} & 87.2\,{\scriptsize$\pm$0.5} & 0.176\,{\scriptsize$\pm$0.004} & 54.1\,{\scriptsize$\pm$0.8} \\
RELOAD ($K{=}8$; RK4) & 93.8\,{\scriptsize$\pm$0.2} & 87.7\,{\scriptsize$\pm$0.3} & 0.171\,{\scriptsize$\pm$0.003} & 53.9\,{\scriptsize$\pm$0.9} \\
\midrule
Midpoint $-$ Euler ($K{=}4$) & $-0.2$ & $+1.7$ & $-0.022$ & $-0.2$ \\
 & {\scriptsize\ci{-0.41}{+0.08}} & {\scriptsize\ci{+0.7}{+2.8}} & {\scriptsize\ci{-0.035}{-0.009}} & {\scriptsize\ci{-0.62}{+0.16}} \\
\bottomrule
\end{tabular}}
\end{minipage}
\caption{Test-time integrators for the trained loop of Von, without retraining (three seeds). $K$ counts evaluations of $G$: Euler takes $K$ steps of size $1/K$; the midpoint method and fourth-order Runge--Kutta (RK4) take two steps of two and four evaluations. Left: longer-reasoning accuracy, with error bars of one standard deviation over seeds; shaded: more passes than in training. Right: accuracy (\%) and Brier score, mean $\pm$ standard deviation; in and out of distribution all Brier scores lie within 0.002 of each other. Last row: paired midpoint $-$ Euler difference at $K{=}4$ with 95\% interval. Both higher-order configurations were fixed on development data before testing.}
\label{fig:integrators}
\end{figure*}
\begin{figure*}[t]
\centering
\includegraphics[width=\linewidth]{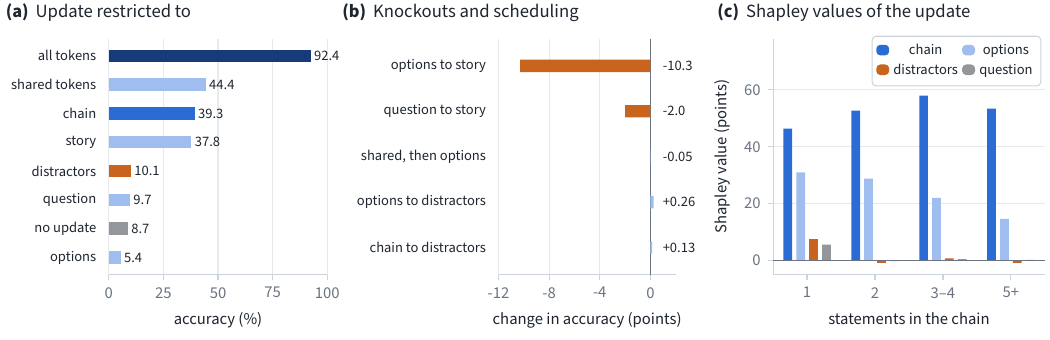}
\caption{Where the trained loop writes (Von, seed 1, StepGame, four passes). (a) Accuracy when each pass updates only one token group, $h\leftarrow h+\tfrac14\,m\odot(G(h)-h)$; the block still reads all tokens, and the others keep their state from before the loop. Stories have a unique chain and at least one distractor; shared tokens (question, story, separators) do not attend to the options. (b) Accuracy change relative to the ordinary loop when attention edges (query to key) are removed in the block, or when four passes over the shared tokens precede four over the options (first three rows: all stories; last two: parsed stories). (c) Shapley values of the accuracy gain over token groups by chain length; per bin they sum, with a small share of the special tokens, to the gain of the full update over no update.}
\label{fig:int_writes}
\end{figure*}

We analyze the trained loop first through the ODE view of Section~\ref{sec:method}, then through interventions on its updates and ablations of its structure.

\paragraph{One trajectory.} If all budgets discretize Equation~\eqref{eq:ode}, their outputs should approach those of a fine reference budget at Euler's rate, and they do: the change of the option log-probabilities relative to eight passes follows $1/K-1/8$ with a fitted exponent of 1.01 (Appendix Table~\ref{tab:convergence}). A single-pass model looped only at inference converges the same way, so the rate is a property of the update, while loop training changes where the flow ends. Unlike in a loop that converges to a fixed point and so forgets small changes to its state \citep{zela2026sfprm}, a random perturbation added between passes largely persists to the end of our trained loop, and large ones even grow (Appendix~\ref{app:interventions}; Von, one seed). In our models, two or three passes capture most of the accuracy gain.

\paragraph{Smaller steps or a longer trajectory.} Extra passes can refine the steps over the trained interval or continue the trajectory past $t=1$. Smaller steps do not hurt: six or eight passes leave accuracy unchanged in and out of distribution and still help Von on longer reasoning (Figures~\ref{fig:frontier} and~\ref{fig:integrators}). A longer trajectory helps only far outside the training range, on object tracking with eight swaps, and hurts StepGame, largely because the norm of the state keeps growing (Appendix~\ref{app:interventions}; Von, one seed).

\paragraph{A better integrator at the same cost.} The four evaluations of $G$ used by four Euler steps can instead feed the explicit midpoint method, a second-order method that first takes half a step to estimate the slope in the middle of the interval and then takes the full step with that slope,
\begin{equation}
\begin{aligned}
h_{k+\frac12} &= h_k + \tfrac{\Delta_k}{2}\bigl(G(h_k) - h_k\bigr),\\
h_{k+1} &= h_k + \Delta_k\bigl(G(h_{k+\frac12}) - h_{k+\frac12}\bigr),
\end{aligned}
\label{eq:midpoint}
\end{equation}
and its global error falls with the square of the step size rather than linearly. We apply it to the trained loop without changing any weights, with two steps of length $\Delta_0=1/3$ and $\Delta_1=2/3$ chosen on development data, which costs four passes; the second-order error bound motivates but does not cover this configuration (Appendix~\ref{app:ode}). For Von, the midpoint method is more accurate on longer reasoning than four Euler passes, with a lower Brier score, and as accurate as eight (Figure~\ref{fig:integrators}). For the other five models, where eight Euler passes add little over four, it is not significantly better, and for the Qwen3 models it slightly worsens the Brier score (Appendix Table~\ref{tab:integrators_all}). A better integrator thus helps only where four Euler passes are still far from the end of the trajectory.

\begin{table}[t]
\centering
\footnotesize\setlength{\tabcolsep}{3pt}
\caption{Design choices (Von, development data, three seeds each; Appendix~\ref{app:loopvariants}). All variants match the compute of the block loop at $K{=}4$ (the 28-layer loop at $K{=}2$: 56 instead of 55 layer applications). Gain: gain in longer-reasoning accuracy over $K{=}1$. $\Delta$: paired difference from a retrained block loop on longer reasoning and over all 15 development files; $^{*}$: 95\% interval excludes zero. Bold: best gain.}
\label{tab:design}
\fitbox{\begin{tabular}{lrrr}
\toprule
Variant & Gain & $\Delta$ Longer & $\Delta$ All  \\
\midrule
\rowcolor{reloadbg}Block loop, layers 9--17 (RELOAD) & $\mathbf{+8.7}$ & -- & --  \\
\midrule
\multicolumn{4}{l}{\textit{Layers 9--17 split into looped units}} \\
$4+5$ & $+7.9$ & $-0.8$ & $-0.5$  \\
$3+3+3$ & $+5.9$ & $-4.3^{*}$ & $-1.8^{*}$  \\
$2+2+2+2+1$ & $+4.0$ & $-4.7^{*}$ & $-1.8^{*}$  \\
Each layer in place, 9 layers & $+1.5$ & $-7.8^{*}$ & $-3.3^{*}$  \\
Each layer in place, 28 layers & $+1.5$ & $-7.4^{*}$ & $-4.4^{*}$  \\
\midrule
\multicolumn{4}{l}{\textit{Update and readout}} \\
Full step $h_{k+1}=G(h_k)$ & $+7.3$ & $-1.5$ & $-0.9^{*}$  \\
Full step, ACT-weighted output & $+7.5$ & $-1.7^{*}$ & $-1.1^{*}$  \\
\bottomrule
\end{tabular}}
\end{table}
\paragraph{Where the passes act.} On StepGame we freeze the trained weights and restrict the update of every pass to one group of tokens (Figure~\ref{fig:int_writes}; Von, one seed; Appendix~\ref{app:interventions}). The gain requires updating both the shared tokens (question and story) and the options: updating only the shared tokens recovers less than half of the gain over skipping the looped block, and updating only the options none of it. Within the story, updating only the statements on the chain is about as good as updating all of them, whereas updating only the distractors barely helps; a Shapley decomposition gives the story about two thirds of the gain, the options most of the rest and the distractors almost nothing. Four passes over the shared tokens followed by four over the options match the ordinary loop, so the options can read the story once its states are final. On object tracking, the answer read out after each pass improves gradually, and the object's final location becomes linearly readable step by step (Appendix Table~\ref{tab:probe}). The passes are thus spent on the statements that must be composed.

\paragraph{Which design choices matter.} Table~\ref{tab:design} summarizes ablations of the loop structure on Von's development data. Looping each layer in place before the next, as in per-layer loops \citep{frey2026adaptive}, removes most of the gain at the same compute, for every step size we tested and whether 9 or all 28 layers loop; splitting the block into units interpolates between the two. The state moves even farther in the layer loop, so a small update does not explain this: what the layer loop lacks is that later layers feed back into earlier ones. Damping matters for Qwen3-0.6B, which loses 6 points on longer reasoning without it, but little for Von and Ettin-400M (Appendix Tables~\ref{tab:injection_fullstep} and~\ref{tab:loop_variants}). Reading out a learned, input-dependent weighted average of the states after each pass, as in adaptive computation time \citep{graves2016adaptive}, does not improve accuracy (Appendix~\ref{app:loopvariants}), and input injection, which writes the prefix output back into the loop state before each pass \citep{huang2026topology}, did not reliably help in any variant we tested on five models (Appendix~\ref{app:injection}).

\begin{rqanswer}{4}
All budgets follow one ODE trajectory, so for Von a midpoint integrator with four passes matches eight Euler passes. The passes act on the statements that must be composed, and the gain needs a looped block of layers rather than per-layer loops.
\end{rqanswer}

\section{Limitations and future work}
Looping helps most on multi-step reasoning and little on knowledge-heavy multiple choice (Appendix~\ref{app:perset}), so it complements rather than replaces model scale. It also helps causal decoders much less than bidirectional encoders, for reasons we have not isolated. We study models up to 4B parameters trained with up to four passes, and run most controls and all interventions on Von, the interventions with one seed. Extra passes refine one learned map, so their benefit levels off once that map is computed accurately; objectives that make longer trajectories useful, such as a longer integration time for harder inputs, could raise this ceiling. The controller follows the length of the input more than the depth of reasoning it requires; features that expose the structure of a problem, or exits along a shared trajectory, could allocate passes more precisely. Finally, longer reasoning is tested on templated stories and synthetic tasks, and longer natural-language reasoning would show how far the gains carry.

\section{Conclusion}
A loop retrofitted into a conventionally pretrained decision model gives it test-time compute without generating tokens, as long as the loop is trained together with the model. Trained on all budgets at once, a damped loop over a middle block yields one model that runs at any number of passes. Four passes improve accuracy and Brier score without adding parameters, most for bidirectional encoders and on problems with longer chains of reasoning than any seen in training. In our controls, matched training compute and more data do not close this gap, and adding layers instead comes close only with nearly twice the parameters. A simple controller keeps most of the gain at about three quarters of the compute. Because the passes are Euler steps of one ODE, a better integrator can stand in for extra passes: for Von, a midpoint integrator matches eight passes with four.

\section*{Acknowledgements}
This research has been funded by the Federal Ministry of Education and Research of Germany and the state of North-Rhine Westphalia as part of the Lamarr Institute for Machine Learning and Artificial Intelligence.

\bibliography{references}

\begin{thebibliography}{64}
\providecommand{\natexlab}[1]{#1}
\providecommand{\url}[1]{\texttt{#1}}
\expandafter\ifx\csname urlstyle\endcsname\relax
  \providecommand{\doi}[1]{doi: #1}\else
  \providecommand{\doi}{doi: \begingroup \urlstyle{rm}\Url}\fi

\bibitem[Bae et~al.(2025)Bae, Fisch, Harutyunyan, Ji, Kim, and
  Schuster]{bae2025relaxed}
Bae, S., Fisch, A., Harutyunyan, H., Ji, Z., Kim, S., and Schuster, T.
\newblock Relaxed recursive transformers: Effective parameter sharing with
  layer-wise {LoRA}.
\newblock In \emph{ICLR}, 2025.

\bibitem[Bai et~al.(2019)Bai, Kolter, and Koltun]{bai2019deq}
Bai, S., Kolter, J.~Z., and Koltun, V.
\newblock Deep equilibrium models.
\newblock In \emph{NeurIPS}, 2019.

\bibitem[Balef et~al.(2026)Balef, Koshil, and
  Eggensperger]{rezaeibalef2026onelayer}
Balef, A.~R., Koshil, M., and Eggensperger, K.
\newblock Is one layer enough? {U}nderstanding inference dynamics in tabular
  foundation models.
\newblock In \emph{ICML}, 2026.

\bibitem[Banino et~al.(2021)Banino, Balaguer, and
  Blundell]{banino2021pondernet}
Banino, A., Balaguer, J., and Blundell, C.
\newblock {PonderNet}: Learning to ponder.
\newblock \emph{arXiv preprint arXiv:2107.05407}, 2021.

\bibitem[Brier(1950)]{brier1950verification}
Brier, G.~W.
\newblock Verification of forecasts expressed in terms of probability.
\newblock \emph{Monthly Weather Review}, 78\penalty0 (1):\penalty0 1--3, 1950.

\bibitem[Butcher(2016)]{butcher2016numerical}
Butcher, J.~C.
\newblock \emph{Numerical Methods for Ordinary Differential Equations}.
\newblock Wiley, 3rd edition, 2016.

\bibitem[Chen et~al.(2026)Chen, Li, Liang, Lao, and Liu]{chen2026trainingfree}
Chen, L., Li, J., Liang, C., Lao, N., and Liu, Q.
\newblock Training-free looped transformers.
\newblock \emph{arXiv preprint arXiv:2605.23872}, 2026.

\bibitem[Chen et~al.(2025)Chen, Zhang, and Tao]{justlogic}
Chen, M.~K., Zhang, X., and Tao, D.
\newblock {JustLogic}: A comprehensive benchmark for evaluating deductive
  reasoning in large language models.
\newblock \emph{arXiv preprint arXiv:2501.14851}, 2025.

\bibitem[Chen et~al.(2018)Chen, Rubanova, Bettencourt, and
  Duvenaud]{chen2018neural}
Chen, R. T.~Q., Rubanova, Y., Bettencourt, J., and Duvenaud, D.
\newblock Neural ordinary differential equations.
\newblock In \emph{NeurIPS}, 2018.

\bibitem[Chung et~al.(2026)]{chung2026jevtom}
Chung, K. et~al.
\newblock {Jev-ToM}: Evaluating theory of mind in typed decision models.
\newblock \emph{arXiv preprint arXiv:2610.18610}, 2026.

\bibitem[Dehghani et~al.(2019)Dehghani, Gouws, Vinyals, Uszkoreit, and
  Kaiser]{dehghani2019universal}
Dehghani, M., Gouws, S., Vinyals, O., Uszkoreit, J., and Kaiser, {\L}.
\newblock Universal transformers.
\newblock In \emph{ICLR}, 2019.

\bibitem[Deu{\ss}er et~al.(2026)Deu{\ss}er, Ramamurthy, Sparrenberg, and
  Sifa]{deusser2026jev}
Deu{\ss}er, T., Ramamurthy, R., Sparrenberg, L., and Sifa, R.
\newblock Evaluating and benchmarking the {System One} model {Jev}.
\newblock \emph{arXiv preprint arXiv:2609.37647}, 2026.

\bibitem[Elbayad et~al.(2020)Elbayad, Gu, Grave, and Auli]{elbayad2020depth}
Elbayad, M., Gu, J., Grave, E., and Auli, M.
\newblock Depth-adaptive transformer.
\newblock In \emph{ICLR}, 2020.

\bibitem[Frey et~al.(2026)Frey, Shomali, Bashir, Berghaus, Koehler, and
  Ali]{frey2026adaptive}
Frey, M., Shomali, B., Bashir, A.~H., Berghaus, D., Koehler, J., and Ali, M.
\newblock Adaptive loops and memory in transformers: Think harder or know more?
\newblock In \emph{Latent \& Implicit Thinking Workshop at ICLR}, 2026.
\newblock arXiv:2603.08391.

\bibitem[Gan et~al.(2026)Gan, Lee, and Kang]{gan2026sansi}
Gan, S., Lee, Y.-J., and Kang, D.
\newblock {SanSi}: A looped typed decision model for {System~1.5} thinking.
\newblock \emph{arXiv preprint arXiv:2610.07730}, 2026.

\bibitem[Geiping et~al.(2025)Geiping, McLeish, Jain, Kirchenbauer, Singh,
  Bartoldson, Kailkhura, Bhatele, and Goldstein]{geiping2025scaling}
Geiping, J., McLeish, S., Jain, N., Kirchenbauer, J., Singh, S., Bartoldson,
  B.~R., Kailkhura, B., Bhatele, A., and Goldstein, T.
\newblock Scaling up test-time compute with latent reasoning: A recurrent depth
  approach.
\newblock In \emph{NeurIPS}, 2025.

\bibitem[Giannou et~al.(2023)Giannou, Rajput, Sohn, Lee, Lee, and
  Papailiopoulos]{giannou2023looped}
Giannou, A., Rajput, S., Sohn, J., Lee, K., Lee, J.~D., and Papailiopoulos, D.
\newblock Looped transformers as programmable computers.
\newblock In \emph{ICML}, 2023.

\bibitem[Grabisch \& Roubens(1999)Grabisch and Roubens]{grabisch1999axiomatic}
Grabisch, M. and Roubens, M.
\newblock An axiomatic approach to the concept of interaction among players in
  cooperative games.
\newblock \emph{International Journal of Game Theory}, 28\penalty0
  (4):\penalty0 547--565, 1999.

\bibitem[Graves(2016)]{graves2016adaptive}
Graves, A.
\newblock Adaptive computation time for recurrent neural networks.
\newblock \emph{arXiv preprint arXiv:1603.08983}, 2016.

\bibitem[Haber \& Ruthotto(2018)Haber and Ruthotto]{haber2017stable}
Haber, E. and Ruthotto, L.
\newblock Stable architectures for deep neural networks.
\newblock \emph{Inverse Problems}, 34\penalty0 (1):\penalty0 014004, 2018.

\bibitem[Hairer et~al.(1993)Hairer, N{\o}rsett, and Wanner]{hairer1993solving}
Hairer, E., N{\o}rsett, S.~P., and Wanner, G.
\newblock \emph{Solving Ordinary Differential Equations {I}: Nonstiff
  Problems}.
\newblock Springer, 2nd edition, 1993.

\bibitem[Huang et~al.(2026{\natexlab{a}})Huang, Geng, and Kolter]{huang2026eqr}
Huang, B., Geng, Z., and Kolter, J.~Z.
\newblock Equilibrium reasoners: Learning attractors enables scalable
  reasoning.
\newblock In \emph{ICML}, 2026{\natexlab{a}}.

\bibitem[Huang et~al.(2026{\natexlab{b}})]{huang2026topology}
Huang, B. et~al.
\newblock Towards looped models done right, part {I}: Topology, input
  injection, recurrent-state design.
\newblock \emph{arXiv preprint arXiv:2610.23036}, 2026{\natexlab{b}}.

\bibitem[Iserles(2009)]{iserles2009first}
Iserles, A.
\newblock \emph{A First Course in the Numerical Analysis of Differential
  Equations}.
\newblock Cambridge University Press, 2nd edition, 2009.

\bibitem[Janzing et~al.(2020)Janzing, Minorics, and
  Bl{\"o}baum]{janzing2020feature}
Janzing, D., Minorics, L., and Bl{\"o}baum, P.
\newblock Feature relevance quantification in explainable {AI}: A causal
  problem.
\newblock In \emph{AISTATS}, 2020.

\bibitem[Jeddi et~al.(2026)Jeddi, Ciccone, and Taati]{jeddi2026loopformer}
Jeddi, A., Ciccone, M., and Taati, B.
\newblock {LoopFormer}: Elastic-depth looped transformers for latent reasoning
  via shortcut modulation.
\newblock In \emph{ICLR}, 2026.

\bibitem[Jolicoeur-Martineau(2025)]{jolicoeur2025trm}
Jolicoeur-Martineau, A.
\newblock Less is more: Recursive reasoning with tiny networks.
\newblock \emph{arXiv preprint arXiv:2510.04871}, 2025.

\bibitem[Kazemi et~al.(2023)]{boardgameqa}
Kazemi, M. et~al.
\newblock {BoardgameQA}: A dataset for natural language reasoning with
  contradictory information.
\newblock In \emph{NeurIPS Datasets and Benchmarks Track}, 2023.

\bibitem[Kim et~al.(2024)Kim, Kim, Park, Lee, Song, Kim, Kim, Kim, Lee, Kim,
  Ahn, Yang, Lee, Park, Gim, Cha, Lee, and Kim]{kim2024solar}
Kim, S., Kim, D., Park, C., Lee, W., Song, W., Kim, Y., Kim, H., Kim, Y., Lee,
  H., Kim, J., Ahn, C., Yang, S., Lee, S., Park, H., Gim, G., Cha, M., Lee, H.,
  and Kim, S.
\newblock {SOLAR} 10.7{B}: Scaling large language models with simple yet
  effective depth up-scaling.
\newblock In \emph{NAACL (Industry Track)}, pp.\  23--35, 2024.

\bibitem[Ladhak et~al.(2026)]{ladhak2026elastic}
Ladhak, F. et~al.
\newblock One model to loop them all: Elastic latent recurrence for language
  models.
\newblock \emph{arXiv preprint arXiv:2610.12582}, 2026.

\bibitem[Liu et~al.(2023)]{logiqa2}
Liu, H. et~al.
\newblock {LogiQA} 2.0: An improved dataset for logical reasoning in natural
  language understanding.
\newblock \emph{IEEE/ACM Transactions on Audio, Speech, and Language
  Processing}, 31:\penalty0 2947--2962, 2023.

\bibitem[Lu et~al.(2018)Lu, Zhong, Li, and Dong]{lu2018beyond}
Lu, Y., Zhong, A., Li, Q., and Dong, B.
\newblock Beyond finite layer neural networks: Bridging deep architectures and
  numerical differential equations.
\newblock In \emph{ICML}, 2018.

\bibitem[Lundberg \& Lee(2017)Lundberg and Lee]{lundberg2017unified}
Lundberg, S.~M. and Lee, S.-I.
\newblock A unified approach to interpreting model predictions.
\newblock In \emph{NeurIPS}, 2017.

\bibitem[McLeish et~al.(2025)McLeish, Li, Kirchenbauer, Kalra, Bartoldson,
  Kailkhura, Schwarzschild, Geiping, Goldstein, and
  Goldblum]{mcleish2025retrofitted}
McLeish, S., Li, A., Kirchenbauer, J., Kalra, D.~S., Bartoldson, B.~R.,
  Kailkhura, B., Schwarzschild, A., Geiping, J., Goldstein, T., and Goldblum,
  M.
\newblock Teaching pretrained language models to think deeper with retrofitted
  recurrence.
\newblock \emph{arXiv preprint arXiv:2511.07384}, 2025.

\bibitem[Mirzaee \& Kordjamshidi(2022)Mirzaee and Kordjamshidi]{resq}
Mirzaee, R. and Kordjamshidi, P.
\newblock Transfer learning with synthetic corpora for spatial role labeling
  and reasoning.
\newblock In \emph{EMNLP}, pp.\  6148--6165, 2022.

\bibitem[Movahedi et~al.(2026)Movahedi, Milovanovi{\'c}, Libo~Feigin, Theus,
  Hofmann, Boeva, Rusch, and Orvieto]{movahedi2026fprm}
Movahedi, S., Milovanovi{\'c}, V., Libo~Feigin, S., Theus, A., Hofmann, T.,
  Boeva, V., Rusch, T.~K., and Orvieto, A.
\newblock Fixed-point reasoners: Stable and adaptive deep looped transformers.
\newblock \emph{arXiv preprint arXiv:2606.18206}, 2026.

\bibitem[Palmer(2026)]{kev}
Palmer, J.
\newblock {Kev}: Small {Jev}-like decision models you can train and run
  yourself.
\newblock \url{https://github.com/jaredpalmer/kev}, 2026.
\newblock Apache-2.0.

\bibitem[Panisa(2026)]{von}
Panisa, V.~H.
\newblock {Von}: An open-source {System One} decision model.
\newblock \url{https://github.com/wfzyx/von}, 2026.
\newblock Apache-2.0.

\bibitem[Perez et~al.(2018)Perez, Strub, de~Vries, Dumoulin, and
  Courville]{perez2018film}
Perez, E., Strub, F., de~Vries, H., Dumoulin, V., and Courville, A.
\newblock {FiLM}: Visual reasoning with a general conditioning layer.
\newblock In \emph{AAAI}, 2018.

\bibitem[Qi et~al.(2025)]{proverqa}
Qi, C. et~al.
\newblock Large language models meet symbolic provers for logical reasoning
  evaluation.
\newblock In \emph{ICLR}, 2025.

\bibitem[Queiruga et~al.(2020)Queiruga, Erichson, Taylor, and
  Mahoney]{queiruga2020continuous}
Queiruga, A.~F., Erichson, N.~B., Taylor, D., and Mahoney, M.~W.
\newblock Continuous-in-depth neural networks.
\newblock \emph{arXiv preprint arXiv:2008.02389}, 2020.

\bibitem[Saunshi et~al.(2025)Saunshi, Dikkala, Li, Kumar, and
  Reddi]{saunshi2025reasoning}
Saunshi, N., Dikkala, N., Li, Z., Kumar, S., and Reddi, S.~J.
\newblock Reasoning with latent thoughts: On the power of looped transformers.
\newblock In \emph{ICLR}, 2025.

\bibitem[Schuster et~al.(2022)Schuster, Fisch, Gupta, Dehghani, Bahri, Tran,
  Tay, and Metzler]{schuster2022calm}
Schuster, T., Fisch, A., Gupta, J., Dehghani, M., Bahri, D., Tran, V.~Q., Tay,
  Y., and Metzler, D.
\newblock Confident adaptive language modeling.
\newblock In \emph{NeurIPS}, 2022.

\bibitem[Shapley(1953)]{shapley1953value}
Shapley, L.~S.
\newblock A value for $n$-person games.
\newblock In Kuhn, H.~W. and Tucker, A.~W. (eds.), \emph{Contributions to the
  Theory of Games {II}}, pp.\  307--317. Princeton University Press, 1953.

\bibitem[Shi et~al.(2022)Shi, Zhang, and Lipani]{shi2022stepgame}
Shi, Z., Zhang, Q., and Lipani, A.
\newblock {StepGame}: A new benchmark for robust multi-hop spatial reasoning in
  texts.
\newblock In \emph{AAAI}, pp.\  11321--11329, 2022.

\bibitem[Shomali et~al.(2026)Shomali, Frey, Berghaus, Koehler, and
  Ali]{shomali2026loopmtp}
Shomali, B., Frey, M., Berghaus, D., Koehler, J., and Ali, M.
\newblock {LoopMTP}: A looped transformer guided by latent multi-token
  prediction.
\newblock \emph{arXiv preprint arXiv:2608.03624}, 2026.

\bibitem[Snell et~al.(2025)Snell, Lee, Xu, and Kumar]{snell2024scaling}
Snell, C., Lee, J., Xu, K., and Kumar, A.
\newblock Scaling {LLM} test-time compute optimally can be more effective than
  scaling parameters for reasoning.
\newblock In \emph{ICLR}, 2025.

\bibitem[Song et~al.(2026)Song, Yan, Zhang, and Sun]{song2026lev}
Song, D., Yan, Z., Zhang, Y., and Sun, L.
\newblock {LEV}: Thinking fast and deep with looped decision models.
\newblock \emph{arXiv preprint arXiv:2610.20951}, 2026.

\bibitem[Sprague et~al.(2024)Sprague, Ye, Bostrom, Chaudhuri, and
  Durrett]{musr}
Sprague, Z., Ye, X., Bostrom, K., Chaudhuri, S., and Durrett, G.
\newblock {MuSR}: Testing the limits of chain-of-thought with multistep soft
  reasoning.
\newblock In \emph{ICLR}, 2024.

\bibitem[Tafjord et~al.(2021)Tafjord, Dalvi, and Clark]{tafjord2021proofwriter}
Tafjord, O., Dalvi, B., and Clark, P.
\newblock {ProofWriter}: Generating implications, proofs, and abductive
  statements over natural language.
\newblock In \emph{Findings of ACL}, 2021.

\bibitem[Tsuzuku et~al.(2018)Tsuzuku, Sato, and Sugiyama]{tsuzuku2018lipschitz}
Tsuzuku, Y., Sato, I., and Sugiyama, M.
\newblock {Lipschitz}-margin training: Scalable certification of perturbation
  invariance for deep neural networks.
\newblock In \emph{NeurIPS}, 2018.

\bibitem[Wang et~al.(2025)Wang, Li, Sun, Chen, Liu, Wu, Lu, Song, and
  Abbasi~Yadkori]{wang2025hrm}
Wang, G., Li, J., Sun, Y., Chen, X., Liu, C., Wu, Y., Lu, M., Song, S., and
  Abbasi~Yadkori, Y.
\newblock Hierarchical reasoning model.
\newblock \emph{arXiv preprint arXiv:2506.21734}, 2025.

\bibitem[Wang et~al.(2026)]{wang2026recurtrace}
Wang, Y. et~al.
\newblock {RecurTrace}: Adaptive latent reasoning with loop-time memory.
\newblock \emph{arXiv preprint arXiv:2609.03379}, 2026.

\bibitem[Warner et~al.(2025)]{warner2025modernbert}
Warner, B. et~al.
\newblock Smarter, better, faster, longer: A modern bidirectional encoder for
  fast, memory efficient, and long context finetuning and inference.
\newblock In \emph{ACL}, pp.\  2526--2547, 2025.

\bibitem[Wei et~al.(2022)Wei, Wang, Schuurmans, Bosma, Ichter, Xia, Chi, Le,
  and Zhou]{wei2022chain}
Wei, J., Wang, X., Schuurmans, D., Bosma, M., Ichter, B., Xia, F., Chi, E.~H.,
  Le, Q.~V., and Zhou, D.
\newblock Chain-of-thought prompting elicits reasoning in large language
  models.
\newblock In \emph{NeurIPS}, 2022.

\bibitem[Weller et~al.(2026)]{weller2025ettin}
Weller, O. et~al.
\newblock {Seq vs Seq}: An open suite of paired encoders and decoders.
\newblock In \emph{ICLR}, 2026.
\newblock arXiv:2507.11412.

\bibitem[Wu et~al.(2024)]{wu2024llamapro}
Wu, C. et~al.
\newblock {LLaMA Pro}: Progressive {LLaMA} with block expansion.
\newblock In \emph{ACL}, pp.\  6518--6537, 2024.

\bibitem[Xin et~al.(2020)Xin, Tang, Lee, Yu, and Lin]{xin2020deebert}
Xin, J., Tang, R., Lee, J., Yu, Y., and Lin, J.
\newblock {DeeBERT}: Dynamic early exiting for accelerating {BERT} inference.
\newblock In \emph{ACL}, 2020.

\bibitem[Yang et~al.(2025)]{yang2025qwen3}
Yang, A. et~al.
\newblock {Qwen3} technical report.
\newblock \emph{arXiv preprint arXiv:2505.09388}, 2025.

\bibitem[Yang et~al.(2024)Yang, Lee, Nowak, and Papailiopoulos]{yang2024looped}
Yang, L., Lee, K., Nowak, R., and Papailiopoulos, D.
\newblock Looped transformers are better at learning learning algorithms.
\newblock In \emph{ICLR}, 2024.

\bibitem[Yu et~al.(2020)Yu, Jiang, Dong, and Feng]{reclor}
Yu, W., Jiang, Z., Dong, Y., and Feng, J.
\newblock {ReClor}: A reading comprehension dataset requiring logical
  reasoning.
\newblock In \emph{ICLR}, 2020.

\bibitem[Zela et~al.(2026)Zela, Maranjyan, Puigdemont, Berghaus, Boreiko, and
  Cevher]{zela2026sfprm}
Zela, A., Maranjyan, A., Puigdemont, P., Berghaus, D., Boreiko, V., and Cevher,
  V.
\newblock Spherical fixed-point reasoners: Adaptive loop halting on the
  hypersphere.
\newblock \emph{arXiv preprint arXiv:2610.25646}, 2026.

\bibitem[Zhou et~al.(2020)Zhou, Xu, Ge, McAuley, Xu, and Wei]{zhou2020pabee}
Zhou, W., Xu, C., Ge, T., McAuley, J., Xu, K., and Wei, F.
\newblock {BERT} loses patience: Fast and robust inference with early exit.
\newblock In \emph{NeurIPS}, 2020.

\bibitem[Zhu et~al.(2025)]{zhu2025ouro}
Zhu, R.-J. et~al.
\newblock Scaling latent reasoning via looped language models.
\newblock \emph{arXiv preprint arXiv:2510.25741}, 2025.

\end{thebibliography}
\bibliographystyle{icml2026}

\newpage
\appendix
\onecolumn
\FloatBarrier
\section{Per-dataset results}\label{app:perset}
Table~\ref{tab:main_full} gives every budget of the loop-trained models, and Tables~\ref{tab:perset}, \ref{tab:rbench} and~\ref{tab:calib} give the per-dataset results behind Table~\ref{tab:main}. On three knowledge-heavy multiple-choice benchmarks outside our test groups, evaluated zero-shot (MMLU, ARC-Easy and OpenBookQA; MMLU is also a source of Kev transfer), four passes change mean accuracy by $-0.4$ (Qwen3-0.6B) to $+1.0$ points (Ettin-400M) relative to the single pass.

\begin{table}[!htbp]
\centering
\footnotesize\setlength{\tabcolsep}{4pt}
\caption{Table~\ref{tab:main} with every budget of the loop-trained checkpoint, rows ordered by compute: accuracy (\%) and Brier score (lower is better), mean $\pm$ standard deviation over three seeds. Groups, variants and the Layers column as in Table~\ref{tab:main}. Bold: best per model and column.}
\label{tab:main_full}
\fitbox{\begin{tabular}{llrrrrrrr}
\toprule
 & & \multicolumn{2}{c}{In-distribution} & \multicolumn{2}{c}{Longer reasoning} & \multicolumn{2}{c}{Out-of-distribution} & \\
\cmidrule(lr){3-4}\cmidrule(lr){5-6}\cmidrule(lr){7-8}
Model & Variant & Acc. & Brier & Acc. & Brier & Acc. & Brier & Layers \\
\midrule
 & Single pass & 92.8\,{\scriptsize$\pm$0.3} & 0.101\,{\scriptsize$\pm$0.004} & 70.6\,{\scriptsize$\pm$1.3} & 0.404\,{\scriptsize$\pm$0.025} & 50.9\,{\scriptsize$\pm$0.2} & 0.728\,{\scriptsize$\pm$0.005} & 28  \\
\rowcolor{reloadbg}\cellcolor{white} & RELOAD ($K{=}1$) & 93.0\,{\scriptsize$\pm$0.4} & 0.096\,{\scriptsize$\pm$0.003} & 68.6\,{\scriptsize$\pm$1.6} & 0.437\,{\scriptsize$\pm$0.029} & 52.4\,{\scriptsize$\pm$1.2} & 0.701\,{\scriptsize$\pm$0.015} & 28  \\
\rowcolor{reloadbg}\cellcolor{white} & RELOAD ($K{=}2$) & \textbf{93.9}\,{\scriptsize$\pm$0.1} & 0.086\,{\scriptsize$\pm$0.002} & 80.7\,{\scriptsize$\pm$0.7} & 0.276\,{\scriptsize$\pm$0.024} & 54.0\,{\scriptsize$\pm$0.7} & 0.677\,{\scriptsize$\pm$0.007} & 37  \\
\rowcolor{reloadbg}\cellcolor{white} & RELOAD + controller & 93.6\,{\scriptsize$\pm$0.2} & 0.088\,{\scriptsize$\pm$0.002} & 85.4\,{\scriptsize$\pm$1.0} & 0.203\,{\scriptsize$\pm$0.014} & 53.8\,{\scriptsize$\pm$1.0} & 0.682\,{\scriptsize$\pm$0.008} & 35/53/39  \\
\rowcolor{reloadbg}\cellcolor{white} & RELOAD ($K{=}3$) & \textbf{93.9}\,{\scriptsize$\pm$0.2} & \textbf{0.085}\,{\scriptsize$\pm$0.002} & 84.1\,{\scriptsize$\pm$0.8} & 0.223\,{\scriptsize$\pm$0.016} & 54.0\,{\scriptsize$\pm$0.8} & \textbf{0.675}\,{\scriptsize$\pm$0.007} & 46  \\
\rowcolor{reloadbg}\cellcolor{white}\multirow{-6}{*}{Von (395M)}& RELOAD ($K{=}4$) & \textbf{93.9}\,{\scriptsize$\pm$0.2} & \textbf{0.085}\,{\scriptsize$\pm$0.002} & \textbf{85.7}\,{\scriptsize$\pm$0.4} & \textbf{0.198}\,{\scriptsize$\pm$0.009} & \textbf{54.1}\,{\scriptsize$\pm$0.7} & \textbf{0.675}\,{\scriptsize$\pm$0.007} & 55  \\
\midrule
 & Single pass & 93.0\,{\scriptsize$\pm$0.4} & 0.100\,{\scriptsize$\pm$0.003} & 72.3\,{\scriptsize$\pm$0.6} & 0.384\,{\scriptsize$\pm$0.008} & 47.9\,{\scriptsize$\pm$0.2} & 0.714\,{\scriptsize$\pm$0.015} & 28  \\
\rowcolor{reloadbg}\cellcolor{white} & RELOAD ($K{=}1$) & 93.3\,{\scriptsize$\pm$0.2} & 0.095\,{\scriptsize$\pm$0.003} & 70.4\,{\scriptsize$\pm$1.2} & 0.407\,{\scriptsize$\pm$0.016} & 48.4\,{\scriptsize$\pm$0.5} & 0.723\,{\scriptsize$\pm$0.017} & 28  \\
\rowcolor{reloadbg}\cellcolor{white} & RELOAD ($K{=}2$) & \textbf{94.0}\,{\scriptsize$\pm$0.1} & \textbf{0.085}\,{\scriptsize$\pm$0.001} & 82.6\,{\scriptsize$\pm$1.9} & 0.250\,{\scriptsize$\pm$0.032} & 50.2\,{\scriptsize$\pm$0.7} & 0.697\,{\scriptsize$\pm$0.016} & 37  \\
\rowcolor{reloadbg}\cellcolor{white} & RELOAD + controller & 93.9\,{\scriptsize$\pm$0.2} & 0.087\,{\scriptsize$\pm$0.001} & \textbf{87.0}\,{\scriptsize$\pm$1.7} & 0.183\,{\scriptsize$\pm$0.027} & 50.0\,{\scriptsize$\pm$0.6} & 0.696\,{\scriptsize$\pm$0.013} & 35/52/37  \\
\rowcolor{reloadbg}\cellcolor{white} & RELOAD ($K{=}3$) & \textbf{94.0}\,{\scriptsize$\pm$0.2} & \textbf{0.085}\,{\scriptsize$\pm$0.000} & 86.5\,{\scriptsize$\pm$1.9} & 0.194\,{\scriptsize$\pm$0.031} & 50.4\,{\scriptsize$\pm$0.9} & 0.689\,{\scriptsize$\pm$0.017} & 46  \\
\rowcolor{reloadbg}\cellcolor{white}\multirow{-6}{*}{Ettin-400M}& RELOAD ($K{=}4$) & \textbf{94.0}\,{\scriptsize$\pm$0.2} & \textbf{0.085}\,{\scriptsize$\pm$0.000} & \textbf{87.0}\,{\scriptsize$\pm$1.8} & \textbf{0.179}\,{\scriptsize$\pm$0.029} & \textbf{50.5}\,{\scriptsize$\pm$0.8} & \textbf{0.686}\,{\scriptsize$\pm$0.017} & 55  \\
\midrule
 & Single pass & 93.9\,{\scriptsize$\pm$0.3} & 0.089\,{\scriptsize$\pm$0.003} & 80.2\,{\scriptsize$\pm$1.4} & 0.292\,{\scriptsize$\pm$0.013} & 48.7\,{\scriptsize$\pm$0.7} & 0.717\,{\scriptsize$\pm$0.028} & 28  \\
\rowcolor{reloadbg}\cellcolor{white} & RELOAD ($K{=}1$) & 94.3\,{\scriptsize$\pm$0.0} & 0.083\,{\scriptsize$\pm$0.002} & 78.4\,{\scriptsize$\pm$0.3} & 0.325\,{\scriptsize$\pm$0.016} & 51.8\,{\scriptsize$\pm$0.6} & 0.663\,{\scriptsize$\pm$0.011} & 28  \\
\rowcolor{reloadbg}\cellcolor{white} & RELOAD ($K{=}2$) & \textbf{94.6}\,{\scriptsize$\pm$0.1} & \textbf{0.078}\,{\scriptsize$\pm$0.002} & 86.3\,{\scriptsize$\pm$1.4} & 0.198\,{\scriptsize$\pm$0.009} & 52.7\,{\scriptsize$\pm$0.7} & 0.654\,{\scriptsize$\pm$0.009} & 37  \\
\rowcolor{reloadbg}\cellcolor{white} & RELOAD + controller & 94.5\,{\scriptsize$\pm$0.1} & 0.079\,{\scriptsize$\pm$0.002} & 90.6\,{\scriptsize$\pm$0.7} & 0.135\,{\scriptsize$\pm$0.005} & 52.6\,{\scriptsize$\pm$0.9} & 0.655\,{\scriptsize$\pm$0.009} & 34/54/39  \\
\rowcolor{reloadbg}\cellcolor{white} & RELOAD ($K{=}3$) & \textbf{94.6}\,{\scriptsize$\pm$0.1} & \textbf{0.078}\,{\scriptsize$\pm$0.002} & 89.4\,{\scriptsize$\pm$0.9} & 0.150\,{\scriptsize$\pm$0.007} & 53.0\,{\scriptsize$\pm$0.8} & 0.650\,{\scriptsize$\pm$0.009} & 46  \\
\rowcolor{reloadbg}\cellcolor{white}\multirow{-6}{*}{Ettin-1B}& RELOAD ($K{=}4$) & \textbf{94.6}\,{\scriptsize$\pm$0.2} & \textbf{0.078}\,{\scriptsize$\pm$0.003} & \textbf{90.7}\,{\scriptsize$\pm$0.8} & \textbf{0.134}\,{\scriptsize$\pm$0.005} & \textbf{53.2}\,{\scriptsize$\pm$0.9} & \textbf{0.649}\,{\scriptsize$\pm$0.010} & 55  \\
\midrule
 & Single pass & 86.1\,{\scriptsize$\pm$0.4} & 0.194\,{\scriptsize$\pm$0.009} & 48.1\,{\scriptsize$\pm$0.8} & 0.629\,{\scriptsize$\pm$0.015} & 52.6\,{\scriptsize$\pm$0.7} & 0.642\,{\scriptsize$\pm$0.009} & 28  \\
\rowcolor{reloadbg}\cellcolor{white} & RELOAD ($K{=}1$) & 86.3\,{\scriptsize$\pm$0.1} & 0.190\,{\scriptsize$\pm$0.000} & 49.1\,{\scriptsize$\pm$1.5} & 0.626\,{\scriptsize$\pm$0.014} & 52.1\,{\scriptsize$\pm$0.9} & 0.645\,{\scriptsize$\pm$0.017} & 28  \\
\rowcolor{reloadbg}\cellcolor{white} & RELOAD ($K{=}2$) & 87.8\,{\scriptsize$\pm$0.2} & 0.171\,{\scriptsize$\pm$0.003} & 53.6\,{\scriptsize$\pm$0.6} & 0.577\,{\scriptsize$\pm$0.008} & 53.3\,{\scriptsize$\pm$1.0} & \textbf{0.637}\,{\scriptsize$\pm$0.015} & 38  \\
\rowcolor{reloadbg}\cellcolor{white} & RELOAD + controller & 87.9\,{\scriptsize$\pm$0.2} & 0.171\,{\scriptsize$\pm$0.003} & 54.3\,{\scriptsize$\pm$1.0} & 0.571\,{\scriptsize$\pm$0.013} & 52.8\,{\scriptsize$\pm$0.8} & 0.641\,{\scriptsize$\pm$0.018} & 37/44/41  \\
\rowcolor{reloadbg}\cellcolor{white} & RELOAD ($K{=}3$) & \textbf{88.0}\,{\scriptsize$\pm$0.2} & \textbf{0.169}\,{\scriptsize$\pm$0.003} & 54.6\,{\scriptsize$\pm$0.9} & 0.565\,{\scriptsize$\pm$0.010} & \textbf{53.4}\,{\scriptsize$\pm$0.9} & 0.639\,{\scriptsize$\pm$0.016} & 48  \\
\rowcolor{reloadbg}\cellcolor{white}\multirow{-6}{*}{Qwen3-0.6B}& RELOAD ($K{=}4$) & \textbf{88.0}\,{\scriptsize$\pm$0.2} & \textbf{0.169}\,{\scriptsize$\pm$0.003} & \textbf{55.2}\,{\scriptsize$\pm$0.6} & \textbf{0.561}\,{\scriptsize$\pm$0.009} & \textbf{53.4}\,{\scriptsize$\pm$1.0} & 0.642\,{\scriptsize$\pm$0.016} & 58  \\
\midrule
 & Single pass & 89.3\,{\scriptsize$\pm$0.5} & 0.153\,{\scriptsize$\pm$0.009} & 57.7\,{\scriptsize$\pm$1.4} & 0.532\,{\scriptsize$\pm$0.015} & 54.6\,{\scriptsize$\pm$1.4} & 0.630\,{\scriptsize$\pm$0.022} & 28  \\
\rowcolor{reloadbg}\cellcolor{white} & RELOAD ($K{=}1$) & 89.4\,{\scriptsize$\pm$0.4} & 0.152\,{\scriptsize$\pm$0.005} & 56.2\,{\scriptsize$\pm$1.3} & 0.563\,{\scriptsize$\pm$0.020} & 55.0\,{\scriptsize$\pm$1.0} & 0.628\,{\scriptsize$\pm$0.013} & 28  \\
\rowcolor{reloadbg}\cellcolor{white} & RELOAD ($K{=}2$) & 90.5\,{\scriptsize$\pm$0.3} & 0.136\,{\scriptsize$\pm$0.004} & 59.4\,{\scriptsize$\pm$1.0} & 0.522\,{\scriptsize$\pm$0.019} & 56.2\,{\scriptsize$\pm$1.0} & 0.612\,{\scriptsize$\pm$0.013} & 38  \\
\rowcolor{reloadbg}\cellcolor{white} & RELOAD + controller & 90.6\,{\scriptsize$\pm$0.5} & 0.135\,{\scriptsize$\pm$0.005} & 59.9\,{\scriptsize$\pm$1.1} & 0.511\,{\scriptsize$\pm$0.018} & 56.2\,{\scriptsize$\pm$1.1} & 0.610\,{\scriptsize$\pm$0.012} & 38/51/43  \\
\rowcolor{reloadbg}\cellcolor{white} & RELOAD ($K{=}3$) & 90.8\,{\scriptsize$\pm$0.3} & 0.132\,{\scriptsize$\pm$0.003} & 59.9\,{\scriptsize$\pm$1.5} & 0.510\,{\scriptsize$\pm$0.017} & \textbf{56.6}\,{\scriptsize$\pm$0.9} & 0.606\,{\scriptsize$\pm$0.010} & 48  \\
\rowcolor{reloadbg}\cellcolor{white}\multirow{-6}{*}{Qwen3-1.7B}& RELOAD ($K{=}4$) & \textbf{90.9}\,{\scriptsize$\pm$0.2} & \textbf{0.131}\,{\scriptsize$\pm$0.003} & \textbf{60.2}\,{\scriptsize$\pm$1.4} & \textbf{0.506}\,{\scriptsize$\pm$0.017} & 56.5\,{\scriptsize$\pm$0.8} & \textbf{0.605}\,{\scriptsize$\pm$0.008} & 58  \\
\midrule
 & Single pass & 90.9\,{\scriptsize$\pm$1.4} & 0.132\,{\scriptsize$\pm$0.014} & 62.0\,{\scriptsize$\pm$4.5} & 0.507\,{\scriptsize$\pm$0.045} & 60.4\,{\scriptsize$\pm$0.9} & 0.578\,{\scriptsize$\pm$0.016} & 36  \\
\rowcolor{reloadbg}\cellcolor{white} & RELOAD ($K{=}1$) & 91.9\,{\scriptsize$\pm$0.3} & 0.121\,{\scriptsize$\pm$0.004} & 64.2\,{\scriptsize$\pm$0.6} & 0.490\,{\scriptsize$\pm$0.011} & 60.5\,{\scriptsize$\pm$0.5} & 0.581\,{\scriptsize$\pm$0.006} & 36  \\
\rowcolor{reloadbg}\cellcolor{white} & RELOAD ($K{=}2$) & 92.9\,{\scriptsize$\pm$0.5} & 0.104\,{\scriptsize$\pm$0.006} & 71.0\,{\scriptsize$\pm$0.2} & 0.414\,{\scriptsize$\pm$0.009} & 61.6\,{\scriptsize$\pm$0.4} & 0.570\,{\scriptsize$\pm$0.006} & 48  \\
\rowcolor{reloadbg}\cellcolor{white} & RELOAD + controller & 92.8\,{\scriptsize$\pm$0.5} & 0.105\,{\scriptsize$\pm$0.006} & 71.8\,{\scriptsize$\pm$0.2} & 0.398\,{\scriptsize$\pm$0.009} & 61.5\,{\scriptsize$\pm$0.1} & 0.570\,{\scriptsize$\pm$0.007} & 46/62/51  \\
\rowcolor{reloadbg}\cellcolor{white} & RELOAD ($K{=}3$) & 92.9\,{\scriptsize$\pm$0.5} & 0.103\,{\scriptsize$\pm$0.006} & 71.8\,{\scriptsize$\pm$0.2} & 0.397\,{\scriptsize$\pm$0.009} & 61.9\,{\scriptsize$\pm$0.3} & \textbf{0.569}\,{\scriptsize$\pm$0.006} & 60  \\
\rowcolor{reloadbg}\cellcolor{white}\multirow{-6}{*}{Qwen3-4B}& RELOAD ($K{=}4$) & \textbf{93.0}\,{\scriptsize$\pm$0.5} & \textbf{0.102}\,{\scriptsize$\pm$0.007} & \textbf{72.1}\,{\scriptsize$\pm$0.4} & \textbf{0.393}\,{\scriptsize$\pm$0.009} & \textbf{62.0}\,{\scriptsize$\pm$0.2} & \textbf{0.569}\,{\scriptsize$\pm$0.006} & 72  \\
\bottomrule
\end{tabular}}
\end{table}

\begin{table}[!htbp]
\centering
\footnotesize\setlength{\tabcolsep}{4pt}
\caption{Accuracy (\%) per test set and macro Brier score over the four sets (lower is better), mean $\pm$ standard deviation over three seeds. All arms of a model share hyperparameters, data and seeds, so paired intervals in the text are tighter than these spreads. Controller as in Table~\ref{tab:main}. Layers: layer applications per question averaged over the four sets (about two passes for the controller, against about 2.5 over the three groups of Table~\ref{tab:main}). Bold: best per model and column.}
\label{tab:perset}
\fitbox{\begin{tabular}{lrrrrrrr}
\toprule
 & \multicolumn{5}{c}{Accuracy (\%)} & & \\
\cmidrule(lr){2-6}
 & ProofWriter & StepGame & Kev decision & Kev transfer & Macro & Macro Brier $\downarrow$ & Layers \\
\midrule
\textit{Von (ModernBERT-large, 395M)} & & & & & & & \\
\midrule
Single pass & 98.88\,{\scriptsize$\pm$0.11} & 82.23\,{\scriptsize$\pm$0.95} & 85.61\,{\scriptsize$\pm$0.17} & 76.37\,{\scriptsize$\pm$1.76} & 85.77\,{\scriptsize$\pm$0.15} & 0.203\,{\scriptsize$\pm$0.001} & 28  \\
\rowcolor{reloadbg}RELOAD ($K{=}2$) & 99.66\,{\scriptsize$\pm$0.02} & 88.30\,{\scriptsize$\pm$0.35} & \textbf{86.14}\,{\scriptsize$\pm$0.43} & 77.13\,{\scriptsize$\pm$1.33} & 87.81\,{\scriptsize$\pm$0.32} & 0.175\,{\scriptsize$\pm$0.006} & 37  \\
\rowcolor{reloadbg}RELOAD + controller & 99.39\,{\scriptsize$\pm$0.08} & 90.70\,{\scriptsize$\pm$0.56} & 85.50\,{\scriptsize$\pm$0.58} & 77.03\,{\scriptsize$\pm$1.42} & 88.16\,{\scriptsize$\pm$0.38} & 0.167\,{\scriptsize$\pm$0.005} & 37.1  \\
\rowcolor{reloadbg}RELOAD ($K{=}4$) & \textbf{99.71}\,{\scriptsize$\pm$0.02} & \textbf{90.88}\,{\scriptsize$\pm$0.38} & 86.03\,{\scriptsize$\pm$0.85} & \textbf{77.49}\,{\scriptsize$\pm$1.92} & \textbf{88.53}\,{\scriptsize$\pm$0.35} & \textbf{0.163}\,{\scriptsize$\pm$0.006} & 55  \\
\midrule
\textit{Ettin-400M} & & & & & & & \\
\midrule
Single pass & 98.56\,{\scriptsize$\pm$0.23} & 82.90\,{\scriptsize$\pm$0.43} & \textbf{86.92}\,{\scriptsize$\pm$0.22} & 68.29\,{\scriptsize$\pm$0.30} & 84.17\,{\scriptsize$\pm$0.09} & 0.224\,{\scriptsize$\pm$0.001} & 28  \\
\rowcolor{reloadbg}RELOAD ($K{=}2$) & 99.52\,{\scriptsize$\pm$0.07} & 89.13\,{\scriptsize$\pm$1.03} & 86.78\,{\scriptsize$\pm$0.50} & 69.41\,{\scriptsize$\pm$1.60} & 86.21\,{\scriptsize$\pm$0.51} & 0.189\,{\scriptsize$\pm$0.005} & 37  \\
\rowcolor{reloadbg}RELOAD + controller & 99.14\,{\scriptsize$\pm$0.11} & 91.38\,{\scriptsize$\pm$0.80} & 86.72\,{\scriptsize$\pm$0.50} & 69.72\,{\scriptsize$\pm$1.33} & 86.74\,{\scriptsize$\pm$0.28} & 0.184\,{\scriptsize$\pm$0.003} & 37.0  \\
\rowcolor{reloadbg}RELOAD ($K{=}4$) & \textbf{99.56}\,{\scriptsize$\pm$0.05} & \textbf{91.45}\,{\scriptsize$\pm$0.71} & 86.67\,{\scriptsize$\pm$0.43} & \textbf{70.68}\,{\scriptsize$\pm$0.93} & \textbf{87.09}\,{\scriptsize$\pm$0.25} & \textbf{0.179}\,{\scriptsize$\pm$0.005} & 55  \\
\midrule
\textit{Ettin-1B} & & & & & & & \\
\midrule
Single pass & 98.67\,{\scriptsize$\pm$0.11} & 88.15\,{\scriptsize$\pm$0.80} & 87.00\,{\scriptsize$\pm$1.20} & 68.75\,{\scriptsize$\pm$2.12} & 85.64\,{\scriptsize$\pm$0.64} & 0.202\,{\scriptsize$\pm$0.005} & 28  \\
\rowcolor{reloadbg}RELOAD ($K{=}2$) & 99.57\,{\scriptsize$\pm$0.08} & 91.55\,{\scriptsize$\pm$0.77} & \textbf{87.47}\,{\scriptsize$\pm$0.43} & \textbf{73.83}\,{\scriptsize$\pm$0.89} & 88.11\,{\scriptsize$\pm$0.40} & 0.168\,{\scriptsize$\pm$0.002} & 37  \\
\rowcolor{reloadbg}RELOAD + controller & 99.34\,{\scriptsize$\pm$0.14} & 93.73\,{\scriptsize$\pm$0.45} & 87.39\,{\scriptsize$\pm$0.39} & 73.32\,{\scriptsize$\pm$1.40} & 88.45\,{\scriptsize$\pm$0.48} & 0.164\,{\scriptsize$\pm$0.002} & 37.1  \\
\rowcolor{reloadbg}RELOAD ($K{=}4$) & \textbf{99.61}\,{\scriptsize$\pm$0.06} & \textbf{93.77}\,{\scriptsize$\pm$0.50} & 87.31\,{\scriptsize$\pm$0.34} & 73.78\,{\scriptsize$\pm$0.76} & \textbf{88.61}\,{\scriptsize$\pm$0.24} & \textbf{0.159}\,{\scriptsize$\pm$0.001} & 55  \\
\midrule
\textit{Qwen3-0.6B (Kev-style)} & & & & & & & \\
\midrule
Single pass & 90.19\,{\scriptsize$\pm$1.28} & 65.77\,{\scriptsize$\pm$0.49} & 84.61\,{\scriptsize$\pm$0.46} & \textbf{68.50}\,{\scriptsize$\pm$0.93} & 77.27\,{\scriptsize$\pm$0.39} & 0.306\,{\scriptsize$\pm$0.006} & 28  \\
\rowcolor{reloadbg}RELOAD ($K{=}2$) & 93.02\,{\scriptsize$\pm$0.81} & 69.80\,{\scriptsize$\pm$0.58} & 84.44\,{\scriptsize$\pm$0.29} & 64.94\,{\scriptsize$\pm$3.15} & 78.05\,{\scriptsize$\pm$0.92} & 0.296\,{\scriptsize$\pm$0.006} & 38  \\
\rowcolor{reloadbg}RELOAD + controller & 93.10\,{\scriptsize$\pm$0.73} & 70.07\,{\scriptsize$\pm$0.71} & \textbf{84.64}\,{\scriptsize$\pm$0.38} & 64.68\,{\scriptsize$\pm$3.09} & 78.12\,{\scriptsize$\pm$0.87} & 0.295\,{\scriptsize$\pm$0.006} & 37.8  \\
\rowcolor{reloadbg}RELOAD ($K{=}4$) & \textbf{93.42}\,{\scriptsize$\pm$0.62} & \textbf{70.70}\,{\scriptsize$\pm$0.59} & 84.42\,{\scriptsize$\pm$0.22} & 65.09\,{\scriptsize$\pm$2.64} & \textbf{78.41}\,{\scriptsize$\pm$0.61} & \textbf{0.293}\,{\scriptsize$\pm$0.007} & 58  \\
\midrule
\textit{Qwen3-1.7B (Kev-style)} & & & & & & & \\
\midrule
Single pass & 92.81\,{\scriptsize$\pm$0.71} & 73.33\,{\scriptsize$\pm$1.16} & 85.97\,{\scriptsize$\pm$0.55} & 73.58\,{\scriptsize$\pm$2.33} & 81.42\,{\scriptsize$\pm$0.32} & 0.258\,{\scriptsize$\pm$0.007} & 28  \\
\rowcolor{reloadbg}RELOAD ($K{=}2$) & 95.25\,{\scriptsize$\pm$0.28} & 74.78\,{\scriptsize$\pm$0.43} & 86.06\,{\scriptsize$\pm$0.43} & 76.47\,{\scriptsize$\pm$1.46} & 83.14\,{\scriptsize$\pm$0.44} & 0.233\,{\scriptsize$\pm$0.009} & 38  \\
\rowcolor{reloadbg}RELOAD + controller & 95.30\,{\scriptsize$\pm$0.32} & 75.13\,{\scriptsize$\pm$0.58} & 86.00\,{\scriptsize$\pm$0.52} & 76.58\,{\scriptsize$\pm$1.42} & 83.25\,{\scriptsize$\pm$0.59} & 0.232\,{\scriptsize$\pm$0.010} & 38.7  \\
\rowcolor{reloadbg}RELOAD ($K{=}4$) & \textbf{95.67}\,{\scriptsize$\pm$0.22} & \textbf{75.33}\,{\scriptsize$\pm$0.64} & \textbf{86.42}\,{\scriptsize$\pm$0.30} & \textbf{76.73}\,{\scriptsize$\pm$1.44} & \textbf{83.54}\,{\scriptsize$\pm$0.57} & \textbf{0.227}\,{\scriptsize$\pm$0.009} & 58  \\
\midrule
\textit{Qwen3-4B (Kev-style)} & & & & & & & \\
\midrule
Single pass & 95.74\,{\scriptsize$\pm$0.59} & 76.47\,{\scriptsize$\pm$3.83} & 85.94\,{\scriptsize$\pm$0.89} & 80.74\,{\scriptsize$\pm$1.31} & 84.72\,{\scriptsize$\pm$1.20} & 0.217\,{\scriptsize$\pm$0.011} & 36  \\
\rowcolor{reloadbg}RELOAD ($K{=}2$) & 97.49\,{\scriptsize$\pm$0.24} & 82.53\,{\scriptsize$\pm$0.13} & 87.06\,{\scriptsize$\pm$0.77} & 81.25\,{\scriptsize$\pm$0.46} & 87.08\,{\scriptsize$\pm$0.18} & 0.187\,{\scriptsize$\pm$0.004} & 48  \\
\rowcolor{reloadbg}RELOAD + controller & 97.48\,{\scriptsize$\pm$0.25} & 82.95\,{\scriptsize$\pm$0.35} & 86.75\,{\scriptsize$\pm$0.58} & 81.25\,{\scriptsize$\pm$0.30} & 87.11\,{\scriptsize$\pm$0.22} & 0.186\,{\scriptsize$\pm$0.004} & 47.7  \\
\rowcolor{reloadbg}RELOAD ($K{=}4$) & \textbf{97.69}\,{\scriptsize$\pm$0.27} & \textbf{83.13}\,{\scriptsize$\pm$0.34} & \textbf{87.08}\,{\scriptsize$\pm$0.65} & \textbf{81.35}\,{\scriptsize$\pm$0.49} & \textbf{87.31}\,{\scriptsize$\pm$0.29} & \textbf{0.184}\,{\scriptsize$\pm$0.005} & 72  \\
\bottomrule
\end{tabular}}
\end{table}

\begin{table}[!htbp]
\centering
\footnotesize\setlength{\tabcolsep}{4pt}
\caption{Zero-shot accuracy (\%) on the eight reasoning benchmarks never used for training or model selection, and macro Brier score (three seeds). Chance: 33\% for ParaRules, JustLogic, ProverQA and BoardgameQA, 25\% for ReClor and LogiQA~2.0, 50\% for ReSQ and about 40\% for MuSR. Bold: best per model and column.}
\label{tab:rbench}
\fitbox{\begin{tabular}{llrrrrrrrrrr}
\toprule
Model & & ParaRules & JustLogic & ProverQA & BoardgameQA & MuSR & ReClor & LogiQA 2.0 & ReSQ & Macro & Brier $\downarrow$ \\
\midrule
Von & Single pass & 54.3 & 40.3 & 57.9 & 53.4 & 48.8 & \textbf{36.7} & 32.7 & \textbf{57.9} & 47.7 & 0.776  \\
\rowcolor{reloadbg}\cellcolor{white} & RELOAD ($K{=}4$) & \textbf{67.3} & \textbf{45.8} & \textbf{59.1} & \textbf{57.8} & \textbf{51.8} & 36.5 & \textbf{33.6} & 57.7 & \textbf{51.2} & \textbf{0.719}  \\
\midrule
Ettin-400M & Single pass & 49.2 & 41.0 & 48.0 & \textbf{53.0} & 45.8 & 37.3 & 32.6 & 55.6 & 45.3 & 0.748  \\
\rowcolor{reloadbg}\cellcolor{white} & RELOAD ($K{=}4$) & \textbf{57.4} & \textbf{48.5} & \textbf{51.9} & 52.8 & \textbf{46.0} & \textbf{37.9} & \textbf{33.2} & \textbf{56.0} & \textbf{48.0} & \textbf{0.722}  \\
\midrule
Ettin-1B & Single pass & 43.8 & 46.6 & 47.4 & 54.7 & \textbf{47.6} & 39.4 & 33.2 & 56.9 & 46.2 & 0.754  \\
\rowcolor{reloadbg}\cellcolor{white} & RELOAD ($K{=}4$) & \textbf{57.4} & \textbf{50.0} & \textbf{52.9} & \textbf{64.9} & 47.5 & \textbf{40.1} & \textbf{33.7} & \textbf{58.1} & \textbf{50.6} & \textbf{0.685}  \\
\midrule
Qwen3-0.6B & Single pass & \textbf{71.8} & \textbf{55.9} & \textbf{56.0} & \textbf{53.9} & 40.2 & 39.7 & 34.2 & 53.6 & 50.6 & 0.668  \\
\rowcolor{reloadbg}\cellcolor{white} & RELOAD ($K{=}4$) & 70.5 & 53.6 & 54.7 & 53.7 & \textbf{51.0} & \textbf{41.9} & \textbf{35.4} & \textbf{54.7} & \textbf{51.9} & \textbf{0.662}  \\
\midrule
Qwen3-1.7B & Single pass & 75.5 & 49.9 & 57.8 & \textbf{55.3} & 40.2 & 44.9 & 38.3 & 56.1 & 52.2 & 0.661  \\
\rowcolor{reloadbg}\cellcolor{white} & RELOAD ($K{=}4$) & \textbf{77.7} & \textbf{50.6} & \textbf{59.2} & \textbf{55.3} & \textbf{45.2} & \textbf{48.3} & \textbf{38.4} & \textbf{57.2} & \textbf{54.0} & \textbf{0.640}  \\
\midrule
Qwen3-4B & Single pass & 80.4 & \textbf{52.2} & 58.3 & 60.4 & \textbf{56.8} & 50.7 & 43.5 & \textbf{60.3} & 57.8 & 0.615  \\
\rowcolor{reloadbg}\cellcolor{white} & RELOAD ($K{=}4$) & \textbf{87.6} & 48.6 & \textbf{59.6} & \textbf{61.9} & 55.3 & \textbf{56.7} & \textbf{47.3} & 59.5 & \textbf{59.6} & \textbf{0.607}  \\
\bottomrule
\end{tabular}}
\end{table}

\begin{table}[!htbp]
\centering
\footnotesize
\caption{Brier score (lower is better) of the single pass and RELOAD ($K{=}4$) on the held-out test sets (three seeds), raw and, in parentheses, after temperature scaling fitted on development data. Bold: the lower raw score. $^{*}$: significant paired improvement.}
\label{tab:calib}
\fitbox{\begin{tabular}{llrrrr}
\toprule
Model & & ProofWriter & StepGame & Kev decision & Kev transfer \\
\midrule
Von & Single pass & 0.019 (0.018) & 0.242 (0.231) & 0.205 (0.199) & 0.344 (0.333)  \\
\rowcolor{reloadbg}\cellcolor{white} & RELOAD ($K{=}4$) & \textbf{0.005}$^{*}$ (0.005) & \textbf{0.124}$^{*}$ (0.120) & \textbf{0.201} (0.194) & \textbf{0.323} (0.311)  \\
Ettin-400M & Single pass & 0.024 (0.023) & 0.234 (0.224) & 0.193 (0.189) & 0.447 (0.434)  \\
\rowcolor{reloadbg}\cellcolor{white} & RELOAD ($K{=}4$) & \textbf{0.007}$^{*}$ (0.007) & \textbf{0.117}$^{*}$ (0.114) & \textbf{0.192} (0.188) & \textbf{0.399}$^{*}$ (0.390)  \\
Ettin-1B & Single pass & 0.022 (0.021) & 0.171 (0.161) & 0.195 (0.188) & 0.421 (0.405)  \\
\rowcolor{reloadbg}\cellcolor{white} & RELOAD ($K{=}4$) & \textbf{0.007}$^{*}$ (0.006) & \textbf{0.086}$^{*}$ (0.085) & \textbf{0.189} (0.182) & \textbf{0.356}$^{*}$ (0.347)  \\
Qwen3-0.6B & Single pass & 0.146 (0.141) & 0.419 (0.417) & \textbf{0.226} (0.215) & \textbf{0.432} (0.426)  \\
\rowcolor{reloadbg}\cellcolor{white} & RELOAD ($K{=}4$) & \textbf{0.101}$^{*}$ (0.098) & \textbf{0.370}$^{*}$ (0.368) & \textbf{0.226} (0.215) & 0.476 (0.452)  \\
Qwen3-1.7B & Single pass & 0.111 (0.107) & 0.338 (0.333) & 0.203 (0.193) & 0.380 (0.366)  \\
\rowcolor{reloadbg}\cellcolor{white} & RELOAD ($K{=}4$) & \textbf{0.069}$^{*}$ (0.066) & \textbf{0.315}$^{*}$ (0.307) & \textbf{0.199} (0.189) & \textbf{0.325}$^{*}$ (0.318)  \\
Qwen3-4B & Single pass & 0.069 (0.066) & 0.315 (0.304) & 0.204 (0.188) & 0.281 (0.274)  \\
\rowcolor{reloadbg}\cellcolor{white} & RELOAD ($K{=}4$) & \textbf{0.038}$^{*}$ (0.037) & \textbf{0.235}$^{*}$ (0.225) & \textbf{0.193}$^{*}$ (0.181) & \textbf{0.269} (0.260)  \\
\bottomrule
\end{tabular}}
\end{table}

\FloatBarrier
\section{Additional results}\label{app:controls}
\paragraph{Training compute.} Table~\ref{tab:compute} extends the compute control of Table~\ref{tab:controls} with six epochs of single-pass training and with the zero-shot benchmarks in the out-of-distribution group. From two to twelve epochs, the out-of-distribution Brier score of the single pass rises from 0.728 to 0.884.

\begin{table*}[!htbp]
\centering
\footnotesize
\caption{Training compute (Von, three seeds; groups as in Table~\ref{tab:main}): single-pass models trained longer versus RELOAD ($K{=}4$) trained for the standard two epochs. Compute: training layer applications relative to the two-epoch single pass.}
\label{tab:compute}
\fitbox{\begin{tabular}{lrrrrrrr}
\toprule
 & & \multicolumn{2}{c}{In-distribution} & \multicolumn{2}{c}{Longer reasoning} & \multicolumn{2}{c}{Out-of-distribution} \\
\cmidrule(lr){3-4}\cmidrule(lr){5-6}\cmidrule(lr){7-8}
Model & Compute & Acc. & Brier & Acc. & Brier & Acc. & Brier \\
\midrule
Single pass, 2 epochs & 1.0 & 92.8 & 0.101 & 70.6 & 0.404 & 50.9 & 0.728 \\
Single pass, 6 epochs & 3.0 & 94.1 & 0.102 & 73.9 & 0.433 & 50.2 & 0.848 \\
Single pass, 12 epochs & 6.0 & 94.1 & 0.107 & 74.9 & 0.455 & 49.8 & 0.884 \\
\rowcolor{reloadbg}RELOAD ($K{=}4$), 2 epochs & 5.9 & 93.9 & 0.085 & 85.7 & 0.198 & 54.1 & 0.675 \\
\bottomrule
\end{tabular}}
\end{table*}

\paragraph{Training data.} Table~\ref{tab:scaling} gives the gain of four passes for Von trained on the original 46k-question mixture and on nested broad mixtures. RELOAD trained on the $2\times$ mixture beats the single pass trained on the $4\times$ mixture. Broad data also matters out of domain: on Kev transfer the loop did not help with the original reasoning-heavy mixture, but helps clearly at $2\times$ and no longer hurts at $4\times$.

\begin{table}[!htbp]
\centering
\footnotesize
\caption{Macro accuracy (\%) over the held-out ProofWriter, StepGame, Kev decision and Kev transfer test sets for Von trained on the original 46k-question mixture and on nested broad mixtures with two and four times as many questions (three seeds).}
\label{tab:scaling}
\fitbox{\begin{tabular}{lrrl}
\toprule
Training mixture & Single pass & RELOAD ($K{=}4$) & Gain \\
\midrule
Original (45,576 questions) & 79.79 & 82.21 & $+2.42$ \ci{+1.53}{+3.32} \\
Broad, $2\times$ (102,707) & 82.93 & 86.97 & $+4.04$ \ci{+3.09}{+5.00} \\
Broad, $4\times$ (193,038) & 85.77 & 88.53 & $+2.76$ \ci{+2.10}{+3.46} \\
\bottomrule
\end{tabular}}
\end{table}

\paragraph{Depth up-scaling.} Table~\ref{tab:dus} compares RELOAD and depth up-scaling per test set for Von and Ettin-400M.

\begin{table*}[!htbp]
\centering
\footnotesize
\caption{RELOAD versus depth up-scaling \citep{kim2024solar} at equal compute ($K{=}4$; held-out test accuracy, \%, three seeds). Depth up-scaling gives passes 2--4 their own trained copies of the block.}
\label{tab:dus}
\begin{tabular}{llrrrrrr}
\toprule
Model & Variant & Params & ProofWriter & StepGame & Kev dec. & Kev transfer & Macro \\
\midrule
Von & Single pass & 395M & 98.88 & 82.23 & 85.61 & 76.37 & 85.77 \\
\rowcolor{reloadbg}\cellcolor{white} & RELOAD ($K{=}4$) & 395M & 99.71 & 90.88 & 86.03 & 77.49 & 88.53 \\
 & Depth up-scaling & 726M & 99.68 & 90.38 & 86.14 & 79.78 & 88.99 \\
\midrule
Ettin-400M & Single pass & 395M & 98.56 & 82.90 & 86.92 & 68.29 & 84.17 \\
\rowcolor{reloadbg}\cellcolor{white} & RELOAD ($K{=}4$) & 395M & 99.56 & 91.45 & 86.67 & 70.68 & 87.09 \\
 & Depth up-scaling & 726M & 99.51 & 91.77 & 86.47 & 73.63 & 87.84 \\
\bottomrule
\end{tabular}
\end{table*}

\paragraph{What makes the retrofit work.} \citet{gan2026sansi} and \citet{song2026lev} report no gain from loops added to conventionally pretrained models. Both train LoRA adapters on a frozen network, and \citet{song2026lev} also use a gated update. To find which difference matters, we fix the backbone (Von), the data and the budgets and vary the trainable weights and the update (Table~\ref{tab:retrofit}). \emph{LoRA}: the network is frozen, and rank-16 adapters ($\alpha=32$) on every linear layer are trained together with the scorer, with the learning rate ($3\times10^{-4}$ out of $\{10^{-4}, 3\times10^{-4}\}$) chosen on development data. \emph{Gated update}: as in \citet{song2026lev}, the first pass runs the block unchanged, and every later pass updates each block layer $L_i$ as $h\leftarrow h+g\odot(L_i(h)-h)$ with per-channel gates $g$ initialized at zero and no damping; all weights are fine-tuned. Both configurations were fixed on development data before the test sets were evaluated once. With LoRA, the damped loop still gains 13 points on longer reasoning over a LoRA single pass, close to its gain with full fine-tuning, whereas the gated update gains nothing even with full fine-tuning. The update rule, not the restriction to adapters, thus decides whether the retrofitted loop helps, and full fine-tuning adds about five points on longer reasoning on top.

\begin{table}[!htbp]
\centering
\footnotesize
\caption{Retrofit ablation (Von, held-out test, three seeds). Accuracy (\%) at four passes and, below it, the gain over the single pass with the same trainable weights (for the gated update, the fully fine-tuned single pass), in points with 95\% intervals from the crossed bootstrap. $^{*}$: interval excludes zero.}
\label{tab:retrofit}
\fitbox{\begin{tabular}{llrrr}
\toprule
Trainable & Update & In-dist. & Longer & OOD \\
\midrule
\rowcolor{reloadbg}All weights & Damped (RELOAD) & 93.9 & 85.7 & 54.1 \\
\rowcolor{reloadbg} & & $+1.1^{*}$ {\scriptsize\ci{+0.6}{+1.6}} & $+15.1^{*}$ {\scriptsize\ci{+12.3}{+17.9}} & $+3.2^{*}$ {\scriptsize\ci{+2.1}{+4.2}} \\
LoRA & Damped & 93.4 & 81.0 & 52.7 \\
 & & $+0.8^{*}$ {\scriptsize\ci{+0.2}{+1.5}} & $+13.2^{*}$ {\scriptsize\ci{+10.5}{+15.9}} & $+2.9^{*}$ {\scriptsize\ci{+1.7}{+4.2}} \\
All weights & Gated & 92.9 & 68.6 & 51.2 \\
 & & $+0.1$ {\scriptsize\ci{-0.4}{+0.5}} & $-2.0$ {\scriptsize\ci{-4.6}{+0.5}} & $+0.3$ {\scriptsize\ci{-1.0}{+1.4}} \\
\bottomrule
\end{tabular}}
\end{table}

\paragraph{Inference cost.} Table~\ref{tab:cost} gives latency, throughput and memory for all six models.

\begin{table*}[!htbp]
\centering
\footnotesize
\caption{Inference cost on one A100-40GB GPU in bf16 (1,600 test questions, seed 1): latency per question at batch size 1, and throughput and peak memory with batching, for the single pass, RELOAD at $K{=}2$ and $K{=}4$, and RELOAD with the controller ($\lambda{=}0.005$, timed end to end, including its features and logistic regressions). With batching, requests are first routed and then batched by their number of passes, continuing from the stored states before the loop.}
\label{tab:cost}
\fitbox{\begin{tabular}{lrrrrrrrrrrr}
\toprule
 & & \multicolumn{4}{c}{ms per question (batch 1)} & \multicolumn{4}{c}{questions/s (batched)} & \multicolumn{2}{c}{peak GB} \\
\cmidrule(lr){3-6}\cmidrule(lr){7-10}\cmidrule(lr){11-12}
 & & & \multicolumn{3}{c}{RELOAD} & & \multicolumn{3}{c}{RELOAD} & & RELOAD \\
\cmidrule(lr){4-6}\cmidrule(lr){8-10}\cmidrule(lr){12-12}
Model & Params & Single pass & $K{=}2$ & ctrl. & $K{=}4$ & Single pass & $K{=}2$ & ctrl. & $K{=}4$ & Single pass & $K{=}4$ \\
\midrule
Von & 395M & 22.6 & 28.2 & 28.3 & 41.0 & 511 & 421 & 345 & 306 & 2.8 & 2.8 \\
Ettin-400M & 395M & 22.6 & 28.2 & 29.0 & 41.5 & 510 & 420 & 339 & 305 & 2.8 & 2.8 \\
Ettin-1B & 1030M & 22.2 & 28.3 & 28.6 & 40.7 & 303 & 239 & 206 & 169 & 6.7 & 6.7 \\
Qwen3-0.6B & 597M & 35.5 & 44.9 & 46.9 & 66.7 & 360 & 275 & 232 & 187 & 3.9 & 3.9 \\
Qwen3-1.7B & 1722M & 35.8 & 45.7 & 44.6 & 67.7 & 222 & 168 & 153 & 113 & 10.2 & 10.2 \\
Qwen3-4B & 4024M & 46.7 & 59.6 & 61.6 & 86.7 & 105 & 80 & 72 & 54 & 23.4 & 23.4 \\
\bottomrule
\end{tabular}}
\end{table*}

\paragraph{Controller baselines.} Table~\ref{tab:ctrl_baselines} compares the controller with an input-independent mix of the next lower and next higher fixed budget at the same mean number of passes and, for Von, with post-hoc halting rules. These run the budgets in order and stop once the answer distribution changes by less than a threshold in total variation, or once the top probability exceeds a threshold. Each budget is its own trajectory, so a rule that stops at budget $K$ has also paid for all smaller budgets.

\begin{table}[!htbp]
\centering
\footnotesize\setlength{\tabcolsep}{4pt}
\caption{Controller baselines (three seeds). Top: controller ($\lambda{=}0.005$) minus a mix of the neighboring fixed budgets at the same mean number of passes, paired accuracy difference in points with 95\% intervals. Bottom: Von, mean accuracy over ProofWriter, StepGame, Kev decision and Kev transfer, and layer applications per question, for the controller and for post-hoc halting rules (total variation $\tau$ between consecutive budgets, or top probability above $c$).}
\label{tab:ctrl_baselines}
\fitbox{\begin{tabular}{lrrrr}
\toprule
Model & Mean passes & In-distribution & Longer reasoning & Out-of-distribution \\
\midrule
Von & 2.63 & $-0.26$ \ci{-0.48}{-0.04} & $+2.52$ \ci{+1.55}{+3.59} & $-0.24$ \ci{-0.65}{+0.22} \\
Ettin-400M & 2.50 & $-0.13$ \ci{-0.31}{+0.06} & $+2.45$ \ci{+1.24}{+3.63} & $-0.36$ \ci{-0.80}{+0.06} \\
Ettin-1B & 2.59 & $-0.06$ \ci{-0.26}{+0.14} & $+2.45$ \ci{+1.64}{+3.34} & $-0.25$ \ci{-0.57}{+0.06} \\
Qwen3-0.6B & 2.30 & $-0.02$ \ci{-0.20}{+0.17} & $+0.31$ \ci{-0.29}{+1.07} & $-0.58$ \ci{-0.94}{-0.23} \\
Qwen3-1.7B & 2.59 & $-0.13$ \ci{-0.34}{+0.09} & $+0.12$ \ci{-0.58}{+0.91} & $-0.16$ \ci{-0.67}{+0.32} \\
Qwen3-4B & 2.42 & $-0.11$ \ci{-0.30}{+0.10} & $+0.42$ \ci{-0.13}{+1.00} & $-0.32$ \ci{-0.75}{+0.10} \\
\bottomrule
\end{tabular}}

\vspace{6pt}
\begin{tabular}{lrr}
\toprule
Von & Accuracy (\%) & Layers \\
\midrule
\rowcolor{reloadbg}RELOAD + controller, $\lambda{=}0.005$ & 88.16 & 37.1 \\
\rowcolor{reloadbg}RELOAD + controller, $\lambda{=}0.002$ & 88.35 & 40.0 \\
Halting, $\tau{=}0.2$ & 88.12 & 59.0 \\
Halting, $\tau{=}0.1$ & 88.37 & 62.0 \\
Halting, $c{=}0.95$ & 88.19 & 57.4 \\
Halting, $c{=}0.99$ & 88.42 & 67.6 \\
\bottomrule
\end{tabular}
\end{table}

\paragraph{StepGame by story and chain length.} Table~\ref{tab:hops} gives Von's StepGame accuracy and the controller's mean number of passes by the number of statements in the story and in the chain.

\begin{table}[!htbp]
\centering
\footnotesize
\caption{StepGame test accuracy (\%) for Von (three seeds) by the number of statements in the story and in the chain linking the two queried objects, for the single pass, RELOAD ($K{=}4$) and RELOAD with the controller ($\lambda{=}0.005$), with the controller's mean number of passes. Training stories have 1--5 statements.}
\label{tab:hops}
\fitbox{\begin{tabular}{lrrrrr}
\toprule
Statements & $n$ & Single pass & RELOAD ($K{=}4$) & RELOAD + controller & Controller passes \\
\midrule
\multicolumn{6}{l}{\emph{Statements in the story}} \\
1--5 & 1000 & 93.9 & 96.1 & 96.0 & 2.46 \\
6--10 & 1000 & 70.6 & 85.7 & 85.4 & 3.80 \\
\midrule
\multicolumn{6}{l}{\emph{Statements in the chain}} \\
1 & 806 & 98.0 & 97.9 & 97.9 & 2.64 \\
2 & 459 & 89.8 & 93.1 & 93.0 & 3.17 \\
3 & 267 & 71.2 & 91.6 & 91.0 & 3.46 \\
4 & 181 & 57.8 & 80.8 & 80.5 & 3.65 \\
5 & 128 & 57.3 & 76.6 & 76.6 & 3.76 \\
6--10 & 159 & 47.0 & 70.4 & 70.0 & 3.89 \\
\bottomrule
\end{tabular}}
\end{table}

\paragraph{Object tracking and linear probes.} To see how the answer forms across passes, we use synthetic object tracking, where the state after every step is known: four objects in four boxes, a sequence of swaps, and the question where one object ends up. On 4,000 new problems each with 3, 6 and 8 swaps (training covers up to 4), we run Von and Ettin-400M with four passes, read out the answer after each pass by applying the rest of the network to the intermediate state, and fit linear probes on the option-marker states that predict where the object is after each swap (3,000 problems for training and 1,000 for testing; chance 25\%). The answer forms gradually, although the model is never trained to answer after an intermediate pass, and the object's final location becomes linearly readable step by step (Table~\ref{tab:probe}). Four passes solve 6-swap problems almost perfectly and recover much of the accuracy on 8-swap problems, where the single-pass models fall below 50\%. Intermediate swap states remain only partly linearly readable (at most about 50\% for the middle swap of 6-swap problems), which, together with the growing readability of the final location, is consistent with refining the whole answer state, as in the ODE view, where the state after pass $k$ of four approximates $h(k/4)$. The probes leave open intermediate computations represented nonlinearly or at other token positions.

\begin{table*}[!htbp]
\centering
\footnotesize
\caption{Object tracking with 3, 6 and 8 swaps (training covers up to 4; three seeds). Answer: accuracy of the answer read out after each pass of a four-pass run (for the single pass, its normal output). Probe: accuracy of a linear probe on the option-marker states at the output of the looped block that predicts the object's final location (chance 25\%).}
\label{tab:probe}
\begin{tabular}{llrrrrrr}
\toprule
 & & \multicolumn{2}{c}{3 swaps} & \multicolumn{2}{c}{6 swaps} & \multicolumn{2}{c}{8 swaps} \\
\cmidrule(lr){3-4}\cmidrule(lr){5-6}\cmidrule(lr){7-8}
Model & State & Answer & Probe & Answer & Probe & Answer & Probe \\
\midrule
 & Single pass & 100.0 & 95.4 & 76.5 & 40.1 & 44.2 & 27.8 \\
\rowcolor{reloadbg}\cellcolor{white} & RELOAD ($K{=}4$), after pass 1 & 99.1 & 92.9 & 56.8 & 35.0 & 34.1 & 25.4 \\
\rowcolor{reloadbg}\cellcolor{white} & RELOAD ($K{=}4$), after pass 2 & 100.0 & 99.3 & 80.7 & 58.2 & 46.2 & 34.0 \\
\rowcolor{reloadbg}\cellcolor{white} & RELOAD ($K{=}4$), after pass 3 & 100.0 & 99.8 & 94.8 & 80.6 & 63.1 & 51.2 \\
\rowcolor{reloadbg}\cellcolor{white}\multirow{-5}{*}{Von}& RELOAD ($K{=}4$), after pass 4 & 100.0 & 99.9 & 98.3 & 92.1 & 74.6 & 66.2 \\
\midrule
 & Single pass & 100.0 & 92.3 & 83.2 & 40.0 & 48.4 & 26.7 \\
\rowcolor{reloadbg}\cellcolor{white} & RELOAD ($K{=}4$), after pass 1 & 99.8 & 83.4 & 69.5 & 38.3 & 41.4 & 25.8 \\
\rowcolor{reloadbg}\cellcolor{white} & RELOAD ($K{=}4$), after pass 2 & 100.0 & 99.4 & 94.9 & 66.5 & 67.9 & 38.2 \\
\rowcolor{reloadbg}\cellcolor{white} & RELOAD ($K{=}4$), after pass 3 & 100.0 & 99.7 & 98.3 & 87.5 & 80.8 & 63.3 \\
\rowcolor{reloadbg}\cellcolor{white}\multirow{-5}{*}{Ettin-400M}& RELOAD ($K{=}4$), after pass 4 & 100.0 & 99.9 & 99.1 & 94.5 & 85.8 & 78.3 \\
\bottomrule
\end{tabular}
\end{table*}

\FloatBarrier
\section{Datasets}\label{app:data}
All tasks share one format: a context (the \emph{state}), a question and a fixed list of answer options, of which the model selects one.

\paragraph{Training mixture.} The broad mixture of 193,038 questions has four parts.
(i) Ten public classification and question-answering sets posed as decisions (52,000 questions): AG News topic, Yelp and Amazon review rating, DBpedia entity type, Banking77 customer intent, TREC question type, MNLI entailment, IMDb and SST-5 sentiment, and BoolQ yes/no reading comprehension.
(ii) Adversarial natural-language inference from ANLI R1--R3 and WANLI, posed as entailment, neutral or contradiction (24,000).
(iii) Kev's policy and compositional decision sets (9,044): synthetic documents with rules, such as a refund or access policy, and questions whose answer follows from combining several rules.
(iv) Multi-step reasoning data (107,994): object tracking (objects in boxes are swapped several times; which box holds a given object at the end?), Boolean circuits (evaluate a chain of logic gates), ProofWriter, CLUTRR (infer a family relation such as ``aunt'' from a chain of 2--4 stated relations) and StepGame stories with 1--5 statements.

\paragraph{ProofWriter} \citep{tafjord2021proofwriter} tests logical deduction in natural language. Each problem is a short synthetic text of facts and if-then rules about people, animals and attributes (``The lion sees the cow. If something is green then it eats the tiger.''), and the model decides whether a statement is true (it follows), false (its negation follows) or unknown (neither follows, under the open-world assumption), with proofs of depth 0 to 5. We train on the training split and test on the held-out test split.

\paragraph{StepGame} \citep{shi2022stepgame} describes the relative positions of objects in a sequence of sentences (``N and W are next to each other with W on the right'') and asks for the spatial relation between two of them, one of nine labels (left, right, above, below, the four diagonals, overlap). A story of $k$ statements describes a path through $k{+}1$ objects, and the question asks about two objects on this path, so answering requires chaining one to $k$ statements, and the others act as distractors. We call $k$ the story length and the number of statements to chain the chain length, which we recover by parsing the stories. Training covers stories of 1--5 statements. The test set has 200 problems for each story length from 1 to 10, of which 806 need one statement, 726 two or three, 309 four or five and 159 six to ten.

\paragraph{Kev decision and Kev transfer} come from the Kev decision benchmark \citep{kev}. Kev decision is the held-out test split of the training sources (the ten public sets and the policy minimal pairs). Kev transfer contains decisions from six sources never seen in training (MMLU, Emotion, TweetEval offensive-language detection, QNLI, PAWS paraphrase identification and SciQ) and rule structures held out from the policy data. The Kev repository is released under Apache-2.0, and its decision suites are built from public datasets under their own licenses.

\begin{sloppypar}
\paragraph{Zero-shot reasoning benchmarks.} These are used only for evaluation; parentheses give the number of questions.
\emph{ParaRules} \citep{tafjord2021proofwriter}: ProofWriter-style theories rewritten by crowd workers in natural, varied language (``That guy Alan sure is nice.''), with true, false or unknown answers and proof depths 0--5 (1,755, at most 400 per depth).
\emph{JustLogic} \citep{justlogic}: deductive arguments built from natural sentences, where the answer depends only on the stated premises and not on world knowledge; true, false or uncertain; argument depth 1--7 (1,050).
\emph{ProverQA} \citep{proverqa}: first-order-logic problems written in natural language and checked by a theorem prover, with reasoning chains of up to 9 steps; true, false or uncertain (1,500).
\emph{BoardgameQA} \citep{boardgameqa}: the state of a fictional board game and rules with exceptions and preferences between conflicting rules; the answer is whether a statement is proved, disproved or unknown, at depths 1--3 (3,000).
\emph{MuSR} \citep{musr}: long narratives (murder mysteries, object placement, team allocation) followed by a question that requires combining many clues, such as who is the most likely murderer (752; stories are truncated to 1,024 tokens).
\emph{ReClor} \citep{reclor}: logical reasoning questions from standardized graduate-admission exams, such as identifying the flaw in an argument, with four options (500, validation split).
\emph{LogiQA~2.0} \citep{logiqa2}: logical reasoning questions from Chinese civil-service exams in English translation, with four options (1,551).
\emph{ReSQ} \citep{resq}: yes/no spatial questions about short descriptions of real photographs (661).
\end{sloppypar}

\FloatBarrier
\section{Training details}\label{app:training}
\paragraph{Mixture.} All models are fine-tuned for two epochs on the broad mixture of Appendix~\ref{app:data}. None of the questions added for the broad mixture overlaps any development or test item. Six of the 2,000 StepGame test stories, all with a single statement, also occur in the training data; excluding them changes in-distribution accuracy by at most 0.04 points and the Brier score by at most 0.0003 for every model and budget and leaves the other groups unchanged, so we keep all 2,000.

\paragraph{Recipe.} Full fine-tuning with AdamW, a cosine schedule with 8\% warmup, gradient clipping at 1.0, identical token-budget batches for every arm and bf16 autocast. For the encoders the learning rate is $1.5\times10^{-5}$ for the network and $7.5\times10^{-5}$ for the looped block and the scorer; on Von's development data, training the looped block faster than the rest of the network improved both the single-pass and the looped models. For the Qwen3 models it is $10^{-5}$ for the network, including the block, and $5\times10^{-5}$ for the head, with activation checkpointing; Qwen3-4B uses 8-bit AdamW to fit into GPU memory. Learning rates were chosen on development data with one seed, identically for all arms of a model: on Von for the encoders (the Ettin encoders use the same values) and on Qwen3-1.7B for the Qwen3 models.

\paragraph{Controller data.} The controller is fitted on the development splits of the training tasks and of Kev transfer; on development splits of ARC-Challenge, bAbI, BBH, CommonsenseQA and FOLIO, which are not among our test or zero-shot benchmarks; and on development sets with longer reasoning chains (StepGame stories with up to 10 statements, longer CLUTRR chains and deeper synthetic problems). A LogiQA development set that shares passages with LogiQA~2.0 is excluded.

\FloatBarrier
\section{The loop as a numerical integrator}\label{app:ode}
We state the standard results behind the ODE view of the loop and check them on our models; textbook treatments are \citet{hairer1993solving}, \citet{butcher2016numerical} and \citet{iserles2009first}. Let $F(h)=G(h)-h$ and let $h(t)$ solve Equation~\eqref{eq:ode} on $[0,1]$. Write $h_K$ for the result of $K$ Euler steps of size $1/K$ (Equation~\eqref{eq:loop}) and, for even $K$, $h^{\mathrm{mid}}_K$ for the result of $K/2$ explicit midpoint steps of size $2/K$, which also evaluate $G$ $K$ times.

\paragraph{Error bounds.} Fix a norm on the hidden states and assume that $F$ is twice continuously differentiable, with bounded first and second derivatives, on an open neighborhood of the trajectory. By the convergence of Euler's method and of second-order Runge--Kutta methods (\citealp[Section~II.3]{hairer1993solving}; \citealp{butcher2016numerical,iserles2009first}), there are constants $C_E$, $C_M$ and $K_0$, depending on $F$ and $h_0$ but not on $K$, such that for $K\ge K_0$ (even $K$ for the midpoint method) the numerical states and midpoint stages stay in this neighborhood and
\begin{equation*}
\|h_K - h(1)\| \le \frac{C_E}{K}, \qquad \|h^{\mathrm{mid}}_K - h(1)\| \le \frac{C_M}{K^2}.
\end{equation*}
Let $Q=\mathrm{softmax}\circ f\circ S$ map a hidden state to the option probabilities and assume $\|Q(u)-Q(v)\|_\infty \le L\|u-v\|$ for $u,v$ in the neighborhood. Then the option probabilities $p_K = Q(h_K)$ satisfy $\|p_K - p_\infty\|_\infty \le L C_E / K$, where $p_\infty=Q(h(1))$, and likewise with $L C_M/K^2$ for the midpoint method.

\paragraph{Margins.} Let $\gamma>0$ be the gap between the largest and the second-largest entry of $p_\infty$. If every probability is perturbed by less than $\gamma/2$, the largest entry stays the largest, an argument similar to margin-based robustness certificates for Lipschitz networks \citep{tsuzuku2018lipschitz}. So for $K\ge K_0$ the answer at budget $K$ equals that of the flow whenever $K > 2LC_E/\gamma$ for Euler and $K > (2LC_M/\gamma)^{1/2}$ for the midpoint method. For fixed constants, a smaller margin raises these sufficient budgets, more slowly for the midpoint method, although its constant decides whether it saves evaluations at a given budget. Training determines $p_\infty$ and its margins, and the number of passes how closely the computation follows the flow. If the answer of the flow is wrong, all sufficiently large budgets give the same wrong answer, whereas small budgets can differ from it in either direction.

\paragraph{Scope.} These are asymptotic statements without certified constants for the trained transformers, so we use them to explain the behavior of the loop rather than to predict it. The two-step midpoint configuration with steps $1/3$ and $2/3$ of Section~\ref{sec:analysis} is not in the family above; the bound motivates it but does not cover it.

\paragraph{Convergence toward a fine budget.} Table~\ref{tab:convergence} measures how far the result at budget $K$ lies from that at budget 8 on 1,795 questions from the test sets and from synthetic object tracking (Von, three seeds): for the answer, the mean absolute difference of the option log-probabilities; for the state, the distance of the option-marker states relative to their displacement at budget 8. A fit of $c\,(1/K-1/8)^{\alpha}$ over budgets 1 to 6 gives $\alpha$ close to 1, the order of Euler's method, both for the trained loop and for the single-pass model looped only at inference.

\begin{table}[!htbp]
\centering
\footnotesize
\caption{Distance from budget 8 at budget $K$ and fitted exponent $\alpha$ (Von, mean over three seeds).}
\label{tab:convergence}
\begin{tabular}{lrrrrrr}
\toprule
 & $K{=}1$ & 2 & 3 & 4 & 6 & $\alpha$ \\
\midrule
\rowcolor{reloadbg}RELOAD, answer (log-probability) & 1.77 & 0.78 & 0.42 & 0.25 & 0.08 & 1.01 \\
\rowcolor{reloadbg}RELOAD, option-marker states & 0.355 & 0.132 & 0.070 & 0.041 & 0.014 & 1.07 \\
Looped only at inference, option-marker states & 0.376 & 0.129 & 0.067 & 0.040 & 0.013 & 1.10 \\
\bottomrule
\end{tabular}
\end{table}

\paragraph{Empirical check.} As a reference for $h(1)$ we use the classical fourth-order Runge--Kutta method with four steps (16 evaluations of $G$), which changes 0.09\% of the answers on the development sets compared with three steps. Against this reference, the mean largest probability error of Von on the development sets (three seeds) falls with a fitted order of 0.91 to 0.98 for Euler with 1 to 16 steps and 1.78 to 1.89 for the midpoint method with 1 to 4 steps, close to the orders 1 and 2 of the bounds. Between one and two steps, Runge--Kutta has an effective order of only 2.6 to 2.7 against the same reference, so these measurements do not show its nominal fourth order; error in the reference and the bfloat16 arithmetic may also affect the fit. The fitted order also varies between questions: for Euler, among the 11,672 question-seed pairs with a probability error of at least $10^{-4}$ at 4, 8 and 16 steps, its 10th and 90th percentiles are 0.29 and 1.92. The bounds hold for each question, but their constants and the budget from which they apply differ between questions, so a fit at a few budgets need not show the formal order on every question.

\paragraph{All six models.} Table~\ref{tab:integrators_all} gives the test comparison for every model, with the configurations fixed on Von's development data and evaluated once. For the Qwen3 models the midpoint method raises the Brier score by 0.001 to 0.005 in distribution and by 0.004 to 0.011 out of distribution; four of these six differences are significant.

\begin{table}[!htbp]
\centering
\footnotesize\setlength{\tabcolsep}{4pt}
\caption{Midpoint method with two steps against Euler at test time for all six models (three seeds; differences with 95\% intervals from the crossed bootstrap; $^{*}$: interval excludes zero). $K$ counts evaluations of $G$. Longer: longer reasoning; In: in distribution; OOD: out of distribution. Accuracy in points; Brier score in absolute units (negative is better).}
\label{tab:integrators_all}
\fitbox{%
\begin{tabular}{lrrrrrr}
\toprule
 & RELOAD ($K{=}8$; Euler) & \multicolumn{4}{c}{RELOAD ($K{=}4$; midpoint)} & RELOAD ($K{=}4$; midpoint) \\
 & $-$ RELOAD ($K{=}4$; Euler) & \multicolumn{4}{c}{$-$ RELOAD ($K{=}4$; Euler)} & $-$ RELOAD ($K{=}8$; Euler) \\
\cmidrule(lr){2-2}\cmidrule(lr){3-6}\cmidrule(lr){7-7}
Model & Longer acc. & Longer acc. & Longer Brier & In acc. & OOD acc. & Longer acc. \\
\midrule
Von & $+1.5$ & $+1.7^{*}$ [$+0.7$, $+2.8$] & $-0.022^{*}$ [$-0.035$, $-0.009$] & $-0.2$ [$-0.4$, $+0.1$] & $-0.2$ [$-0.6$, $+0.2$] & $+0.2$ [$-0.5$, $+0.9$] \\
Ettin-400M & $+0.4$ & $+0.4$ [$-0.7$, $+1.6$] & $-0.004$ [$-0.015$, $+0.007$] & $-0.2$ [$-0.5$, $+0.0$] & $-0.2$ [$-0.6$, $+0.2$] & $+0.0$ [$-0.7$, $+0.8$] \\
Ettin-1B & $+0.6$ & $+0.6$ [$-0.2$, $+1.4$] & $-0.008$ [$-0.017$, $+0.001$] & $-0.2$ [$-0.4$, $+0.0$] & $+0.1$ [$-0.3$, $+0.6$] & $+0.0$ [$-0.5$, $+0.5$] \\
Qwen3-0.6B & $-0.5$ & $-0.4$ [$-1.5$, $+0.6$] & $0.000$ [$-0.006$, $+0.007$] & $-0.2$ [$-0.6$, $+0.1$] & $-0.1$ [$-0.6$, $+0.3$] & $+0.0$ [$-1.2$, $+1.2$] \\
Qwen3-1.7B & $+0.1$ & $+0.2$ [$-1.0$, $+1.5$] & $0.000$ [$-0.009$, $+0.008$] & $-0.3$ [$-0.8$, $+0.1$] & $-0.3$ [$-0.9$, $+0.2$] & $+0.2$ [$-0.7$, $+1.1$] \\
Qwen3-4B & $+0.3$ & $+0.3$ [$-0.8$, $+1.5$] & $-0.003$ [$-0.012$, $+0.006$] & $+0.0$ [$-0.2$, $+0.3$] & $-0.1$ [$-0.5$, $+0.3$] & $+0.1$ [$-0.7$, $+0.9$] \\
\bottomrule
\end{tabular}}
\end{table}

\paragraph{Midpoint steps in the controller.} With a single step, that is with two passes, the midpoint method is less accurate than two Euler passes in distribution ($-0.5$ points [$-0.8$, $-0.1$], Brier $+0.007$; Von, test) and not significantly different on longer reasoning and out of distribution, so a midpoint step pays off only from four passes on. A controller that chooses between one pass and the midpoint method with two or four passes, fitted on development data like the controller of Section~\ref{sec:controller}, is therefore no better overall: it is 2.1 points [$+0.9$, $+3.4$] more accurate on longer reasoning than the Euler controller and not significantly different in and out of distribution, but uses 0.1 to 0.2 more passes on average, and its macro accuracy-compute curve coincides with that of the Euler controller. Keeping the Euler controller's choices and running only its four-pass choices with the midpoint method leaves the compute unchanged and raises accuracy on longer reasoning by 1.5 points [$+0.5$, $+2.6$] (Brier $-0.019$), with in-distribution and out-of-distribution accuracy within 0.05 points. This combination was chosen after the fixed-budget test results of its parts were known.

\FloatBarrier
\section{Controlled StepGame stories}\label{app:horizon}
\paragraph{Generator.} A story with $S$ statements describes a path through $S{+}1$ distinct objects named by capital letters. Each statement is one unit step in one of the eight directions, drawn uniformly and independently, and is written with an original StepGame template drawn with its frequency in the training data. We exclude 13 of the 209 templates whose wording does not match the relation they encode (ambiguous, inverted or missing an object), keeping 196. As in StepGame, the statements appear in random order; the question asks for the relation between the two ends of a chain of $L$ statements, and the label is the direction of the summed offset. A parser recovers the planted chain, the label and a unique geometry for every generated item. In the main design every statement belongs to the chain ($S{=}L$, $L=1,\dots,16$, 300 stories each); in the distractor design the chain sits at a random position on a longer path with $S{=}L{+}5$; and in a third design the story length is fixed at $S{=}10$ with $L=1,\dots,10$.

\paragraph{Fidelity.} We also generate a replica of the real test design (200 stories for each story length from 1 to 10) whose chain-length histogram, label distribution and story lengths match the real test set. Per cell of story length and chain length, the generated stories are 1.4 to 3.9 points easier than the real ones for every model and budget. The difference disappears ($-1.1$ to $+2.3$ points) when the real stories that use one of the excluded templates are removed, so it comes from those templates rather than from the generator.

\paragraph{Reasoning horizon.} For each model, budget, design and seed we fit
\begin{equation*}
P(\text{correct}\mid L) = c(L) + \bigl(1-c(L)\bigr)\,u\,\sigma\!\bigl((m-L)/s\bigr)
\end{equation*}
by maximum likelihood over items, where $c(L)$ is the accuracy of always answering the most frequent label at chain length $L$ (12.5\% at $L{=}1$, rising to 19.6\% at $L{=}16$), $u\le1$ the ceiling, $s$ the width and $\sigma$ the logistic function. The horizon is the midpoint $m$, where accuracy lies halfway between ceiling and chance. Beyond the horizon, accuracy levels off near 30\% rather than at $c(L)$, because the models still get part of the direction right, so the floor of the fit is slightly misspecified at long chains. A nonparametric horizon, the first chain length at which accuracy normalized between chance and ceiling falls below one half, does not depend on the floor; it is 0.5 to 1.6 statements lower but orders the models and budgets in the same way, as do three further variants of the definition. With the story length fixed at ten statements, which separates chain length from story length, Von's horizon grows from 4.4 with a single pass to 8.6 with four passes.

\paragraph{Distractors.} Table~\ref{tab:distractors} gives the cost of five statements off the chain at chain lengths of one to five statements. For the Qwen3 models the loop barely reduces this cost.

\begin{table}[!htbp]
\centering
\footnotesize
\caption{Accuracy lost (points) when five statements off the chain are added to the generated StepGame stories, mean over chain lengths 1--5 (three seeds).}
\label{tab:distractors}
\begin{tabular}{lrrr}
\toprule
Model & Single pass & RELOAD ($K{=}4$) & RELOAD ($K{=}8$) \\
\midrule
Von & 18.9 & 4.1 & 2.9 \\
Ettin-400M & 17.0 & 3.2 & 2.0 \\
Ettin-1B & 16.1 & 2.9 & 0.9 \\
Qwen3-0.6B & 19.2 & 19.3 & 19.4 \\
Qwen3-1.7B & 23.4 & 22.9 & 22.6 \\
Qwen3-4B & 21.6 & 19.6 & 19.3 \\
\bottomrule
\end{tabular}
\end{table}

\FloatBarrier
\section{Base models and released decision models}\label{app:baselines}
Table~\ref{tab:baselines} compares our models with untrained base models and released decision models. Base Qwen3 models answer through the probabilities of the option letters after an ``Answer:'' prompt (renormalized over the valid letters), and ModernBERT and Ettin predict the option letter at a masked position. Von and Kev are evaluated as released (both Kev checkpoints with temperature 1); the released models were not trained on ProofWriter or StepGame.
\begin{table}[!htbp]
\centering
\footnotesize\setlength{\tabcolsep}{4pt}
\caption{Held-out test accuracy (\%) on ProofWriter, StepGame, Kev decision and Kev transfer with their macro accuracy and Brier score, and macro accuracy and Brier score on the eight unseen reasoning benchmarks.}
\label{tab:baselines}
\fitbox{\begin{tabular}{lrrrrrrrr}
\toprule
 & ProofWriter & StepGame & Kev dec. & Kev transfer & Macro & Brier & Unseen acc. & Unseen Brier \\
\midrule
\multicolumn{9}{l}{\textit{Base models, zero-shot letter readout}} \\
ModernBERT-large (masked letter) & 32.9 & 9.6 & 36.0 & 45.0 & 30.9 & 0.769 & 35.2 & 0.733 \\
Ettin-1B (masked letter) & 42.4 & 17.1 & 52.8 & 54.6 & 41.7 & 0.672 & 43.6 & 0.672 \\
Qwen3-0.6B-Base & 37.9 & 12.5 & 62.8 & 59.3 & 43.1 & 0.645 & 42.9 & 0.649 \\
Qwen3-1.7B-Base & 36.1 & 14.0 & 65.9 & 71.2 & 46.8 & 0.595 & 43.9 & 0.639 \\
Qwen3-4B-Base & 40.2 & 19.5 & 73.4 & 76.2 & 52.3 & 0.559 & 52.9 & 0.563 \\
\midrule
\multicolumn{9}{l}{\textit{Released decision models}} \\
Von (as released) & 45.1 & 17.7 & 65.8 & 75.5 & 51.0 & 0.678 & 43.5 & 0.814 \\
Kev-0.6B (Qwen3) & 53.2 & 18.2 & 80.8 & 64.2 & 54.1 & 0.568 & 45.3 & 0.734 \\
Kev-4B (Qwen3) & 31.9 & 27.7 & 85.6 & 80.6 & 56.5 & 0.592 & 52.9 & 0.673 \\
\midrule
\multicolumn{9}{l}{\textit{Ours (broad mixture, three seeds)}} \\
Von, single pass & 98.9 & 82.2 & 85.6 & 76.4 & 85.8 & 0.203 & 47.7 & 0.776 \\
\rowcolor{reloadbg}Von, RELOAD ($K{=}4$) & 99.7 & 90.9 & 86.0 & 77.5 & 88.5 & 0.163 & 51.2 & 0.719 \\
Ettin-1B, single pass & 98.7 & 88.1 & 87.0 & 68.8 & 85.6 & 0.202 & 46.2 & 0.754 \\
\rowcolor{reloadbg}Ettin-1B, RELOAD ($K{=}4$) & 99.6 & 93.8 & 87.3 & 73.8 & 88.6 & 0.159 & 50.6 & 0.685 \\
Qwen3-0.6B, single pass & 90.2 & 65.8 & 84.6 & 68.5 & 77.3 & 0.306 & 50.6 & 0.668 \\
\rowcolor{reloadbg}Qwen3-0.6B, RELOAD ($K{=}4$) & 93.4 & 70.7 & 84.4 & 65.1 & 78.4 & 0.293 & 51.9 & 0.662 \\
Qwen3-1.7B, single pass & 92.8 & 73.3 & 86.0 & 73.6 & 81.4 & 0.258 & 52.2 & 0.661 \\
\rowcolor{reloadbg}Qwen3-1.7B, RELOAD ($K{=}4$) & 95.7 & 75.3 & 86.4 & 76.7 & 83.5 & 0.227 & 54.0 & 0.640 \\
Qwen3-4B, single pass & 95.7 & 76.5 & 85.9 & 80.7 & 84.7 & 0.217 & 57.8 & 0.615 \\
\rowcolor{reloadbg}Qwen3-4B, RELOAD ($K{=}4$) & 97.7 & 83.1 & 87.1 & 81.4 & 87.3 & 0.184 & 59.6 & 0.607 \\
\bottomrule
\end{tabular}}
\end{table}

\FloatBarrier
\section{Input injection}\label{app:injection}

\paragraph{Question and setup.}
Input injection, which re-supplies the input to the recurrent state at every iteration, is common in looped and recurrent-depth models \citep{yang2024looped,geiping2025scaling,shomali2026loopmtp}. \citet{huang2026topology} report that, in looped language models trained from scratch, writing the fixed prelude output back into the recurrent state helps context-dependent tasks but hurts multi-step quantitative reasoning. We test whether this helps RELOAD (Section~\ref{sec:method}). Let $e=h_0$ be the prefix output (the prelude of \citet{huang2026topology}). Instead of the update of Equation~\eqref{eq:loop}, the diagonal injection, which follows their parameterization, computes
\begin{equation*}
D(h_k,e)=\alpha\odot h_k+\delta\odot W_{\mathrm{in}}e,
\qquad
h_{k+1}=h_k+\frac{G(D(h_k,e))-h_k}{K},
\end{equation*}
where $\delta=\mathrm{softplus}(b)$ and $\alpha=\exp(-\delta\odot\exp(a))$ with learned per-channel vectors $a,b$, and $W_{\mathrm{in}}$ is a full matrix. We initialize $W_{\mathrm{in}}=0$ and $\alpha=0.99$, so that the write starts almost inactive, and write either before passes 2 to $K$ (before pass 1 the state is $e$ itself) or before every pass. A trained-gate variant gives the gate parameters a higher learning rate, since at the base rate they stay at their initial values, and an embedding-source variant writes the token embeddings instead of $e$. Two further writes start as the exact identity: FiLM \citep{perez2018film}, which adds an input-dependent channel scale to the additive write, and masked cross-attention from the loop state to the prefix output. Table~\ref{tab:injection_variants} lists all variants and the models they were run on.

All comparisons use three seeds and a plain loop (RELOAD without injection) retrained with the same recipe and data as the injection arms. Rounds 1 to 4 use Ettin-400M and Qwen3-0.6B. A final replication round trains the three variants with the best development accuracy on these two models (cross-attention, embedding source with trained gate, and FiLM), selected before any test result of the new models, together with the diagonal write on every pass, on Ettin-1B, Qwen3-1.7B and Von. We evaluate at $K=4$ on the four test sets of Table~\ref{tab:perset} and the eight zero-shot benchmarks, grouped as in Table~\ref{tab:main}, and also report $K=6,8$ as extrapolation beyond training. Group differences are paired against the same model's plain loop, with 95\% bootstrap intervals that resample seeds and then items within each set; with three seeds they are necessarily coarse.

\begin{table}[!htbp]
\centering
\footnotesize\setlength{\tabcolsep}{3pt}
\caption{Input-injection variants. Rounds 1 to 3 change only the write into the damped loop; round 4 also changes the update.}
\label{tab:injection_variants}
\fitbox{\begin{tabular}{llll}
\toprule
Round & Variant & Change & Models \\
\midrule
1 & Diagonal, passes 2--$K$ & Write $D(h,e)$ before passes 2--$K$ & Ettin-400M, Qwen3-0.6B \\
1 & Diagonal, every pass & Same diagonal write before all passes & All five \\
2 & Prefix source, trained gate & Gate parameters use learning rate $3\!\times\!10^{-3}$ & Ettin-400M, Qwen3-0.6B \\
2 & Embedding source, trained gate & Write token embeddings; gate learning rate $3\!\times\!10^{-3}$ & All five \\
3 & FiLM & Input-dependent channel scale and additive write & All five \\
3 & Cross-attention & Masked state-to-prelude cross-attention write & All five \\
4 & Full-step & Replace damped update with $h_{k+1}=G(h_k)$; test two writes & Ettin-400M, Qwen3-0.6B \\
\bottomrule
\end{tabular}}
\end{table}

\begin{figure}[!htbp]
\centering
\includegraphics[width=\linewidth]{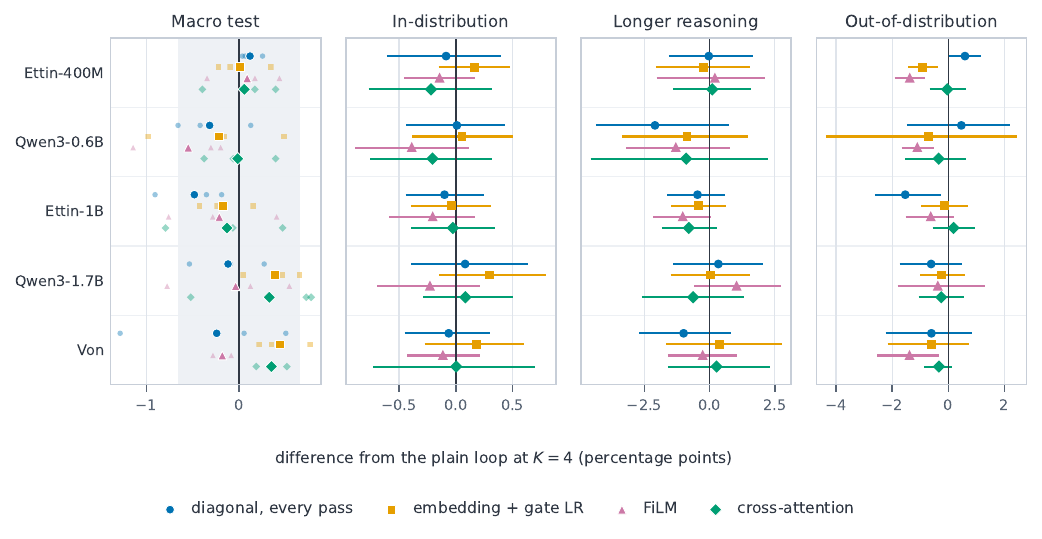}
\caption{Input injection for the four variants run on all five models, relative to each model's plain loop at $K=4$. Left: macro test accuracy; small points are seeds, large points their mean, and the band is the plain-loop seed spread ($\pm0.65$). Other panels: paired differences by group with 95\% bootstrap intervals.}
\label{fig:injection_summary}
\end{figure}

\paragraph{Macro test results.}
Table~\ref{tab:injection_macro} gives the difference in mean accuracy over the four test sets. For every variant and model it lies between $-0.55$ and $+0.45$ points, about the seed-to-seed spread of the plain loop itself (0.65 points; Figure~\ref{fig:injection_summary}, left). Cross-attention and the embedding write gain 0.3 to 0.5 points on Qwen3-1.7B and Von but lose 0.1 to 0.2 on Ettin-1B. The full-step update of round 4 is evaluated separately below.

\begin{table}[!htbp]
\centering
\footnotesize\setlength{\tabcolsep}{4pt}
\caption{Macro test accuracy at $K=4$ (mean over the four test sets and three seeds): difference from the same model's plain loop (RELOAD), in points. --: not run. The single-pass row shows the size of the loop's own gain for reference.}
\label{tab:injection_macro}
\fitbox{\begin{tabular}{lrrrrr}
\toprule
Variant $-$ RELOAD & Ettin-400M & Qwen3-0.6B & Ettin-1B & Qwen3-1.7B & Von \\
\midrule
Single pass ($K=1$) & $-2.60$ & $-3.01$ & -- & -- & -- \\
\midrule
Diagonal, passes 2--$K$ & $+0.40$ & $-0.15$ & -- & -- & -- \\
Diagonal, every pass & $+0.12$ & $-0.32$ & $-0.48$ & $-0.12$ & $-0.24$ \\
Prefix source, trained gate & $-0.27$ & $-0.28$ & -- & -- & -- \\
Embedding source, trained gate & $+0.01$ & $-0.22$ & $-0.17$ & $+0.39$ & $+0.45$ \\
FiLM & $+0.09$ & $-0.55$ & $-0.21$ & $-0.03$ & $-0.18$ \\
Cross-attention & $+0.06$ & $-0.01$ & $-0.13$ & $+0.33$ & $+0.35$ \\
\bottomrule
\end{tabular}}
\end{table}

\paragraph{Group results and replication.}
Figure~\ref{fig:injection_summary} and Table~\ref{tab:injection_groups} give the paired group differences. No variant helps on the three new models: for Ettin-1B, Qwen3-1.7B and Von, no interval lies entirely above zero in any group, so the slight macro gains of cross-attention and the embedding write on Qwen3-1.7B and Von are within noise, and they reverse on Ettin-1B. The only intervals that exclude zero on the new models are negative: diagonal injection on Ettin-1B (OOD $-1.54$) and FiLM on Von (OOD $-1.39$). On the two small models the only positive intervals are out of distribution ($+0.59$ for the diagonal write on Ettin-400M and $+0.75$ for the trained prefix-source gate on Qwen3-0.6B), and the latter comes with $-2.43$ on longer reasoning.

\begin{table}[!htbp]
\centering
\scriptsize\setlength{\tabcolsep}{3pt}
\caption{Paired group accuracy difference (injection variant minus plain loop) at $K=4$, in points with 95\% bootstrap intervals. $^{*}$: interval excludes zero. Longer: longer reasoning; OOD: Kev transfer and zero-shot benchmarks.}
\label{tab:injection_groups}
\fitbox{\begin{tabular}{llrrr}
\toprule
Model & Variant & In & Longer & OOD \\
\midrule
Ettin-400M & Diagonal, passes 2--$K$ & $-0.12$ [$-0.46$, $+0.22$] & $+0.90$ [$-0.27$, $+2.07$] & $-0.83$ [$-3.57$, $+1.15$] \\
 & Diagonal, every pass & $-0.09$ [$-0.61$, $+0.40$] & $-0.03$ [$-1.57$, $+1.67$] & $+0.59^{*}$ [$+0.02$, $+1.15$] \\
 & Prefix source, trained gate & $-0.02$ [$-0.47$, $+0.49$] & $-0.37$ [$-2.67$, $+1.57$] & $-0.40$ [$-3.15$, $+1.55$] \\
 & Embedding source, trained gate & $+0.17$ [$-0.15$, $+0.48$] & $-0.23$ [$-2.07$, $+1.53$] & $-0.92^{*}$ [$-1.44$, $-0.36$] \\
 & FiLM & $-0.14$ [$-0.45$, $+0.17$] & $+0.20$ [$-2.03$, $+2.13$] & $-1.38^{*}$ [$-1.92$, $-0.82$] \\
 & Cross-attention & $-0.22$ [$-0.77$, $+0.32$] & $+0.10$ [$-1.40$, $+1.60$] & $-0.04$ [$-0.67$, $+0.64$] \\
\midrule
Qwen3-0.6B & Diagonal, passes 2--$K$ & $-0.37$ [$-1.00$, $+0.25$] & $-1.33$ [$-3.47$, $+0.60$] & $+0.01$ [$-2.50$, $+2.21$] \\
 & Diagonal, every pass & $+0.01$ [$-0.44$, $+0.43$] & $-2.10$ [$-4.37$, $+0.73$] & $+0.46$ [$-1.48$, $+2.19$] \\
 & Prefix source, trained gate & $-0.25$ [$-0.78$, $+0.28$] & $-2.43^{*}$ [$-3.90$, $-0.90$] & $+0.75^{*}$ [$+0.12$, $+1.34$] \\
 & Embedding source, trained gate & $+0.05$ [$-0.38$, $+0.50$] & $-0.87$ [$-3.37$, $+1.47$] & $-0.71$ [$-4.36$, $+2.45$] \\
 & FiLM & $-0.39$ [$-0.88$, $+0.12$] & $-1.30$ [$-3.20$, $+0.80$] & $-1.11^{*}$ [$-1.67$, $-0.52$] \\
 & Cross-attention & $-0.21$ [$-0.75$, $+0.32$] & $-0.90$ [$-4.57$, $+2.23$] & $-0.34$ [$-1.53$, $+0.64$] \\
\midrule
Ettin-1B & Diagonal, every pass & $-0.10$ [$-0.44$, $+0.25$] & $-0.47$ [$-1.63$, $+0.60$] & $-1.54^{*}$ [$-2.62$, $-0.27$] \\
 & Embedding source, trained gate & $-0.04$ [$-0.40$, $+0.31$] & $-0.43$ [$-1.47$, $+0.63$] & $-0.13$ [$-0.97$, $+0.69$] \\
 & FiLM & $-0.20$ [$-0.59$, $+0.16$] & $-1.03$ [$-2.17$, $+0.07$] & $-0.63$ [$-1.52$, $+0.19$] \\
 & Cross-attention & $-0.03$ [$-0.39$, $+0.35$] & $-0.80$ [$-1.83$, $+0.27$] & $+0.19$ [$-0.56$, $+0.97$] \\
\midrule
Qwen3-1.7B & Diagonal, every pass & $+0.08$ [$-0.39$, $+0.63$] & $+0.33$ [$-1.40$, $+2.07$] & $-0.61$ [$-1.74$, $+0.50$] \\
 & Embedding source, trained gate & $+0.30$ [$-0.15$, $+0.80$] & $+0.03$ [$-1.50$, $+1.53$] & $-0.25$ [$-1.01$, $+0.58$] \\
 & FiLM & $-0.23$ [$-0.69$, $+0.21$] & $+1.03$ [$-0.60$, $+2.73$] & $-0.38$ [$-1.81$, $+1.29$] \\
 & Cross-attention & $+0.08$ [$-0.29$, $+0.50$] & $-0.63$ [$-2.60$, $+1.33$] & $-0.25$ [$-1.06$, $+0.55$] \\
\midrule
Von & Diagonal, every pass & $-0.06$ [$-0.45$, $+0.30$] & $-1.00$ [$-2.70$, $+0.83$] & $-0.60$ [$-2.24$, $+0.86$] \\
 & Embedding source, trained gate & $+0.18$ [$-0.28$, $+0.60$] & $+0.37$ [$-1.67$, $+2.77$] & $-0.59$ [$-2.17$, $+0.73$] \\
 & FiLM & $-0.11$ [$-0.43$, $+0.21$] & $-0.27$ [$-1.60$, $+1.03$] & $-1.39^{*}$ [$-2.53$, $-0.32$] \\
 & Cross-attention & $+0.00$ [$-0.73$, $+0.69$] & $+0.27$ [$-1.60$, $+2.30$] & $-0.33$ [$-0.87$, $+0.15$] \\
\bottomrule
\end{tabular}}
\end{table}

\paragraph{Full-step control.}
Round 4 changes the update as well as the write, so its injection effects are paired against the full-step loop rather than the damped one. Full-step recurrence is similar to the damped loop on Ettin-400M but significantly reduces in-distribution and longer-reasoning accuracy on Qwen3-0.6B (macro $-2.63$ points; Table~\ref{tab:injection_fullstep}). Injection does not recover these losses (macro still $-2.32$ points with the diagonal write and $-1.89$ with cross-attention, relative to the damped loop). The significant out-of-distribution gains over the full step on Qwen3-0.6B leave its out-of-distribution accuracy within 0.6 points of the damped loop, a difference we did not test.

\begin{table}[!htbp]
\centering
\scriptsize\setlength{\tabcolsep}{3pt}
\caption{Round-4 paired group differences at $K=4$, in points with 95\% intervals. Injection arms are compared with the full-step loop, and the last comparison is the full step minus the damped loop (RELOAD). $^{*}$: interval excludes zero.}
\label{tab:injection_fullstep}
\fitbox{\begin{tabular}{llrrr}
\toprule
Comparison & Model & In & Longer & OOD \\
\midrule
Full-step + diagonal $-$ full-step & Ettin-400M & $-0.15$ [$-0.49$, $+0.19$] & $+0.77$ [$-0.37$, $+1.83$] & $-1.28$ [$-3.46$, $+1.50$] \\
 & Qwen3-0.6B & $-0.09$ [$-0.59$, $+0.41$] & $+0.30$ [$-2.03$, $+2.57$] & $+1.29^{*}$ [$+0.30$, $+2.22$] \\
Full-step + cross-attention $-$ full-step & Ettin-400M & $-0.33$ [$-0.73$, $+0.02$] & $+0.40$ [$-0.80$, $+1.63$] & $+0.95$ [$-0.75$, $+3.35$] \\
 & Qwen3-0.6B & $+0.12$ [$-0.54$, $+0.89$] & $+0.93$ [$-1.03$, $+2.97$] & $+0.97^{*}$ [$+0.41$, $+1.55$] \\
Full-step $-$ RELOAD & Ettin-400M & $+0.24$ [$-0.13$, $+0.65$] & $+0.53$ [$-1.07$, $+2.10$] & $+0.96$ [$-0.56$, $+2.33$] \\
 & Qwen3-0.6B & $-2.21^{*}$ [$-2.73$, $-1.68$] & $-6.27^{*}$ [$-8.13$, $-4.33$] & $-0.70$ [$-2.40$, $+0.54$] \\
\bottomrule
\end{tabular}}
\end{table}

\paragraph{Why the writes did not help.}
The state does not stay close to the prefix output $e$: after four passes its norm is about three times that of $e$, and its cosine similarity to $e$ falls below 0.8 (Table~\ref{tab:injection_drift}). The null result is therefore not explained by $e$ already being present in the loop state. Instead, switching the write off at inference on the development sets shows that the trained models make little use of it.

\begin{table}[!htbp]
\centering
\footnotesize\setlength{\tabcolsep}{5pt}
\caption{Top: state drift of the plain loop at $K=4$ ($e$ is the prefix output; mean over seeds and all tokens). Bottom: size of the write relative to the state, and change in development accuracy at $K=4$ when the trained write is switched off at inference (rounds 1--3; range over models and variants).}
\label{tab:injection_drift}
\begin{tabular}{lrrr}
\toprule
Model & $\|h_4-e\|/\|e\|$ & $\|h_4\|/\|e\|$ & $\cos(h_4,e)$ \\
\midrule
Ettin-400M & 2.28 & 2.92 & 0.78 \\
Qwen3-0.6B & 2.42 & 3.03 & 0.71 \\
\bottomrule
\end{tabular}

\vspace{4pt}
\fitbox{\begin{tabular}{lrr}
\toprule
Write & Size relative to state & Accuracy change when off \\
\midrule
Diagonal (all four variants) & 4--9\% & $-0.33$ to $+0.52$ \\
FiLM & 22--26\% & $-0.14$ to $+0.16$ \\
Cross-attention & 1--3\% & $0.00$ to $+0.10$ \\
\bottomrule
\end{tabular}}
\end{table}

Removing the write never costs more than 0.33 points, and on Ettin-400M it slightly helps, so the differences between injection arms and the plain loop reflect how the write changed training, not information supplied at inference. The diagonal gate stays at its initial values at the base learning rate; with a larger rate it trains on Ettin-400M, Ettin-1B, Qwen3-0.6B and Von, without a consistent accuracy benefit, and hardly moves on Qwen3-1.7B. FiLM writes a substantial fraction of the state, but disabling it barely changes accuracy, so most of it appears to be a rescaling absorbed by later layers; cross-attention learns a much smaller write. The switch-off test covers rounds 1--3 only, not the full-step variants or the three replication models, so the conclusion about inference use does not extend to those arms.

\paragraph{Summary.}
Across five pretrained models and the variants tested, input injection does not reliably improve in-distribution or longer-reasoning accuracy; isolated significant out-of-distribution gains do not reproduce on other models, and some variants significantly reduce out-of-distribution accuracy. This supports keeping injection out of RELOAD in this fine-tuning regime, but does not test the distinct setting of looped language models trained from scratch at longer depths and budgets, as in \citet{huang2026topology}.

\FloatBarrier
\section{Interventions on the trained loop}\label{app:interventions}

\paragraph{Question and setup.}
Appendix~\ref{app:injection} changes what the loop receives and retrains the model. Here we keep the trained weights fixed and change how long the loop runs, which tokens it may update and which input it sees, to find out which parts of the input the passes act on and how well the ODE view of Section~\ref{sec:method} describes the trained loop. We use the looped Von model of the first training seed and, where needed, its controller from Table~\ref{tab:main}. Since the right-hand side of Equation~\eqref{eq:ode} does not depend on $t$, the trajectory can be continued past the trained end point $t=1$, so we separate the step size $\Delta t$ from the number of passes $n$,
\begin{equation*}
h_{i+1} = h_i + \Delta t\,\bigl(G(h_i)-h_i\bigr), \qquad i=0,\dots,n-1,
\end{equation*}
which integrates to time $T=n\,\Delta t$. The budgets of the main text have $\Delta t=1/K$ and $n=K$, so $T=1$, and $T=0$ skips the looped block. Besides the held-out test sets, we generate object-tracking problems with up to 12 swaps (training covers up to 4; Table~\ref{tab:probe}). For StepGame, a strict graph parser recovers the chain of statements linking the two queried objects, and we call the statements off the chain distractors; this parser is stricter than the one behind Table~\ref{tab:hops}, so its chain-length bins differ slightly. Without an intervention, every experiment reproduces the ordinary outputs exactly. The results are descriptive, for three reasons: they come from a single seed, some comparisons were chosen after earlier results on the same test data had been seen, and the intervals resample items only.

\begin{figure}[!htbp]
\centering
\includegraphics[width=\linewidth]{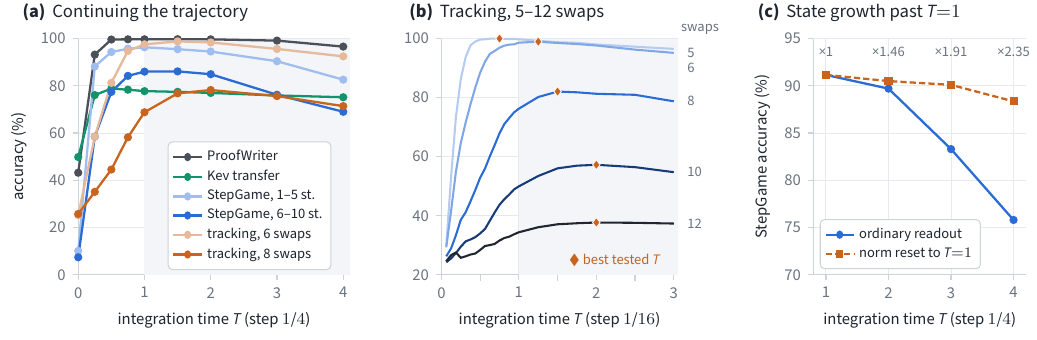}
\caption{Integration time of the trained loop (Von, seed 1, frozen weights). (a) Test accuracy along the four-pass trajectory ($\Delta t=1/4$), continued past the trained end point $T=1$ (shaded); StepGame is split by story length. (b) Generated tracking problems with 5 to 12 swaps along the trajectory with step $1/16$; diamonds mark the best tested time. (c) StepGame accuracy past $T=1$ with the ordinary readout and with every token rescaled to its norm at $T=1$ before the readout; numbers give the mean token norm relative to $T=1$.}
\label{fig:int_time}
\end{figure}

\paragraph{Integration time.}
More passes can be spent on smaller steps over the trained interval or on a longer trajectory. Halving the step at $T=1$ never lowers accuracy by more than 0.1 points and helps on the hardest problems (Section~\ref{sec:analysis}). A longer trajectory behaves differently (Figure~\ref{fig:int_time}a): on 8-swap tracking, far outside the training range, continuing to $T=2$ adds 9.4 points \ci{+7.8}{+11.1}, more than halving the step, whereas on the full test sets accuracy changes little up to $T=2$ and then falls, most on StepGame (about 15 points by $T=4$, short chains included). The best end time grows with the number of swaps (Figure~\ref{fig:int_time}b; selected on the test problems, so optimistic). The end time $T=1$ fixed in training is thus a choice rather than a property of the flow: problems beyond the training distribution can gain from integrating longer, while problems the loop already solves at $T=1$ get worse.

Much of this loss comes from the growing norm of the state, whose direction changes little: rescaling every token to its norm at $T=1$ before the readout recovers most, though not all, of it (Figure~\ref{fig:int_time}c). Choosing the end time per question is harder. A rule that continues only while the answer still changes matches $T=1$ on the test sets but keeps almost none of the gain of $T=2$ on 8-swap tracking, because most problems that improve later have not yet changed their answer at $T=1$; as for the post-hoc halting rules of Section~\ref{sec:controller}, a changing answer is a poor signal that more passes will help. Within the four-pass trajectory, the StepGame answer settles early, after little more than one pass on average and later for longer stories, consistent with Table~\ref{tab:probe}.

\paragraph{Where the loop writes.}
Figure~\ref{fig:int_writes}a--b and Section~\ref{sec:analysis} show that the gain requires updating both the shared tokens and the options; that within the story it comes from the chain rather than the distractors, with updating only the chain \ci{+26.3}{+31.9} points better than updating only the distractors (paired interval); and that it is kept when four passes over the shared tokens precede four over the options, which is possible because shared tokens never attend to the options. Attention knockouts point the same way: removing the attention of the options to the story inside the loop costs 10 points, whereas removing attention to the distractors costs nothing. Noise added to the chain tokens before the loop roughly halves accuracy, while the same noise on the more numerous distractor tokens changes it little, and in the later passes chain tokens receive larger updates than distractors. These accuracies are not isolated importance scores: frozen tokens create combinations of states that the suffix was not trained on, and the groups differ in size.

\begin{figure}[!htbp]
\centering
\includegraphics[width=\linewidth]{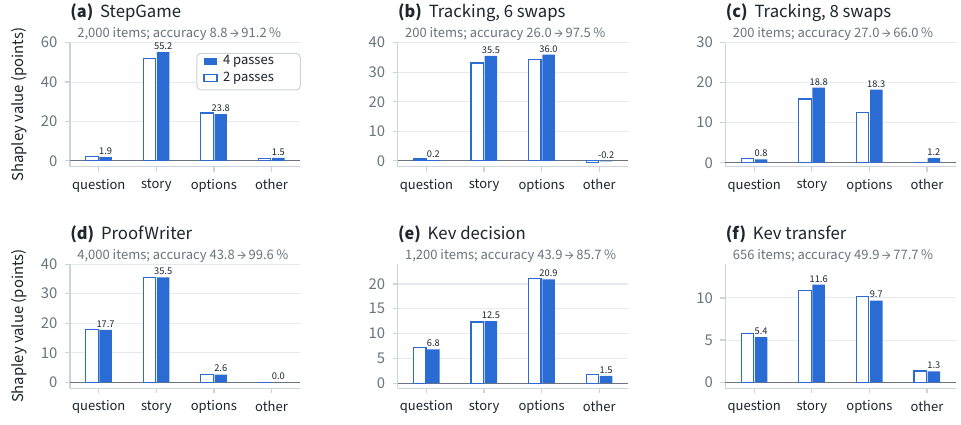}
\caption{Shapley decomposition of the loop's accuracy gain over token groups for every test set (Von, seed 1), with two passes (open bars) and four passes (filled bars, labeled). The value of a set of groups is the accuracy when only these groups are updated, so the values add up to the gain of the loop over no update, shown above each panel. Other: the remaining special tokens. Vertical scales differ between panels. ProofWriter uses 4,000 sampled test questions, and tracking 200 problems per depth.}
\label{fig:int_shapley}
\end{figure}

\begin{figure}[!htbp]
\centering
\includegraphics[width=\linewidth]{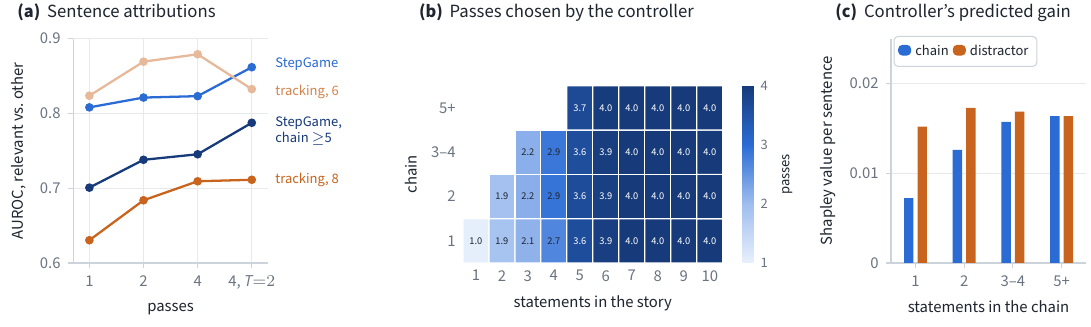}
\caption{Sentence attributions and the controller (Von, seed 1). (a) AUROC with which sentence-deletion Shapley values of the log-probability of the correct option separate relevant sentences from the others, for $K$ passes and for the four-pass trajectory continued to $T=2$. Relevant sentences are chain statements, or swaps whose removal alone changes the answer. (b) Mean number of passes chosen by the controller of Table~\ref{tab:main} on StepGame stories with a unique chain, by story and chain length; empty cells cannot occur. (c) Shapley value per sentence of the controller's predicted gain $\hat P_{4}-\hat P_{1}$ for chain statements and distractors, by chain length.}
\label{fig:int_attr}
\end{figure}

\paragraph{Shapley values.}
Restricted updates and deleted sentences interact: two groups may help only together, so removing or keeping one group alone does not measure its contribution. We therefore use Shapley values \citep{shapley1953value}: for players $N=\{1,\dots,n\}$ and a value $v(S)$ for every coalition $S\subseteq N$, player $i$ receives its marginal contribution averaged over all orders in which the players can join,
\begin{equation*}
\phi_i=\sum_{S\subseteq N\setminus\{i\}}\frac{|S|!\,(n-|S|-1)!}{n!}\,\bigl(v(S\cup\{i\})-v(S)\bigr).
\end{equation*}
the only attribution that is efficient ($\sum_i\phi_i=v(N)-v(\emptyset)$), symmetric, additive over games and zero for players that never change $v$. The pairwise interaction index \citep{grabisch1999axiomatic} measures what two players contribute together beyond their separate values. A game fixes the players, what absence means and the value; we use four:
(i) \emph{Token groups} (Figures~\ref{fig:int_writes}c and~\ref{fig:int_shapley}): the question, the story (or its chain and distractor statements), the options and the remaining special tokens. An absent group is frozen at its state before the loop, and the value is the correctness, or the log-probability of the correct option, with two or four passes. Averaged over questions, the correctness values split the accuracy gain of the loop over no update among the groups.
(ii) \emph{Sentences} (Figure~\ref{fig:int_attr}a): the sentences of a StepGame story or the swaps of a tracking problem. An absent sentence is deleted from the input, and the value is the log-probability of the correct option. We use 243 StepGame stories, in which long chains are overrepresented, and 200 tracking problems each with 6 and 8 swaps.
(iii) \emph{Controller, sentences} (Figure~\ref{fig:int_attr}c): game (ii) with the controller's predicted gain $\hat P_{4}-\hat P_{1}$ as the value.
(iv) \emph{Controller, feature blocks}: the controller's four feature blocks, namely the pooled story state, the pooled option-marker state, the log input length and the number of options. An absent block takes its values from 100 background questions, which gives interventional Shapley values \citep{lundberg2017unified,janzing2020feature}.
All values are computed exactly over the $2^n$ coalitions. They describe dependence under the intervention that defines absence, not a causal mechanism: deleted sentences can break the chain, frozen tokens produce combinations of states the model was not trained on, and the feature-block values depend on the background. The values are also measured against no loop rather than against one pass, and since a group can receive a large value simply because it has many members, for sentences we report means per sentence.

\paragraph{Token groups.}
Figure~\ref{fig:int_shapley} splits the loop's accuracy gain among the token groups for every test set. On StepGame the split is the one summarized in Section~\ref{sec:analysis} (Figure~\ref{fig:int_writes}c); for longer chains the share of the options shrinks, and the strongest pairwise interaction is between the chain and the options. On object tracking, story and options contribute about equally, and on 8-swap tracking four passes raise both contributions over two passes, in line with the accuracy gain. On ProofWriter, the story contributes most and the question, which states what is to be proved, about half as much, while the fixed options true, false and unknown contribute little. On Kev decision, the options contribute more than the story. Which groups matter thus depends on where the task places the content of the decision.

\paragraph{Which sentences the answer depends on.}
In the sentence game (ii), a StepGame sentence is relevant if it lies on the chain, and a swap if removing it alone changes the correct answer. With more passes, the attributions concentrate on the relevant sentences (Figure~\ref{fig:int_attr}a), most for long chains and 8-swap tracking; continuing to $T=2$ separates the StepGame sentences further, without improving accuracy, and the 6-swap tracking sentences less well. The attributions also reveal a surface cue: irrelevant swaps that mention the queried object receive larger attributions than other irrelevant swaps, and off-chain StepGame sentences that mention a queried object lower the probability of the correct answer. Since these groups also differ in position and context, this does not isolate an effect of the name alone.

\begin{figure}[!htbp]
\centering
\includegraphics[width=0.62\linewidth]{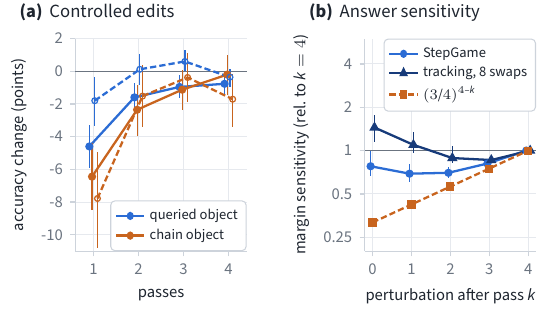}
\caption{Controlled edits and answer sensitivity (Von, seed 1). (a) Change in StepGame test accuracy when an off-chain sentence is attached to a queried object or to an object inside the chain, relative to the same sentence with a letter that occurs nowhere else in the story. Solid: test-story sentences that mention a queried object; dashed: edited sentences that mention none; bars: 95\% intervals. (b) Sensitivity of the margin between the correct and the strongest other option to a perturbation after pass $k$ of four, in the metric of Figure~\ref{fig:int_noise} and relative to $k=4$ (medians and interquartile ranges over 60 StepGame stories and 20 tracking problems), with the reference $(3/4)^{4-k}$ for a block that ignores the perturbation.}
\label{fig:int_cue}
\end{figure}

\paragraph{A controlled test of the cue.}
To isolate the name, we change one letter in one off-chain StepGame sentence, keeping its position, template and other object fixed, so that it mentions a queried object, an object inside the chain, or a letter that occurs nowhere else in the story (Figure~\ref{fig:int_cue}a); every edited story is parsed again and kept only if its chain, and therefore its answer, is unchanged. A sentence that mentions a queried object is no more harmful than the same sentence attached to an object inside the chain: with one pass, what costs accuracy is attaching an irrelevant sentence to the story at all. Relative to a letter that occurs nowhere else, the test-story sentences that mention a queried object cost 4.6 points with one pass and less than one point with four, so the cue is structural rather than lexical, and the passes largely remove it. Edits that close a cycle around the queried object, which can make the story contradictory, cost about 20 points even with four passes and are not shown. The negative attributions of game (ii) may thus arise mainly in coalitions with an incomplete chain, where a sentence attached to a queried object offers a wrong route.

\paragraph{What the controller reads.}
The controller of Table~\ref{tab:main} was released with this checkpoint, so we analyze it directly. On StepGame it matches the accuracy of four passes to within 0.1 points with 79\% of their passes. As in Section~\ref{sec:controller}, its budget grows with story length, from one pass for single-statement stories to four from seven statements on, and for a fixed story length it hardly depends on chain length (Figure~\ref{fig:int_attr}b); the rise with chain length in Table~\ref{tab:hops} comes from long chains occurring in long stories. Sentence attributions of the predicted gain (game iii) agree: chain statements and distractors receive almost equal values per sentence, for short chains the distractors even more (Figure~\ref{fig:int_attr}c), and the attributions do not separate chain statements from other sentences (AUROC 0.46, against 0.82 for the model's own answer). Over the feature blocks (game iv), the two pooled states carry almost all of the attribution and the explicit length and option-count features almost none, so story length reaches the controller through the hidden states. A controller of the same form that we had refitted on development questions before the weights were released gives the same picture.

\begin{figure}[!htbp]
\centering
\includegraphics[width=0.62\linewidth]{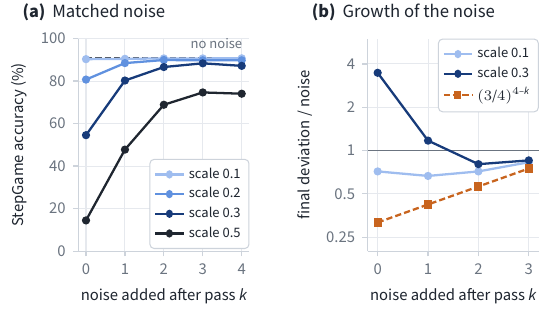}
\caption{Matched perturbations of the loop state (Von, seed 1, StepGame, four passes). (a) Accuracy when the same Gaussian perturbation, scaled by the per-feature standard deviation of the state before the loop, is added after pass $k$. (b) Norm of the resulting deviation of the final state relative to the inserted perturbation, with the reference $(3/4)^{4-k}$ for a block that ignores the perturbation.}
\label{fig:int_noise}
\end{figure}

\paragraph{Perturbations of the state.}
Finally, we add the same Gaussian perturbation, scaled by the per-feature standard deviation of the state before the loop, after pass $k$ of four on StepGame (Figure~\ref{fig:int_noise}); early perturbations do the most damage. A block that ignored the perturbation would keep $3/4$ of it per remaining damped step, $(3/4)^{4-k}$ at the end. The measured deviation is larger than this reference at every scale and insertion point before the last pass, and at larger scales it even exceeds the inserted perturbation. Unlike contractive dynamics near an attracting fixed point \citep{zela2026sfprm}, the trained loop thus behaves like an initial-value problem whose end state depends on $h_0$ (Equation~\eqref{eq:ode}), consistent with the drift of the state away from the anchor in Table~\ref{tab:injection_drift}. Distances between states are not accuracy, however, so this does not show whether the loop repairs perturbations in the directions that matter for the answer.

\paragraph{Linearization of the loop.}
To see which directions grow and whether they matter for the answer, we linearize the loop along the four-pass trajectory with automatic differentiation on 60 StepGame stories and 20 problems with 8 swaps, computing leading eigenvalues on 26 of them. The Jacobian of the field $F(h)=G(h)-h$ of Appendix~\ref{app:ode} has an eigenvalue with positive real part in every item and at every time; its largest real part falls from about 3 at $t=0$ to about 1 at $t=2$, whereas a block that ignored its input would have all eigenvalues at $-1$. The leading eigenvector lies on the question and story tokens in almost every item. The first step is the stiffest, with $\Delta t$ times the largest eigenvalue modulus close to 1, consistent with the gain of the midpoint method (Section~\ref{sec:analysis}), and the field keeps pushing the state outward, so its norm grows almost linearly with $T$ (Figure~\ref{fig:int_time}c). The linearized loop shrinks a random perturbation as much as measured at the smallest noise scale, but stretches a few directions by factors of 20 to 80, including directions that matter for the answer (Figure~\ref{fig:int_cue}b): a perturbation before the loop moves the margin between the correct and the strongest other option about three quarters as much as the same perturbation after the last pass on StepGame, and more on 8-swap tracking, whereas passive damping would reduce it to a third. The trained loop therefore carries perturbations in these directions forward, as the solution of an initial-value problem does, rather than repairing them.

\FloatBarrier
\section{Loop structure}\label{app:loopvariants}

\paragraph{Question and setup.}
RELOAD loops a block of nine consecutive layers as a whole: with layers $A,B,C$ the order of computation is $ABCABC$. We test on Von whether looping smaller units, a different update or a learned readout of the passes improves on this. Every arm uses the recipe of the main experiments with budgets $K\in\{1,2,3,4\}$ and is paired with a retrain of RELOAD on Von on the same cluster, with the same data and seeds. These experiments were run while the design was explored, on development data only; no test set was evaluated for any of them. The development files form three groups: \emph{longer reasoning} (four files: StepGame stories with up to 10 statements, deeper synthetic problems, longer CLUTRR chains and a deeper development set), \emph{Kev} (the Kev decision and transfer development sets) and nine further development sets; \emph{all} is the macro mean over these 15 files, without the LogiQA development set. From the second round on, a decision rule was fixed before training: a variant helps if the paired interval on longer reasoning at matched compute lies above zero and the interval over all files does not lie entirely below zero, and it hurts if the interval on longer reasoning lies below zero. All arms have three seeds; most layer-loop arms were first run with one seed and stopped when they fell far below the baseline, and their second and third seeds, as well as the retrained baseline, were added afterwards for this table.

\paragraph{Variants.}
\emph{Layer loop}: each layer is looped in place before the next, as in the per-layer loops of \citet{frey2026adaptive}, applying $h\leftarrow h+s\,(L_i(h)-h)$ $K$ times per layer $L_i$, which gives the order $AA\,BB\,CC$ at the same compute as the block loop, with step $s=1/K$ as in RELOAD, $s=K^{-1/2}$, or $s=1$ (full step). With \emph{learned output weights}, the output of each looped layer is a learned softmax combination of its input and its $K$ pass outputs, separate for each budget and initialized at the weights implied by the damped update. We loop either layers 9--17 or all 28 layers; the latter is compared at $K=2$, which costs about the same as the baseline at $K=4$ (56 instead of 55 layer applications), and trains every layer at the learning rate of the looped block. \emph{Units}: layers 9--17 are split into consecutive units, each running its $K$ damped passes before the next starts; a single unit of nine layers is RELOAD and nine units of one layer are the layer loop, all at the same compute. \emph{Smaller block}: only layers 10--16 loop, and layers 9 and 17 run once. \emph{Update and readout}: the full step $h_{k+1}=G(h_k)$ without damping, and the same loop with an ACT-style weighted output \citep{graves2016adaptive}, using halting weights in the form of PonderNet \citep{banino2021pondernet} from a per-token router similar to the learned halting of \citet{frey2026adaptive}. After pass $k=1,\dots,K$, a linear router on the state and the relative pass index gives each token a halting probability $\eta_k=\sigma\bigl(w^\top[h_k;\,(k-1)/\max(K-1,1)]+b\bigr)$. The output is $\sum_k \pi_k h_k$ with $\pi_k=\eta_k\prod_{j<k}(1-\eta_j)$, and the last pass of the budget receives the remaining weight, so that $K=1$ is the single pass.

\begin{table}[!htbp]
\centering
\scriptsize\setlength{\tabcolsep}{3pt}
\caption{Loop-structure ablations on Von, development data only (no test set was evaluated). Longer: longer-reasoning development group; All: macro mean over the 15 development files. Accuracy in \%, mean over the listed seeds. Layers: layer applications per question at the reported budget, $K=2$ for the arms over all 28 layers and $K=4$ otherwise. Gain: longer-reasoning accuracy at that budget minus at $K=1$. $\Delta$: paired difference from the retrained block loop at $K=4$, with 95\% intervals from a bootstrap over seeds and items that resamples seeds separately for each development file. $^{*}$: interval excludes zero. Bold: best per column.}
\label{tab:loop_variants}
\fitbox{\begin{tabular}{lrrrrrrrr}
\toprule
Variant & Seeds & Layers & Longer $K{=}1$ & Longer & Gain & All & $\Delta$ Longer & $\Delta$ All  \\
\midrule
\multicolumn{9}{l}{\textit{Baseline}} \\
\rowcolor{reloadbg}Block loop, layers 9--17 (RELOAD) & 3 & 55 & 69.95 & \textbf{78.70} & $\mathbf{+8.74}$ & \textbf{71.24} & -- & --  \\
\midrule
\multicolumn{9}{l}{\textit{Each layer of 9--17 looped in place}} \\
Damped, step $1/K$ & 3 & 55 & 69.39 & 70.90 & $+1.51$ & 67.98 & $-7.79^{*}$ \ci{-8.95}{-6.60} & $-3.26^{*}$ \ci{-3.90}{-2.61}  \\
Learned output weights & 3 & 55 & 69.73 & 71.43 & $+1.70$ & 68.69 & $-7.26^{*}$ \ci{-8.22}{-6.32} & $-2.55^{*}$ \ci{-3.26}{-1.83}  \\
Step $K^{-1/2}$ & 3 & 55 & 70.38 & 71.50 & $+1.13$ & 68.02 & $-7.19^{*}$ \ci{-8.55}{-5.72} & $-3.22^{*}$ \ci{-3.93}{-2.51}  \\
Full step & 3 & 55 & 70.60 & 71.38 & $+0.78$ & 67.51 & $-7.31^{*}$ \ci{-8.71}{-5.99} & $-3.73^{*}$ \ci{-4.46}{-2.97}  \\
\midrule
\multicolumn{9}{l}{\textit{Each of the 28 layers looped in place}} \\
Damped, step $1/K$ & 3 & 56 & 69.77 & 71.30 & $+1.53$ & 66.81 & $-7.39^{*}$ \ci{-8.85}{-5.99} & $-4.43^{*}$ \ci{-5.20}{-3.64}  \\
Learned output weights & 3 & 56 & 70.99 & 72.29 & $+1.30$ & 67.84 & $-6.41^{*}$ \ci{-7.39}{-5.41} & $-3.40^{*}$ \ci{-4.16}{-2.60}  \\
Step $K^{-1/2}$ & 3 & 56 & 70.18 & 71.58 & $+1.41$ & 67.08 & $-7.11^{*}$ \ci{-8.74}{-5.44} & $-4.16^{*}$ \ci{-5.07}{-3.39}  \\
Step $K^{-1/2}$, learned weights & 3 & 56 & 68.89 & 70.44 & $+1.55$ & 66.46 & $-8.25^{*}$ \ci{-9.63}{-6.98} & $-4.77^{*}$ \ci{-5.48}{-4.05}  \\
\midrule
\multicolumn{9}{l}{\textit{Layers 9--17 split into looped units}} \\
$1+7+1$ & 3 & 55 & 70.36 & 78.09 & $+7.73$ & 70.49 & $-0.60$ \ci{-1.60}{+0.45} & $-0.75^{*}$ \ci{-1.37}{-0.10}  \\
$4+5$ & 3 & 55 & 69.98 & 77.92 & $+7.94$ & 70.74 & $-0.78$ \ci{-1.64}{+0.10} & $-0.50$ \ci{-1.11}{+0.09}  \\
$2+5+2$ & 3 & 55 & 70.58 & 77.36 & $+6.78$ & 70.18 & $-1.34$ \ci{-2.96}{+0.33} & $-1.06^{*}$ \ci{-1.77}{-0.35}  \\
$3+3+3$ & 3 & 55 & 68.46 & 74.37 & $+5.91$ & 69.49 & $-4.32^{*}$ \ci{-5.71}{-3.01} & $-1.75^{*}$ \ci{-2.45}{-1.09}  \\
$2+2+2+2+1$ & 3 & 55 & 70.03 & 74.00 & $+3.98$ & 69.42 & $-4.69^{*}$ \ci{-5.94}{-3.54} & $-1.82^{*}$ \ci{-2.50}{-1.13}  \\
\midrule
\multicolumn{9}{l}{\textit{Smaller block}} \\
Block loop, layers 10--16 & 3 & 49 & 69.58 & 78.19 & $+8.60$ & 70.90 & $-0.51$ \ci{-1.34}{+0.36} & $-0.34$ \ci{-0.93}{+0.22}  \\
\midrule
\multicolumn{9}{l}{\textit{Update and readout}} \\
Full step $h_{k+1}=G(h_k)$ & 3 & 55 & 69.92 & 77.21 & $+7.29$ & 70.32 & $-1.48$ \ci{-2.95}{+0.25} & $-0.92^{*}$ \ci{-1.70}{-0.11}  \\
Full step, ACT-weighted output & 3 & 55 & 69.51 & 77.05 & $+7.53$ & 70.09 & $-1.65^{*}$ \ci{-3.16}{-0.02} & $-1.14^{*}$ \ci{-1.95}{-0.33}  \\
\bottomrule
\end{tabular}}
\end{table}

\paragraph{Looping single layers.}
Looping each layer in place removes most of the gain from additional passes (Table~\ref{tab:loop_variants}). From $K=1$ to the matched budget, longer-reasoning accuracy rises by 0.8 to 1.7 points in all eight layer-loop arms, against 8.7 for the block loop, for every step size, with and without learned output weights, and whether 9 or 28 layers loop. All eight are 6.4 to 8.3 points less accurate than the block loop on longer reasoning and 2.6 to 4.8 points over all files. Since accuracy at $K=1$ is similar to the baseline (68.9 to 71.0 against 70.0), the loss lies in what the additional passes add. The state does move: on 600 StepGame development stories of all lengths at $K=4$, the looped section moves the state 3.0 times as far as one pass with step $K^{-1/2}$, with each layer moving it 2.1 times as far as one application of that layer, and 6.8 times as far with the full step (two seeds each), against 1.50 times for the block loop (three seeds). These two arms nevertheless gain only 1.1 and 0.8 points, so a smaller update does not explain the deficit: in the layer loop, later layers of the block never feed back into earlier ones.

\paragraph{Unit size.}
Splitting layers 9--17 into units interpolates between the two orders. Units of seven or five layers keep most of the gain (6.8 to 7.9 points), units of three or two layers lose much of it (5.9 and 4.0), and the $3+3+3$ and $2+2+2+2+1$ splits are 4.3 and 4.7 points less accurate than the block loop on longer reasoning. Over all files every split except $4+5$ is significantly less accurate, by 0.75 to 1.8 points, and no split beats the block loop.

\paragraph{Smaller block.}
Looping only layers 10--16, with layers 9 and 17 run once, uses 49 instead of 55 layer applications at $K=4$ and is not significantly different from the block loop on longer reasoning ($-0.51$ \ci{-1.34}{+0.36}) or over all files ($-0.34$ \ci{-0.93}{+0.22}). The intervals do not exclude a loss of about one point, and we did not evaluate this variant on test data.

\paragraph{Damping and a learned readout.}
Without damping, Von loses little, $-1.48$ \ci{-2.95}{+0.25} on longer reasoning and $-0.92$ \ci{-1.70}{-0.11} over all files, whereas Qwen3-0.6B loses 6.3 points (Table~\ref{tab:injection_fullstep}). The ACT-weighted output does not improve the full-step loop ($-0.17$ \ci{-1.59}{+1.18} on longer reasoning) and is 1.65 points \ci{-3.16}{-0.02} below the damped loop, although the router does learn input-dependent weights. At $K=4$ it gives the option markers almost all weight on passes 1 and 2 (at most about 2\% on passes 3 and 4), and the weight on pass 2 rises with the number of StepGame statements, from 0.00--0.10 for one statement to 0.42--0.61 for nine (range over three seeds). Over all tokens, passes 3 and 4 together receive 6 to 20\% of the weight on StepGame and 28 to 44\% on the deeper synthetic problems, so what they compute reaches the markers through the context tokens in the layers after the loop. A learned readout thus adapts to the input but is not more accurate than the endpoint of the damped loop.

\paragraph{Summary.}
Of the structures tested, the damped loop over a block of layers is the most accurate. Most of the gain requires a looped unit of several layers: units of five or more layers keep most of it, and smaller units lose a substantial part. Removing the damping costs about a point over all files, and neither learned output weights nor an ACT readout improved accuracy. With 16 arms, single intervals that exclude zero are expected by chance, so we read only the consistent patterns: the loss from small units and the absence of any gain over the block loop.

\FloatBarrier
\end{document}